\documentclass{article}

\PassOptionsToPackage{numbers,compress}{natbib}
\usepackage[final,main]{neurips_2026}

\usepackage[utf8]{inputenc}
\usepackage[T1]{fontenc}
\usepackage{hyperref}
\usepackage{url}
\usepackage{graphicx}
\usepackage{booktabs}
\usepackage{amsfonts}
\usepackage{amsmath}
\usepackage{mathtools}
\usepackage{array}
\usepackage{multirow}
\usepackage{threeparttable}
\usepackage{microtype}
\usepackage{xcolor}
\usepackage{colortbl}
\usepackage{float}
\usepackage{placeins}
\usepackage{subcaption}
\usepackage{tabularx}
\usepackage{wrapfig}
\usepackage{needspace}
\usepackage{adjustbox}

\graphicspath{{./}}

\definecolor{iclrcolor}{RGB}{34,139,34}

\DeclareRobustCommand{\eg}{\emph{e.g.}}
\newcommand{\MMdb}[2]{%
  \left\langle\!\left\langle #1 \right\rangle\!\right\rangle_{#2}%
}

\title{RIFLE: Removal of Image Flicker-Banding \\ via Latent Diffusion Enhancement}
\author{
  Libo Zhu$^{1}$\thanks{Equal contribution}~,\enspace
  Zihan Zhou$^{1}$\footnotemark[1]~,\enspace
  Xiaoyang Liu$^{1}$,\enspace
  Zhiyi Zhou$^{1}$,\enspace \\
  \textbf{Weihang Zhang}$^{2}$,\enspace
  \textbf{Keyu Shi}$^{2}$,\enspace 
  \textbf{Yifan Fu}$^{2}$,\enspace
  \textbf{Yulun Zhang}$^{1}$\thanks{Corresponding authors:
  Yulun Zhang, yulun100@gmail.com} \enspace \\
  \textsuperscript{1}Shanghai Jiao Tong University,\enspace
  \textsuperscript{2}Institute of Media Technology and Experience Design, Huawei \enspace
  \vspace{-8mm}
}

\begin{document}

\maketitle
\parfillskip=0pt plus .2\linewidth

\begin{abstract}
Capturing screens is common, but photos of emissive displays are often influenced by \emph{flicker-banding} (FB), some alternating bright--dark stripes due to temporal aliasing between a camera's rolling-shutter readout and display's brightness modulation. Unlike moir\'e degradation, FB remains underexplored despite its frequent and severe impact on readability and perceived quality. We formulate FB removal as a dedicated restoration task and introduce \textbf{R}emoval of \textbf{I}mage \textbf{F}licker-\textbf{B}anding via \textbf{L}atent Diffusion \textbf{E}nhancement, \textbf{\emph{RIFLE}}, a diffusion-based framework designed to remove FB with fine details. We introduce a task-specific \textbf{\emph{Banding-Suppressed High-Frequency Prior (BSHP)}}, which combines gradient-based high-frequency localization with a smooth structural support map to build a compact restoration prior, and injects it into the restoration backbone via multi-stage FiLM modulation. Moreover, a task-specific \textbf{\emph{Masked Loss (ML)}} uses the mask from FB simulation to concentrate supervision on banded regions while preserving global fidelity. To overcome data scarcity, we provide a \textbf{\emph{simulation pipeline}} synthesizing FB in the luminance domain with stochastic jitter in banding angle, spacing, and width. Feathered boundaries and sensor noise are also applied for a more realistic simulation. For evaluation, we collect a \textbf{\emph{paired real-world FB dataset}} with pixel-aligned banding-free references captured via long exposure. Across quantitative metrics and visual comparisons on our real-world dataset, RIFLE consistently outperforms recent image reconstruction baselines. To our knowledge, this is \textbf{\emph{the first work}} on FB simulation and removal. Our dataset and code will be released at \url{https://github.com/libozhu03/RIFLE}
\end{abstract}
\interlinepenalty=10000
\clubpenalty=10000
\widowpenalty=10000
\parfillskip=0pt plus .2\linewidth

\begin{figure*}[t]
  \centering
  \includegraphics[width=\textwidth]{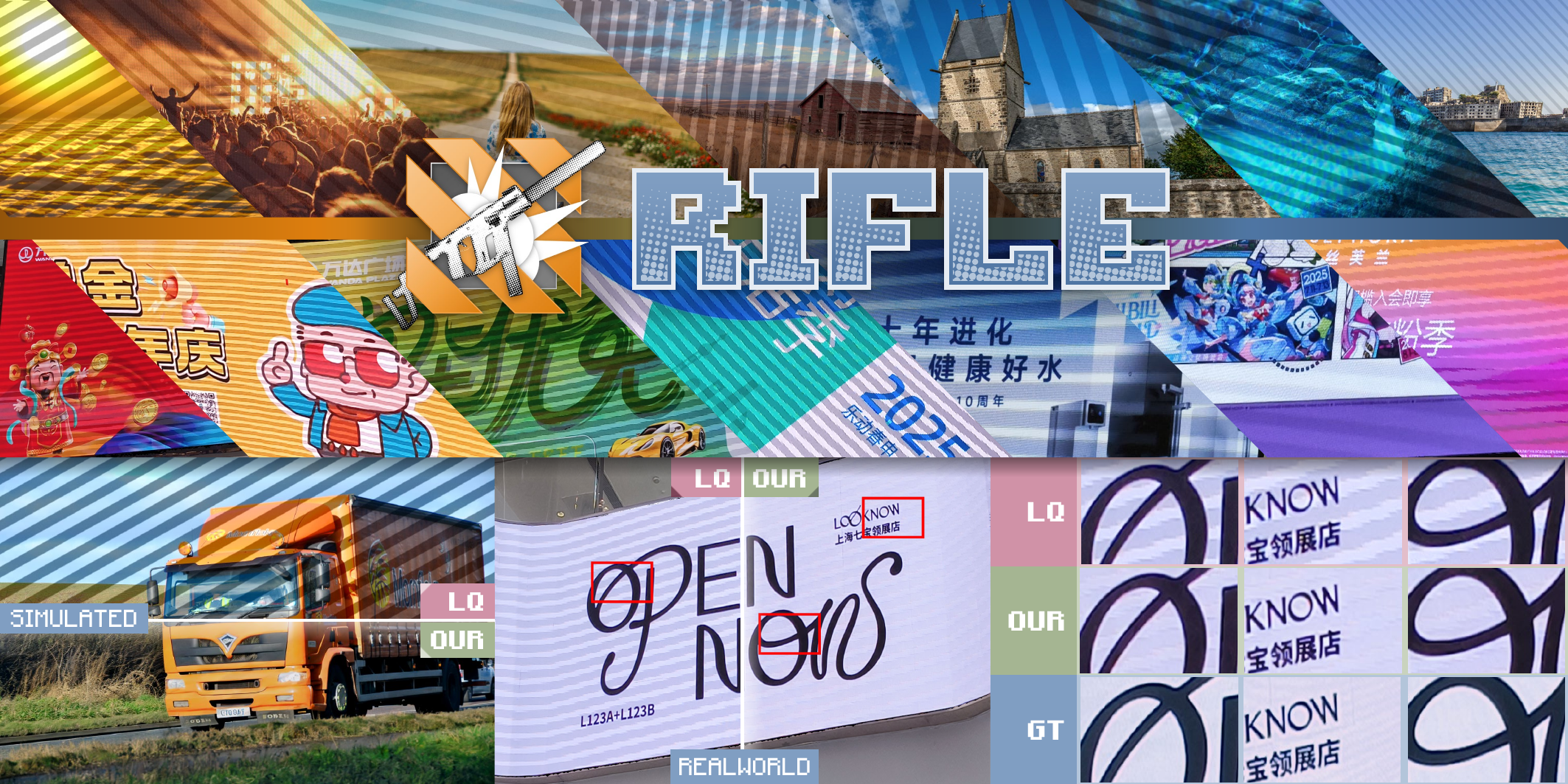}
  \caption{
    Overview of \textbf{\emph{RIFLE}}. The top part shows the datasets we build, including the simulated training dataset (first row) and the real-world testing dataset (second row). The bottom part shows the flicker-banding removal effect on the simulated datasets and real-world scenarios.
  }
  \vspace{-3mm}
  \label{fig:overview_our_work}
\end{figure*}

\vspace{-5mm}
\section{Introduction}
\vspace{-3mm}
Capturing screens is routine in our everyday life: \textup{(\romannumeral1)} students photograph lecture slides on projectors in class, \textup{(\romannumeral2)} professionals document dashboards and error messages on monitors, \textup{(\romannumeral3)} creators share phone or smartwatch interfaces, \textup{(\romannumeral4)} commuters record LED billboards or vehicle clusters, etc. Despite impressive progress in mobile imaging, photos of emissive displays remain prone to characteristic degradations. Among these, moir\'e~\cite{zhang2023realtimeimagedemoireingmobile,xu2023image,liu2025freqformer,mei2025imagedemoireingusingdual,yang2025dsdnetrawdomaindemoireing} from spatial interference between the subpixel display lattice and the camera sampling grid has been widely studied, producing effective learning-based remedies. However, our empirical survey of real-world captures reveals a different, underexplored image degradation that frequently dominates visual quality: \textbf{\emph{Flicker-banding}}.

Flicker-banding (FB) appears as alternating bright and dark stripes that traverse images, often approximately horizontal but sometimes tilted or warped. The cause is temporal: most commodity smartphone sensors employ electronic rolling shutters that expose rows sequentially, while modern displays modulate luminance in time via pulse-width modulation (PWM) or scanning refresh.

When the camera's line readout cadence aliases the display's time-varying emission, the temporal mismatch collapses into spatial striping within a single frame. The perceptual impact is so substantial that FB largely distracts attention, obscures screen elements and small fonts, and distorts tone and contrast. FB often degrades the perceived quality of screen photographs and may result in an unsatisfactory viewing experience, especially when small screen elements must remain legible.

Removing FB is an inherent challenge for three reasons.
\textup{(\romannumeral1)} \textbf{\emph{Missing screen-side metadata.}} During photography, the camera has no access to the screen's driving parameters (\eg, PWM frequency, duty cycle, scanning order), which are device dependent and mode dependent. Therefore, hardware-side FB mitigation solutions are extremely hard.
\textup{(\romannumeral2)} \textbf{\emph{Various morphological types.}} Stripe orientation, spacing, duty cycle, and contrast depend jointly on the sensor readout speed, exposure, gain, and display technology and settings. Different parameters yield diverse and non-stationary patterns that strain single-prior restorers.
\textup{(\romannumeral3)} \textbf{\emph{Partial information loss.}} In severe cases, the dark phase of the modulation produces near-black bars that overwrite scene content. Successful restoration must reason about missing structures, not merely denoise or deblur. All these factors differentiate FB from the classic moir\'e or global deflicker and call for a task-specific image reconstruction solution.

These challenges motivate a dedicated learning-based solution for single-image FB removal. To the best of our knowledge, this is the \textbf{first} work to formulate the task with neural networks. Since large-scale paired training data are unavailable, we first develop a \textbf{\emph{simulation pipeline}} that synthesizes realistic FB in the luminance domain with stochastic variations in banding angle, spacing, and width, together with feathered transitions and sensor noise. For rigorous evaluation, we further collect a paired \textbf{\emph{real-world dataset}} with pixel-aligned banded observations and banding-free references. Building on PiSA-SR~\cite{sun2024pisasr}, we then propose \textbf{R}emoval of \textbf{I}mage \textbf{F}licker-banding via \textbf{L}atent diffusion \textbf{E}nhancement (\textbf{\emph{RIFLE}}), which introduces two task-specific components: \textup{(\MakeUppercase{\romannumeral1})} \textbf{\emph{Banding-Suppressed High-Frequency Prior (BSHP)}}, which provides structural guidance by combining fine-grained high-frequency responses with smooth local support, and \textup{(\MakeUppercase{\romannumeral2})} \textbf{\emph{Masked Loss} (ML)}, which emphasizes banded regions while preserving global fidelity. Experiments on our real-world FB dataset show that RIFLE consistently outperforms recent image reconstruction baselines. 

As shown in Fig.~\ref{fig:overview_our_work}, our main contributions to single-image FB removal are as follows:

\begin{itemize}
\item To the best of our knowledge, it is \textbf{\emph{the first work}} to research the simulation and removal of the flicker-banding (FB) and establishes a strong foundation for future research.
\item We propose a FB simulation pipeline to build a large-scale training dataset and real-world testing dataset for evaluating the FB removal model's performance in practical settings.
\item We present \textbf{\emph{RIFLE}}, a one-step diffusion-based model with a Banding-Suppressed High-Frequency Prior (\textbf{\emph{BSHP}}) that highlights informative high-frequency transitions for task-specific prior-guided FiLM modulation, and a Masked Loss (\textbf{\emph{ML}}) tailored to FB artifacts.
\item Our RIFLE achieves substantial improvement over other recent image reconstruction methods in both the quantitative metrics and visual comparisons. RIFLE is capable of addressing various kinds of FB patterns and can also be directly applied to real-world scenarios.
\end{itemize}

\begin{figure*}[t]
  \centering
  \includegraphics[width=0.99\linewidth]{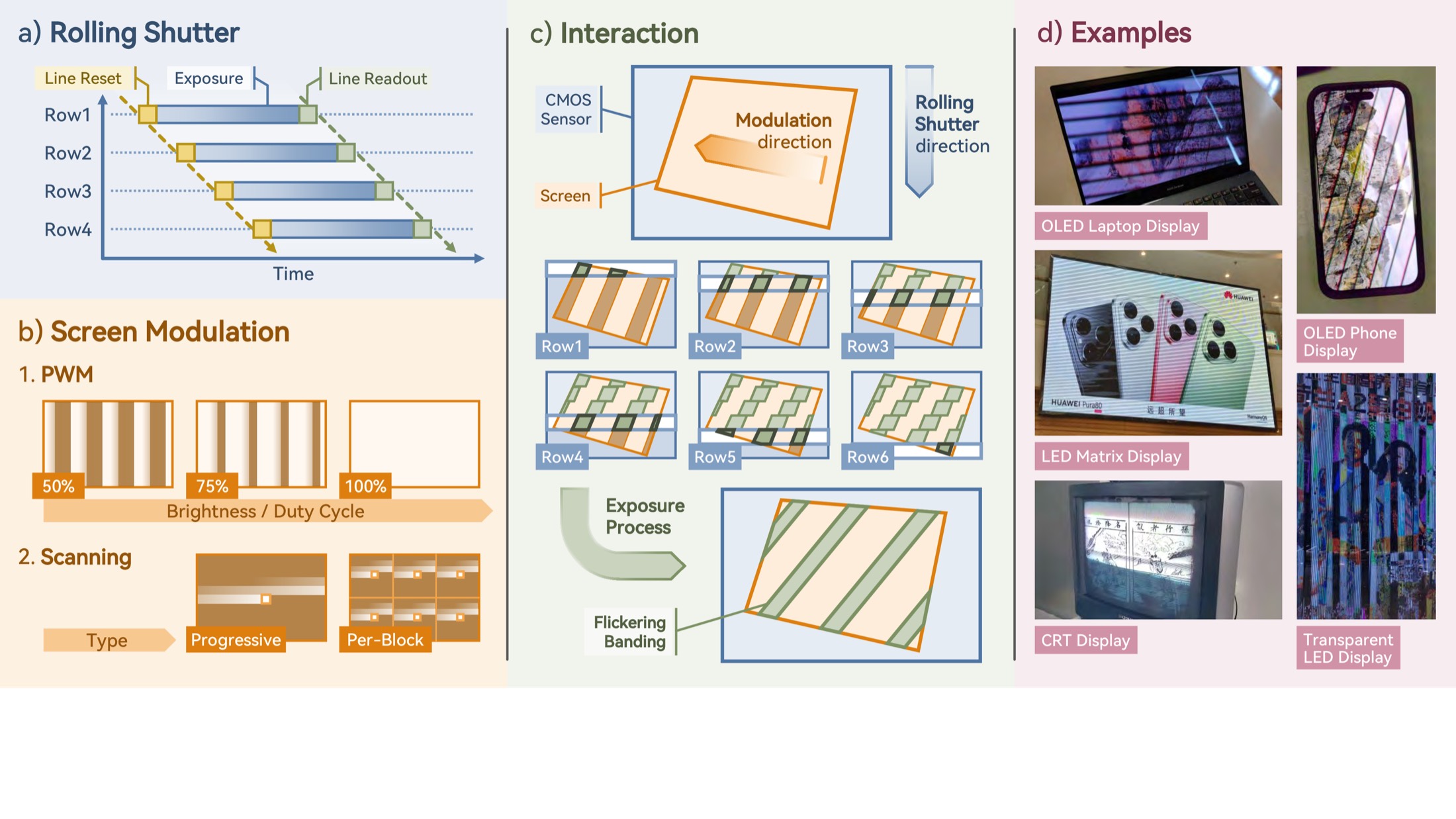}
  \vspace{-13mm}
  \caption{Flicker-banding (FB) when filming screens with smartphone cameras. \textbf{a)} Rolling shutter exposure process. \textbf{b)} Typical display brightness modulation (e.g., PWM and scanning refresh). \textbf{c)} Interaction between the camera exposure and screen modulation leads to the banding artifacts. \textbf{d)} Example banding patterns captured from different modern display technologies.}
  \label{Fig.flickering-banding-explain}
  \vspace{-4mm}
\end{figure*}

\section{Related work}

Researchers have addressed some display-camera artifacts, such as the moir\'eing patterns. Many works~\cite{Sun_2018, yu2022towards,xu2023image,liu2025freqformer} make great progress in constructing the moir\'eing datasets and designing the demoir\'eing models. Other research on the rolling-shutter degradation spans the physics-based and learning-based correction, including joint the rolling-shutter correction and deblurring~\cite{zhong2021towards, Cao2024RSCorr}. Parallel efforts on the flickering artifacts target fluctuations caused by temporal variations in global illumination such as fluorescent-light flicker~\cite{kohler2021rollin, lin2023deflicker} using typical methods like CycleGAN~\cite{zhu2020unpairedimagetoimagetranslationusing}. More recently, BurstDeflicker~\cite{BurstDeflicker_lishenqu} introduces a benchmark for burst deflickering in dynamic scenes under varying ambient illumination. FlickerFormer~\cite{flickerformer} further exploits periodicity and directionality cues across burst frames to improve ambient-light flicker removal.

However, the above flicker-removal studies address ambient-illumination fluctuations rather than the display-camera flicker-banding considered here, which arises from the interaction between rolling-shutter smartphone capture and PWM- or scan-driven screens. Compared with burst deflickering, our setting is more challenging because the banding is stronger, more complex, and not restricted to horizontally aligned stripes, making prior assumptions and temporal cues less effective. To the best of our knowledge, no public paired banding dataset is currently available in either academia or industry for this setting. Our work releases the first dataset for rigorous evaluation and formulates flicker-banding restoration based on diffusion models~\cite{ho2020denoising, xia2023diffir, rombach2022high}.

\section{Preliminaries}

\noindent\textbf{Flicker-Banding (FB).} \label{sec:flicker-banding}
When smartphone cameras capture OLED or LED-matrix displays, periodic bright and dark stripes may appear in the image. These FB artifacts arise from the temporal mismatch between rolling-shutter acquisition and display brightness modulation.

Most smartphone cameras use CMOS sensors with a rolling-shutter readout~\cite{durini2019high}, which exposes and reads image rows sequentially and thus becomes sensitive to time-varying illumination.

OLED displays typically regulate brightness through pulse-width modulation (PWM)~\cite{geffroy2006organic}, while LED matrices often rely on scanning refresh. In both cases, only part of the screen is lit at a given moment, introducing high-frequency temporal fluctuations during image capture.

FB appears when sequential exposure overlaps with the screen's modulated emission~\cite{Sumner2020ImatestFlicker}, as illustrated in Fig.~\ref{Fig.flickering-banding-explain}. This temporal aliasing converts otherwise invisible temporal dynamics into visible stripe patterns in the captured image, whose appearance depends on both camera settings and display modulation. Further details of these display mechanisms are provided in Appendix~\ref{app:display-patterns}.

\noindent\textbf{Common Case: Gray Banding.} Although extreme imaging conditions and exposure parameters can produce colored or curved stripes, everyday photography more often yields straight gray-scale bands, which are the dominant artifact in our dataset and the primary focus of our restoration setting.

\noindent\textbf{Distinction from Moir\'e.} Although both moir\'e and FB arise from display-camera interactions, moir\'e is caused by \textit{spatial} interference between the display lattice and the camera sampling grid, whereas FB results from \textit{temporal} aliasing between the rolling-shutter readout and display modulation. Their different origins make FB a distinct restoration problem.

\begin{figure}[t]
  \centering
  \setlength{\tabcolsep}{2pt}
  \begin{tabular}{cccccc}
    \includegraphics[width=0.155\linewidth]{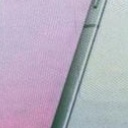} &
    \includegraphics[width=0.155\linewidth]{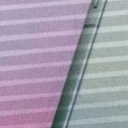} &
    \includegraphics[width=0.155\linewidth]{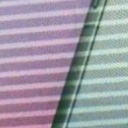} &
    \includegraphics[width=0.155\linewidth]{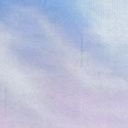} &
    \includegraphics[width=0.155\linewidth]{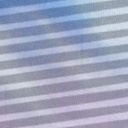} &
    \includegraphics[width=0.155\linewidth]{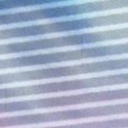} \\
    {\scriptsize GT} & {\scriptsize SIM} & {\scriptsize LQ} & {\scriptsize GT} & {\scriptsize SIM} & {\scriptsize LQ}
  \end{tabular}
  \vspace{-2mm}
  \caption{Visual comparison between our simulated flicker-banding and real-world flicker-banding. GT indicates the real-world non-banding images, on which our simulation pipeline is conducted. SIM indicates our simulation FB images, while LQ indicates real-world FB images.}
  \vspace{-3mm}
  \label{fig:sim-visual-comparison}
\end{figure}

\section{Methods}
\label{sec:methods}
\subsection{Flicker-Banding Dataset Pipeline}
\begin{figure*}[t]
  \centering
  \includegraphics[width=0.99\linewidth]{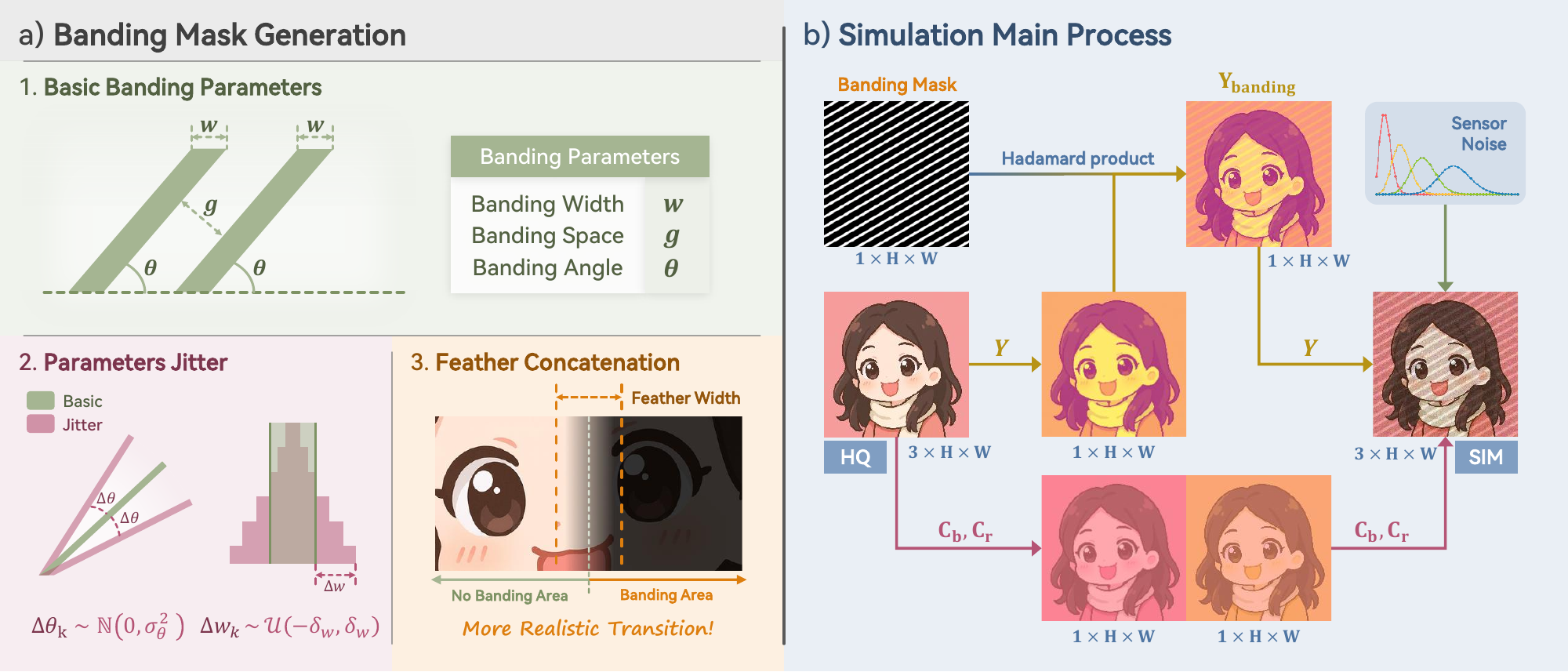}
  \caption{Our flicker-banding simulation pipeline design. \textbf{a)} Stage 1: We generate the general framework based on the basic banding parameters and introduce parameter jitter and feather concatenation for a more realistic transition. \textbf{b)} Stage 2: We overlay the flicker-banding mask on the Y-channel of high-quality (HQ) images and add sensor noise to the reconstructed screen images.}
  \label{fig:simulate}
\end{figure*}

To the best of our knowledge, no existing dataset targets flicker-banding (FB) degradations. We therefore build a simulation pipeline in Fig.~\ref{fig:simulate} for training and a real-world dataset for testing.

\subsubsection{Simulated Flicker-Banding Datasets for Training Datasets} \label{sec:simulated-dataset}
We simulate training degradations because large-scale paired real FB data are scarce.

Let the high-quality (HQ) RGB image be $I_{HQ}\in[0,1]^{3\times H \times W}$ with pixel coordinates $(x,y)$.
We form a stripe-aligned local coordinate $(u,v)$ by rotating pixel coordinates through banding angle $\theta$:
\begin{equation}
\begin{bmatrix}
u\\ v
\end{bmatrix}
=
\begin{bmatrix}
\cos\theta & \sin\theta\\
-\sin\theta & \cos\theta
\end{bmatrix}
\begin{bmatrix}
x-\tfrac{W-1}{2}\\[2pt]
y-\tfrac{H-1}{2}
\end{bmatrix}, \theta\in[-\pi,\pi).
\label{eq:rot}
\end{equation}
Given nominal stripe width $w>0$ and gap $g>0$, we get the spatial period $P = w + g$.

With a normal-direction phase offset $\phi$ (in pixels), the centerline of the $k$-th stripe is
\begin{equation}
L^c_k(u) = kP + \phi,\qquad k\in\mathbb{Z}.
\end{equation}
The basic FB mask ($1$ indicates the banding area, $0$ indicates the non-banding area) is
\begin{equation}
\mathcal{M}(u,v)=
\begin{cases}
1, & \exists\,k:\ \left|v-L^c_k(u)\right|\le \tfrac{w_0}{2},\\[2pt]
0, & \text{otherwise}.
\end{cases}
\end{equation}

To better mimic real FB, we introduce jitter to stripe orientation, spacing, width, and boundaries.

\Needspace{5\baselineskip}
\noindent\textbf{Orientation angle jitter.}
To allow realistic departures from the nominal orientation $\theta$, we model the local orientation of the $k$-th stripe along its tangential coordinate $u$ as
\begin{equation}
\theta_k(u) = \theta_0 + \Delta\theta_k(u),
\label{equ: angle jitter}
\end{equation}
where $\triangle\theta_k(u)\sim \mathcal{N}(0,{\sigma_{\theta}}^2)$ is a zero-mean Gaussian perturbation with the variance $\sigma_\theta^2$.

\noindent\textbf{Spacing and width fluctuation.}
We denote the $k$-th stripe centerline with spacing jitter as
\begin{equation}
L_k^{\text{c}}(u) = kP + \phi + \Delta g_k,\qquad
\Delta_g \sim \mathcal{U}\!\left(-\delta_g,\delta_g\right),
\end{equation}
where $\delta_g$ is the spacing-jitter amplitude.
The banding width jitter with width-jitter amplitude $\delta_w$ is
\begin{equation}
w_k = w + \Delta w_k,\qquad
\Delta_w \sim \mathcal{U}\!\left(-\delta_w,\delta_w\right).
\end{equation}

\noindent\textbf{Axial edge fluctuation.}
The top and bottom edges meander along the stripe axis:
\begin{equation}
\left\{
\begin{aligned}
v_{k}^{\text{top}}(u) &= v_k^{\text{c}}(u) + \tfrac{1}{2}w_k(u) + \delta_{\text{edge}}\,\eta_{\text{top}}(u)\\
v_{k}^{\text{bot}}(u) &= v_k^{\text{c}}(u) - \tfrac{1}{2}w_k(u) + \delta_{\text{edge}}\,\eta_{\text{bot}}(u)
\end{aligned}
\right.,
\label{equ:edge jitter}
\end{equation}
where $\eta_{\text{top}},\eta_{\text{bot}}$ are low-pass 1D random processes and $\delta_{edge}$ sets the normal jitter amplitude.

We convert $I_{HQ}$ from RGB to Y$\text{C}_\text{b}\text{C}_\text{r}$ and isolate the luminance channel:
\begin{equation}
I^Y_{HQ} =  K_RI^R_{HQ} + K_GI^G_{HQ} + K_BI^B_{HQ},
\label{equ:Y_transition}
\end{equation}
where $[K_R,K_B,K_G]=[0.299,0.114,0.587]$, and $I^{C}\in[0,1]^{1\times H \times W}$ denotes channel $C$ of image $I$. Banding mask is then applied only on luminance channel with darkening factor $v_Y\in(0,1]$:
\begin{equation}
I^Y_{LQ}=\underbrace{v_Y I^Y_{HQ} \mathcal{M}}_{Banding~Area} + \underbrace{I^Y_{HQ} (1 - \mathcal{M})}_{Nonbanding~Area}.
\label{equ:banding_add}
\end{equation}
This luminance-only design is motivated by our observation that real FB mainly appears in the Y channel of $\text{YC}_\text{b}\text{C}_\text{r}$, while the chrominance channels remain nearly unchanged. As further shown in Appendix~\ref{app:simulation-analysis}, swapping only the Y channel between banded and banding-free images transfers the artifact almost entirely. Applying the same mask to all RGB channels therefore introduces a noticeable domain gap in real-world screen photography.

Finally, we obtain the simulated FB image $I_{LQ}$ by recombining the processed channels and adding sensor noise to better emulate real short-exposure capture conditions in practice:
\begin{equation}
I_{LQ} = \mathcal{C}(I^Y_{LQ},I^{C_b}_{HQ},I^{C_r}_{HQ}) + \zeta(\mathcal{C}(I^Y_{LQ},I^{C_b}_{HQ},I^{C_r}_{HQ})),
\label{equ:LQ_overlay}
\end{equation}

Here, $\mathcal{C}$ converts Y$\text{C}_\text{b}\text{C}_\text{r}$ to RGB. The noise term $\zeta$ combines Poisson strength $\alpha$ and Gaussian strength $\sigma_r^{2}$ to mimic mixed noise that arises during short-exposure camera capture:
\begin{equation}
\zeta(I) = \sqrt{\alpha I+\sigma_r^{2}}\;\varepsilon,
\qquad \varepsilon\sim\mathcal N(0,1).
\label{equ:sensro_noise}
\end{equation}

\begin{figure*}[t]
  \centering
  \includegraphics[width=0.99\linewidth]{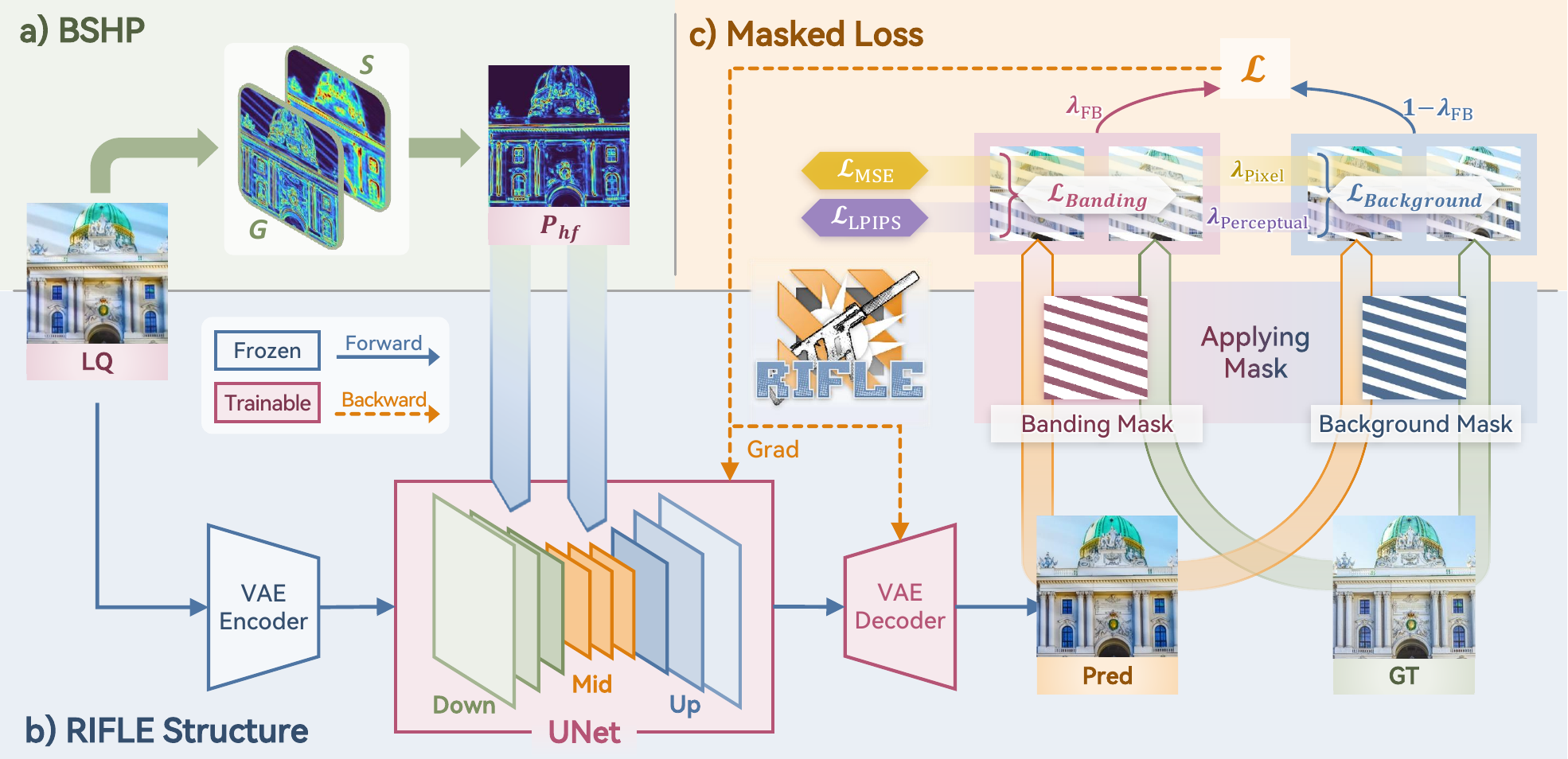}
  \vspace{-1mm}
  \caption{Overview of our model design. \textbf{a)} We propose Banding-Suppressed High-Frequency Prior (\textbf{BSHP}), which combines a fine-grained gradient signal with a smooth structural support map to build a compact single-channel prior. \textbf{b)} We adopt a one-step diffusion restoration model and inject BSHP through multi-stage FiLM modulation for degradation-aware feature recalibration. \textbf{c)} We propose a masked loss (\textbf{ML}) that emphasizes banded regions while preserving global fidelity.}
  \label{fig:structure}
  \vspace{-3mm}
\end{figure*}

We also apply feathered blending at stripe boundaries for smoother transitions. Fig.~\ref{fig:sim-visual-comparison} shows that the simulated FB is visually close to real FB, supporting the validity of the training data.

\subsubsection{Real-World Flicker-Banding Datasets for Testing Datasets}
Although simulation provides abundant training samples, objective evaluation requires a real-world benchmark. We collect 105 valid high-resolution pairs from five scenes on commercial LED/OLED-type emissive advertising displays, using a vivo X200s under indoor ambient lighting. Each pair is captured from a fixed viewpoint: a short exposure produces the banded LQ observation, while a longer exposure produces the banding-free GT reference, with ambient lighting unchanged during paired capture. The scene-specific exposure and ISO settings are reported in Appendix~\ref{app:acquisition}.

We register each LQ image to its GT reference using a homography estimated from matched image features, with ECC refinement when feature-based alignment is insufficient. We crop the common valid field of view and visually discard pairs with registration failures or invalid boundaries. Paired patches are then extracted at identical coordinates; a screen mask selects locations containing sufficient display content without modifying or replacing any LQ or GT pixels. The dataset comprises 105 image pairs at $4096\times3072$ resolution, from which we obtain 424 paired $512\times512$ test patches.

Our intended scope is dark, predominantly luminance, approximately straight FB with arbitrary orientations and varying width, spacing, and strength. The acquisition conditions instantiate this predefined target. Chromatic, strongly curved or nonstationary, multi-frequency or per-block patterns, and severe cases with unrecoverable information loss remain beyond the current capability. We do not claim coverage of CRTs, projectors, fluorescent lamps, ambient LED light sources, or other flicker sources.

\subsection{Design of Flicker-Banding Removing Model}
Although FB is common in screen photography, effective solutions remain limited. Hardware-side correction is difficult because of diverse screen scanning mechanisms and material properties. We therefore adopt the restoration model in Fig.~\ref{fig:structure} and use the one-step diffusion image restoration model PiSA-SR~\cite{sun2024pisasr} as the baseline for its strong efficiency and performance.

The contribution of BSHP and ML lies in their task-specific integration into the complete FB-removal pipeline rather than in FiLM or masked losses themselves. BSHP fuses gradient-based localization with repeatability-based structural support and injects this FB-specific prior through multi-stage FiLM. ML uses the banding mask produced by the same simulation process, coupling degradation synthesis with region-focused supervision for banded image regions.

\subsubsection{Loss Design of Removing Model}

Let the low-quality (LQ), prediction, and ground truth (GT) images be $x_{\text{LQ}}$, $x_{\text{Pred}}$, and $x_{\text{GT}} \in \mathbb{R}^{B\times C\times H\times W}$, and let the simulation mask be $\mathcal{M}\in\{0,1\}^{B\times 1\times H\times W}$, shared with the simulated FB generation process, where $1$ indicates banding and $0$ indicates background. Since the background of $x_{\text{LQ}}$ is typically close to that of $x_{\text{GT}}$, we emphasize reconstruction in the banding region using $\lambda_{\text{FB}}\in[0,1]$, with $\lambda_{\text{Pixel}}, \lambda_{\text{Perceptual}}\ge 0$ weighting the loss terms.

\noindent\textbf{Area-decoupled masked mean operator.}
Let $I\in\mathbb{R}^{B\times C\times H\times W}$ be an image tensor and $\mathcal{M}\in\{0,1\}^{B\times 1\times H\times W}$ the binary banding mask. Let $\tilde{\mathcal{M}}$ be $\mathcal{M}$ broadcast to the shape of $I$. We define
\begin{equation}
  \MMdb{I}{\mathcal{M}} =
  \frac{\sum I \odot \tilde{\mathcal{M}}}{\sum_{b,c,h,w}\mathbf{1} \odot \tilde{\mathcal{M}} + \varepsilon},
  \label{equ:masked mean}
\end{equation}
where $\varepsilon>0$ ensures numerical stability for the area-normalized pixel loss in each region.

\noindent\textbf{Masked Pixel Loss.} We use mean squared error (MSE) for pixel-level reconstruction:
\begin{equation}
    \mathcal{L}_{\text{Pixel}} = ( x_{\text{Pred}} - x_{\text{GT}} )^2,
    \label{equ:l2loss}
\end{equation}
To emphasize restoration in banding regions, we apply $\mathcal{M}$ to obtain the masked MSE loss:
\begin{equation}
    \mathcal{L}_{\text{Pixel}}^{\text{Masked}} = \lambda_{\text{FB}}\MMdb{\mathcal{L}_{\text{Pixel}}}{\mathcal{M}}
    \quad + (1 - \lambda_{\text{FB}})\MMdb{\mathcal{L}_{\text{Pixel}}}{\mathbf{1}-\mathcal{M}}.
    \label{equ:masked_l2loss}
\end{equation}

\noindent\textbf{Masked Perceptual Loss.} We use LPIPS~\cite{zhang2018unreasonable} to compare masked image pairs. Let $\Phi$ denote the scalar LPIPS distance, including its required input normalization. For $\mathcal R\in\{\mathcal M,\mathbf1-\mathcal M\}$, we apply the same mask to both input images before computing the LPIPS distance:
\begin{equation}
d_{\Phi}(\mathcal R)=\frac{1}{B}\sum_{b=1}^{B}
\Phi\!\left(\widetilde{\mathcal R}_{b}\odot x_{\text{Pred},b},
\widetilde{\mathcal R}_{b}\odot x_{\text{GT},b}\right).
\label{equ:masked_lpips_distance}
\vspace{-2mm}
\end{equation}

Here, $\widetilde{\mathcal R}_{b}$ broadcasts the mask across channels; both inputs are zero outside the selected region. We combine the two batch-averaged scalar distances using the banding weight:
\begin{equation}
    \mathcal{L}_{\text{Perceptual}}^{\text{Masked}} = \lambda_{\text{FB}}d_{\Phi}(\mathcal M)
    + (1 - \lambda_{\text{FB}})d_{\Phi}(\mathbf1-\mathcal M).
    \label{equ:masked_lpipsloss}
\end{equation}

\noindent\textbf{Merged Masked Loss.} The final loss balances pixel fidelity and overall perceptual quality:
\begin{equation}
\mathcal{L} = \lambda_{\text{Pixel}} \mathcal{L}_{\text{Pixel}}^{\text{Masked}} + \lambda_{\text{Perceptual}} \mathcal{L}_{\text{Perceptual}}^{\text{Masked}}.
    \label{equ:merged_loss}
\end{equation}

\begin{figure*}[t]
  \centering
  \setlength{\tabcolsep}{2pt}
  \begin{tabular}{cccc}
    \includegraphics[width=0.235\linewidth]{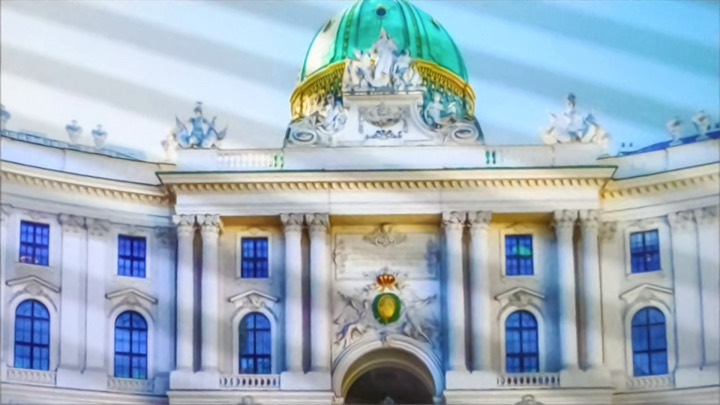} &
    \includegraphics[width=0.235\linewidth]{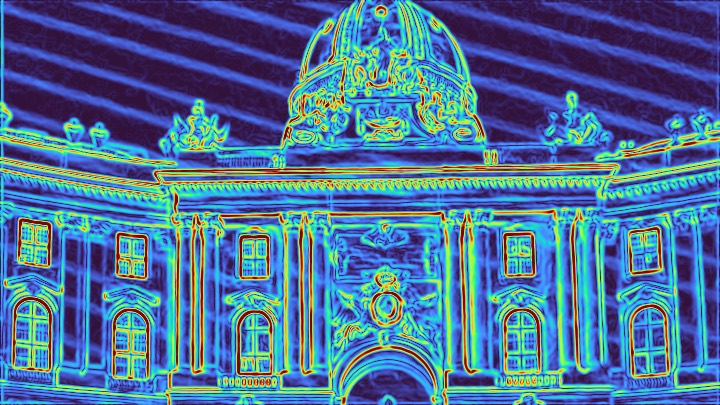} &
    \includegraphics[width=0.235\linewidth]{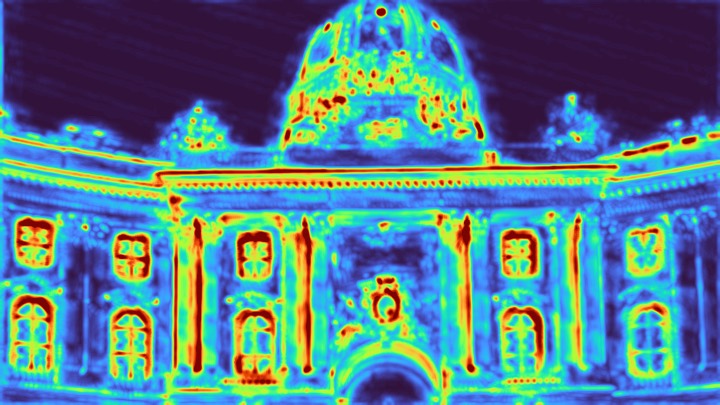} &
    \includegraphics[width=0.235\linewidth]{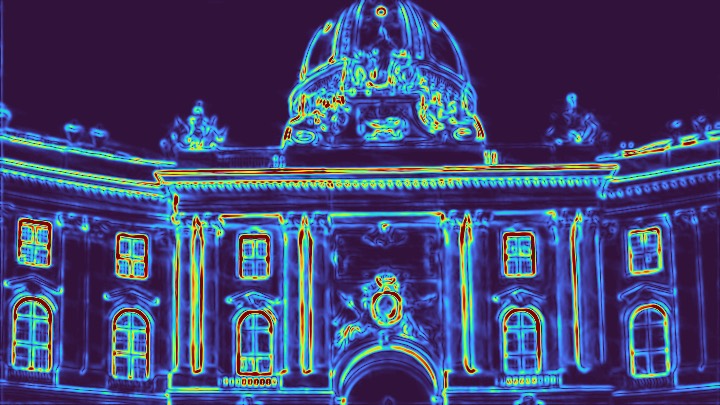} \\
    \scriptsize LQ & \scriptsize Gradient signal $G$ & \scriptsize Repeatability map $S$ & \scriptsize Banding cue $P_{\mathrm{hf}}$
  \end{tabular}
  \vspace{-1.5mm}
  \caption{Visualization of BSHP. Gradient signal $G$ captures fine transitions but also stripe responses, while repeatability map $S$ is smoother and provides structural support. Their fusion yields a compact prior $P_{\mathrm{hf}}$ that preserves informative transitions while suppressing unstable banding responses.}
  \vspace{-2mm}
  \label{fig:bshp-visualization}
\end{figure*}

\subsubsection{Banding-Suppressed High-Frequency Prior (BSHP)}

Inspired by conditional restoration, we cast FB removal as a prior-guided restoration problem. Rather than explicitly estimating a stripe mask, we seek a compact prior that highlights recoverable high-frequency structures while suppressing unstable artifact responses.

For an input image $x_{\text{LQ}}$, we first convert its RGB channels to luminance for prior construction:
\begin{equation}
Y=K_R R+K_G G+ K_BB.
\label{equ:fpe_luma}
\end{equation}

\Needspace{6\baselineskip}
To characterize informative local high-frequency structures in the degraded input, we apply mild Gaussian smoothing before computing the spatially localized gradient signal:
\begin{equation}
\tilde Y=\mathcal{G}_{\sigma_b}(Y),\qquad
G=\mathrm{Norm}\!\left(\sqrt{|\partial_x \tilde Y|^2+|\partial_y \tilde Y|^2}\right).
\label{equ:fpe_gradient}
\end{equation}
Here, $G$ localizes candidate high-frequency structures, but also responds to FB-induced stripe edges.

However, transition strength alone is insufficient for robust prior construction. We therefore introduce a second term that provides smoother and more reliable local structural support:
\begin{equation}
R_{\mathrm{mid}}=
\alpha \bigl(Y-\mathcal{G}_{\sigma_1}(Y)\bigr)
+(1-\alpha)\bigl(Y-\mathcal{G}_{\sigma_2}(Y)\bigr),\qquad
X=\mathrm{Norm}\!\left(|R_{\mathrm{mid}}|\right),
\label{equ:fpe_mid}
\end{equation}
where $\sigma_1<\sigma_2$ and $\alpha\in[0,1]$. We then define a repeatability map by searching over candidate directions $\Theta$ and spatial offsets $\Delta$ across the local neighborhood of each pixel:
\begin{equation}
S(x,y)=
\mathrm{Norm}\!\left(
\max_{\delta\in\Delta,\ \theta\in\Theta}
\mathcal{G}_{\sigma_r}\!\left(
X(x,y)\cdot X\bigl((x,y)+\delta u_\theta\bigr)
\right)
\right),
\label{equ:fpe_repeat}
\end{equation}
where $u_\theta$ is the unit vector of direction $\theta$. We use $\Theta=\{0^\circ,15^\circ,\ldots,165^\circ\}$, $\Delta=\{2,4,\ldots,20\}$ pixels, and $\sigma_r=2.0$. Here, $S$ is a local repeatability score, not an estimator of stripe frequency or angle: it measures the strength of similar luminance responses within the local neighborhood. Spatial band frequency and angle affect the direction-offset responses before aggregation, but the maximum operation retains repetition strength rather than a particular direction or offset. Thus, $\theta$ and $\delta$ cannot be compared one-to-one with the angle, period, phase, width, or mask used in simulation. The simulation parameters generate training degradations, whereas $S$ is computed directly from each LQ image. When $S$ reliably supports relevant structures, its fusion with the localized gradient cue $G$ provides more informative guidance for FB restoration in each image region.

Neither cue alone is sufficient: $G$ is rich in fine details but also contains stripe responses, while $S$ is smoother and cleaner but lacks precise localization. We therefore fuse these two complementary cues multiplicatively at each pixel and define the FB-specific final prior as
\begin{equation}
P_{\mathrm{hf}}=
\mathrm{Norm}\!\left(
G^{\gamma_g}\cdot S^{\gamma_s}
\right),
\label{equ:fpe_final_prior}
\end{equation}
where $\gamma_g$ and $\gamma_s$ balance transition strength and structural support. The gradient term preserves precise localization, while the support term reweights these responses with smoother local context. As shown in Fig.~\ref{fig:bshp-visualization}, $P_{\mathrm{hf}}$ is a single-channel prior that retains the precise localization ability of $G$ while suppressing unstable stripe-dominated responses through the structural support map $S$.

Rather than concatenating the prior to the input image, we inject it into the restoration backbone through FiLM-based feature modulation. We first encode the prior as
\begin{equation}
M=\mathrm{SiLU}\!\left(\mathrm{Conv}_{1\times1}(P_{\mathrm{hf}})\right),
\end{equation}
\begin{equation}
z=\left[\mathrm{Pool}_g(M);\mathrm{Pool}_h(M);\mathrm{Pool}_v(M)\right],
h=\mathrm{SiLU}\!\left(W_s z+b_s\right),
\label{equ:prior_embedding}
\end{equation}
and, for each injected U-Net stage $l$, generate channel-wise modulation parameters:
\begin{equation}
[\gamma_l,\beta_l]=W_l h+b_l,\qquad
\hat F_l=F_l\odot (1+\gamma_l)+\beta_l.
\label{equ:film_mod}
\end{equation}
Here, $F_l$ denotes the intermediate feature tensor at stage $l$, while $\mathrm{Pool}_g$, $\mathrm{Pool}_h$, and $\mathrm{Pool}_v$ denote global, horizontal, and vertical average pooling. This directional FiLM conditioning biases the backbone toward structurally informative high-frequency transitions with minimal overhead.

\section{Experiments}
\label{sec:experiments}

\begin{table*}[t]
\centering
\scriptsize
\caption{Quantitative comparison on the real-world FB testing dataset. Except for Notch Filter and FlickerFormer*, all learning-based baselines are trained or finetuned on our simulated data. FlickerFormer* uses the official checkpoint from the original deflickering task, while FlickerFormer denotes the version finetuned on our simulated data. Best results are highlighted in \textcolor{red}{red}.}
\vspace{-2mm}
\resizebox{\textwidth}{!}{
\begin{tabular}{c | c | c c c c c c}
\toprule
\rowcolor{iclrcolor!30}
{Methods} & Task & SSIM$\uparrow$ & ms-SSIM$\uparrow$ & LPIPS$\downarrow$ & DISTS$\downarrow$ & FSIM$\uparrow$ & GMSD$\downarrow$ \\[-0.75mm]
\midrule
LQ & $-$ & 0.6239 & 0.6758 & 0.3070 & 0.1978 & 0.7947 & 0.2045 \\[0.25mm]
\cline{1-8}
\rule{0pt}{0.5mm}\\[-2mm]
Notch Filter & Filtering & 0.5562 & 0.6313 & 0.3411 & 0.2600 & 0.7196 & 0.2110 \\
MAT~\cite{xie2025mat} & Super-resolution & 0.4747 & 0.6479 & 0.4114 & 0.2243 & 0.7766 & 0.2003 \\
InvSR~\cite{yue2025InvSR} & Super-resolution & 0.6101 & 0.6873 & 0.3833 & 0.2195 & 0.8035 & 0.2108 \\
Step1X~\cite{liu2025step1x-edit} & Image Editing & 0.6164 & 0.6715 & 0.3564 & 0.2066 & 0.7900 & 0.2079 \\
ESDNet~\cite{yu2022towards} & Demoir\'eing & 0.6306 & 0.6941 & 0.2984 & 0.1905 & 0.8107 & 0.1969 \\
FlickerFormer*~\cite{flickerformer} & Deflickering & 0.6252 & 0.6783 & 0.3054 & 0.1968 & 0.7977 & 0.2031 \\
FlickerFormer~\cite{flickerformer} & Deflickering & 0.6658 & 0.7993 & 0.2433 & 0.1534 & 0.9086 & 0.1528 \\
PiSA-SR~\cite{sun2024pisasr} & Super-resolution & 0.6201 & 0.7370 & 0.3808 & 0.2048 & 0.8443 & 0.2085\\
\rowcolor{iclrcolor!10}
RIFLE (ours) & Debanding & \textcolor{red}{0.6843} & \textcolor{red}{0.8318} & \textcolor{red}{0.2209} & \textcolor{red}{0.1410} & \textcolor{red}{0.9281} & \textcolor{red}{0.1376}\\
\cline{1-8}
\bottomrule
\end{tabular}
}
\vspace{-2mm}
\label{tab: Real-World Quantitative Results}
\end{table*}

\begin{figure*}[t]
  \scriptsize
  \centering
  \begin{tabular}{@{}cc@{}}
    \hspace{-0.35cm}
    \begin{adjustbox}{valign=t}
    \begin{tabular}{c}
      \includegraphics[width=0.26\textwidth]{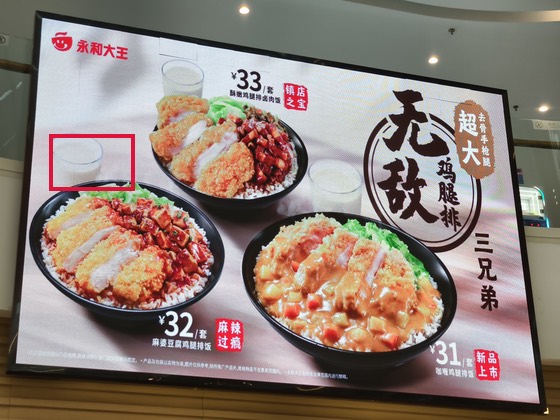}\\
      Real-Flicker: Banding09\_1
    \end{tabular}
    \end{adjustbox}
    \hspace{-0.30cm}
    &
    \begin{adjustbox}{valign=t}
    \begin{tabular}{ccccc}
      \includegraphics[width=0.13\textwidth]{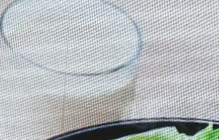}\hspace{-3mm} &
      \includegraphics[width=0.13\textwidth]{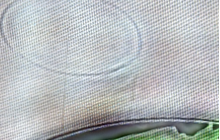}\hspace{-3mm} &
      \includegraphics[width=0.13\textwidth]{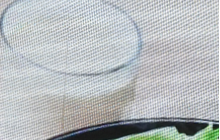}\hspace{-3mm} &
      \includegraphics[width=0.13\textwidth]{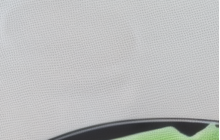}\hspace{-3mm} &
      \includegraphics[width=0.13\textwidth]{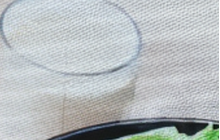}\hspace{-3mm} \\
      LQ \hspace{-3mm} & Notch Filter \hspace{-3mm} & MAT \hspace{-3mm} & InvSR \hspace{-3mm} & Step1X \hspace{-3mm} \\
      \includegraphics[width=0.13\textwidth]{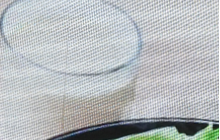}\hspace{-3mm} &
      \includegraphics[width=0.13\textwidth]{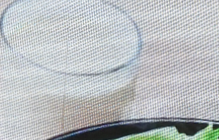}\hspace{-3mm} &
      \includegraphics[width=0.13\textwidth]{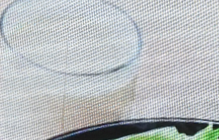}\hspace{-3mm} &
      \includegraphics[width=0.13\textwidth]{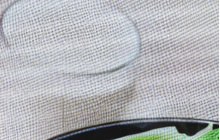}\hspace{-3mm} &
      \includegraphics[width=0.13\textwidth]{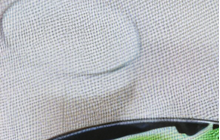}\hspace{-3mm} \\
      ESDNet \hspace{-3mm} & FlickerFormer* \hspace{-3mm} & FlickerFormer \hspace{-3mm} & PiSA-SR \hspace{-3mm} & RIFLE \hspace{-3mm}
    \end{tabular}
    \end{adjustbox}
  \end{tabular}
  \vspace{-0.8mm}

  \begin{tabular}{@{}cc@{}}
    \hspace{-0.35cm}
    \begin{adjustbox}{valign=t}
    \begin{tabular}{c}
      \includegraphics[width=0.26\textwidth]{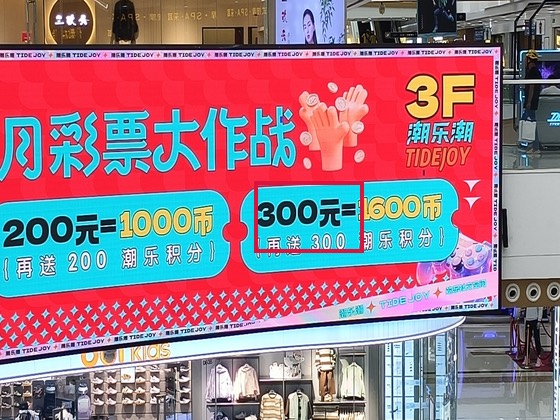}\\
      Real-Flicker: Banding36\_1
    \end{tabular}
    \end{adjustbox}
    \hspace{-0.30cm}
    &
    \begin{adjustbox}{valign=t}
    \begin{tabular}{ccccc}
      \includegraphics[width=0.13\textwidth]{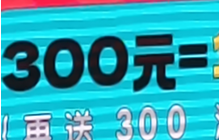}\hspace{-3mm} &
      \includegraphics[width=0.13\textwidth]{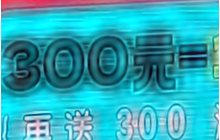}\hspace{-3mm} &
      \includegraphics[width=0.13\textwidth]{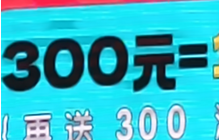}\hspace{-3mm} &
      \includegraphics[width=0.13\textwidth]{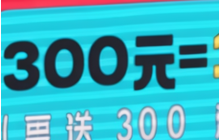}\hspace{-3mm} &
      \includegraphics[width=0.13\textwidth]{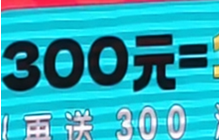}\hspace{-3mm} \\
      LQ \hspace{-3mm} & Notch Filter \hspace{-3mm} & MAT \hspace{-3mm} & InvSR \hspace{-3mm} & Step1X \hspace{-3mm} \\
      \includegraphics[width=0.13\textwidth]{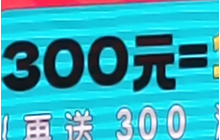}\hspace{-3mm} &
      \includegraphics[width=0.13\textwidth]{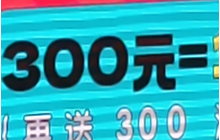}\hspace{-3mm} &
      \includegraphics[width=0.13\textwidth]{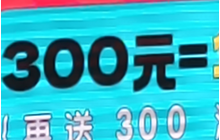}\hspace{-3mm} &
      \includegraphics[width=0.13\textwidth]{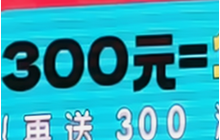}\hspace{-3mm} &
      \includegraphics[width=0.13\textwidth]{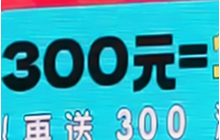}\hspace{-3mm} \\
      ESDNet \hspace{-3mm} & FlickerFormer* \hspace{-3mm} & FlickerFormer \hspace{-3mm} & PiSA-SR \hspace{-3mm} & RIFLE \hspace{-3mm}
    \end{tabular}
    \end{adjustbox}
  \end{tabular}
  \vspace{-0.8mm}

  \begin{tabular}{@{}cc@{}}
    \hspace{-0.35cm}
    \begin{adjustbox}{valign=t}
    \begin{tabular}{c}
      \includegraphics[width=0.26\textwidth]{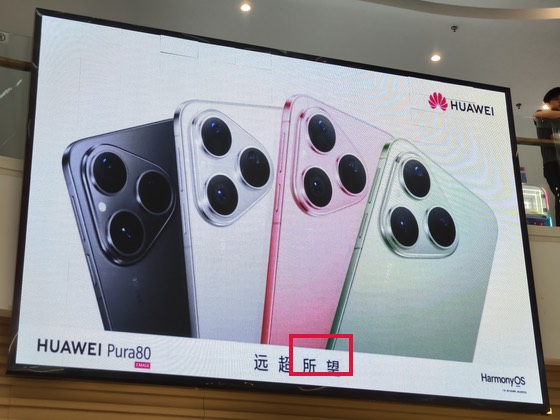}\\
      Real-Flicker: Banding01\_2
    \end{tabular}
    \end{adjustbox}
    \hspace{-0.30cm}
    &
    \begin{adjustbox}{valign=t}
    \begin{tabular}{ccccc}
      \includegraphics[width=0.13\textwidth]{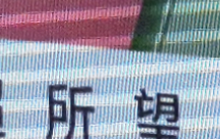}\hspace{-3mm} &
      \includegraphics[width=0.13\textwidth]{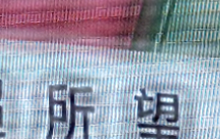}\hspace{-3mm} &
      \includegraphics[width=0.13\textwidth]{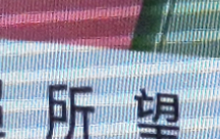}\hspace{-3mm} &
      \includegraphics[width=0.13\textwidth]{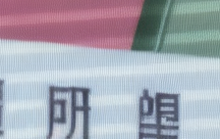}\hspace{-3mm} &
      \includegraphics[width=0.13\textwidth]{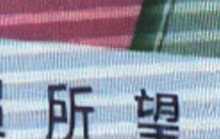}\hspace{-3mm} \\
      LQ \hspace{-3mm} & Notch Filter \hspace{-3mm} & MAT \hspace{-3mm} & InvSR \hspace{-3mm} & Step1X \hspace{-3mm} \\
      \includegraphics[width=0.13\textwidth]{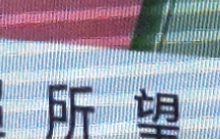}\hspace{-3mm} &
      \includegraphics[width=0.13\textwidth]{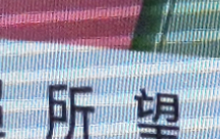}\hspace{-3mm} &
      \includegraphics[width=0.13\textwidth]{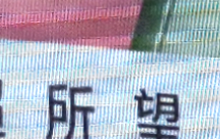}\hspace{-3mm} &
      \includegraphics[width=0.13\textwidth]{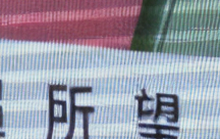}\hspace{-3mm} &
      \includegraphics[width=0.13\textwidth]{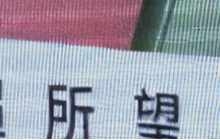}\hspace{-3mm} \\
      ESDNet \hspace{-3mm} & FlickerFormer* \hspace{-3mm} & FlickerFormer \hspace{-3mm} & PiSA-SR \hspace{-3mm} & RIFLE \hspace{-3mm}
    \end{tabular}
    \end{adjustbox}
  \end{tabular}
  \vspace{-4pt}
  \caption{Visual comparison on three representative Real-Flicker cases. RIFLE removes stronger flicker-banding artifacts with cleaner structures and fewer residual stripes than competing baselines.}
  \vspace{-4pt}
  \label{fig:main-visual-comparison}
\end{figure*}

\subsection{Experiments Setup}
\label{exp_all}
\noindent\textbf{Data Construction.} For training, we apply our simulation pipeline to LSDIR~\cite{li2023lsdir} and UHDM~\cite{yu2022towards}. LSDIR is a large-scale super-resolution dataset, while UHDM is a demoir\'eing dataset; both provide abundant high-quality images for synthesizing flicker-banding degradations. We use UHDM to better cover screen-shooting scenarios and LSDIR to improve generalization to more diverse natural images.
For the testing datasets, we use our proposed real-world datasets, consisting of 424 paired patches of 512 $\times$ 512 size. Results from real-world datasets better demonstrate the model's practical value.

\begingroup
\setlength{\parskip}{3pt}
\setlength{\textfloatsep}{10pt plus 2pt minus 2pt}
\setlength{\floatsep}{8pt plus 2pt minus 2pt}
\noindent\textbf{Evaluation Metrics.} No metric directly quantifies FB severity. Since bands affect limited areas while preserving detail elsewhere, global scores can miss visible improvements. We therefore complement a user study with SSIM~\cite{wang2004image}, LPIPS~\cite{zhang2018unreasonable}, DISTS~\cite{ding2020image}, FSIM~\cite{zhang2011fsim}, and GMSD~\cite{Xue2014Gradient}. We omit non-reference metrics because they score banded and banding-free images similarly.

\begin{figure}[t]
\centering
\setlength{\tabcolsep}{1pt}
\renewcommand{\arraystretch}{1.0}
\begin{tabular}{@{}ccc@{}}
  \includegraphics[width=0.3\linewidth]{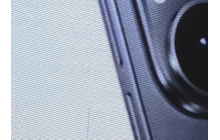} &
  \includegraphics[width=0.3\linewidth]{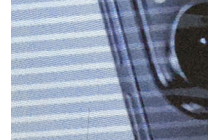} &
  \includegraphics[width=0.3\linewidth]{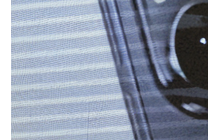} \\
  \footnotesize GT & \footnotesize LQ & \footnotesize Baseline \\
  \includegraphics[width=0.3\linewidth]{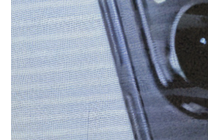} &
  \includegraphics[width=0.3\linewidth]{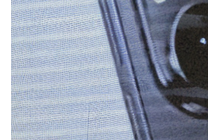} &
  \includegraphics[width=0.3\linewidth]{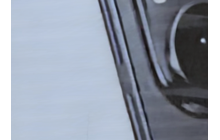} \\
  \footnotesize ML & \footnotesize BSHP & \footnotesize ML+BSHP \\
\end{tabular}
\vspace{-2mm}
\caption{Visual comparison of RIFLE component ablations on a real-world FB image.}
\vspace{-3mm}
\label{fig:ablation-visual-comparison}
\end{figure}

\noindent\textbf{Implementation Details.}
We finetune rank-32 LoRA for 50,000 iterations on paired $512\times512$ images (batch size 4, learning rate $5\times10^{-5}$), using about 43.3 GB of GPU memory and 25.8 hours. The masked-loss weights are $\lambda_{\text{FB}}=0.8$, $\lambda_{\text{Pixel}}=1.0$, and $\lambda_{\text{Perceptual}}=2.0$.

\noindent\textbf{Compared Methods.}
With no prior FB-specific model available, we finetune related baselines on our simulated data: MAT~\cite{xie2025mat} and Step1X~\cite{liu2025step1x-edit} (transformer/image editing), InvSR~\cite{yue2025InvSR} and PiSA-SR~\cite{sun2024pisasr} (diffusion), ESDNet~\cite{yang2025dsdnetrawdomaindemoireing} (demoir\'eing), and FlickerFormer~\cite{flickerformer} (deflickering).

\subsection{Main Results}

\noindent\textbf{Quantitative Results.} Tab.~\ref{tab: Real-World Quantitative Results} reports some quantitative results on our real-world dataset. RIFLE outperforms other competing methods on most metrics. Notably, even raw FB inputs achieve relatively high scores, as reference-based metrics are largely insensitive to thr banding artifacts---FB entails minimal fine-detail loss and non-banding regions closely resemble the corresponding GT reference.

\noindent\textbf{Visual Comparison.} Fig.~\ref{fig:main-visual-comparison} compares real-world results. Competing methods leave visible FB residues despite finetuning on our simulated data. FlickerFormer mainly targets horizontal stripes in its official model; finetuning helps, but stripes and content distortion remain. RIFLE nearly removes mild banding with high fidelity and removes most severe banding with only minor residuals in the output, while other compared methods improve little.



\noindent\textbf{Computational Cost and Additional Evaluation.} At $512\times512$ resolution on an RTX A6000, RIFLE adds 1.239 M parameters, 0.0013 GiB peak inference memory, and 38.2 ms latency (13.7\%) relative to PiSA-SR, with nearly unchanged profiled MACs/FLOPs. Appendix~\ref{app:compute} reports training and inference costs. The user-study protocol and statistics are detailed in Appendix~\ref{app:user-study}; OCR-based text fidelity and full-dataset video evaluation are reported in Appendices~\ref{app:ocr-fidelity} and~\ref{app:video-evaluation}.

\subsection{Ablation Study}

\begin{table}[t]
\centering
\scriptsize
\caption{Ablation study of our proposed components on real-world FB datasets.}
\resizebox{0.9\columnwidth}{!}{
\begin{tabular}{ c | c c c c c c}
\toprule
\rowcolor{iclrcolor!30}
{Methods} & SSIM$\uparrow$ & ms-SSIM$\uparrow$ & LPIPS$\downarrow$ & DISTS$\downarrow$ & FSIM$\uparrow$ & GMSD$\downarrow$ \\[-0.75mm]
\midrule
\rule{0pt}{0.5mm}\\[-3mm]
Baseline & 0.6201 & 0.7370 & 0.3808 & 0.2048 & 0.8443 & 0.2085 \\
ML & 0.6673 & 0.7899 & 0.2520 & 0.1621 & 0.8884 & 0.1513 \\
BSHP & 0.6668 & 0.7928 & 0.2521 & 0.1587 & 0.8920 & 0.1528 \\
\rowcolor{iclrcolor!10}
ML+BSHP & \textcolor{red}{0.6843} & \textcolor{red}{0.8318} & \textcolor{red}{0.2209} & \textcolor{red}{0.1410} & \textcolor{red}{0.9281} & \textcolor{red}{0.1376}\\
\cline{1-7}
\bottomrule
\end{tabular}
}
\vspace{-1mm}
\label{tab:Ablation Quantitative Results}
\end{table}

Tab.~\ref{tab:Ablation Quantitative Results} and Fig.~\ref{fig:ablation-visual-comparison} ablate masked loss (ML) and banding-suppressed high-frequency prior (BSHP). The baseline leaves stripes; ML targets banded regions for cleaner, more faithful outputs. BSHP adds localized high-frequency guidance with smooth structural support, improving removal but sometimes softening texture alone. Together, they best remove bands while preserving edges and detail across all evaluated metrics in our study, especially under heavier FB degradation.

\section{Conclusion}
\vspace{-2mm}
We study flicker-banding (FB) in display photographs and introduce RIFLE with a simulation pipeline and paired real-world benchmark. Its compact prior, FiLM guidance, and region-focused masked loss remove bands while preserving detail. RIFLE outperforms recent baselines quantitatively and visually; ablations confirm both components. To our knowledge, this is the first neural study of FB removal with practical training-data synthesis and paired real-world evaluation data construction. Dark luminance bands in LED/OLED captures define the evaluated scope; chromatic, curved, and extreme information-loss patterns remain challenging. OCR results further show that character-level fidelity is not guaranteed. These limitations motivate richer simulations and text-preserving restoration.

\section*{Acknowledgments}
\vspace{-2mm}
\parfillskip=0pt plus 1fil
This work was supported by Huawei Explore X funds. This work was also sponsored by Al Hundred Schools Program and was carried out using the Ascend AI technology stack.

\newpage
\endgroup

\interlinepenalty=0
\parfillskip=0pt plus 1fil

\appendix
\newpage

\section{Technical appendices and supplementary material}
\label{sec:appendix}
\subsection{Flicker-Banding Patterns across Display Technologies}
\label{app:display-patterns}

\begin{figure*}[h]
\centering
\includegraphics[width=0.99\textwidth]{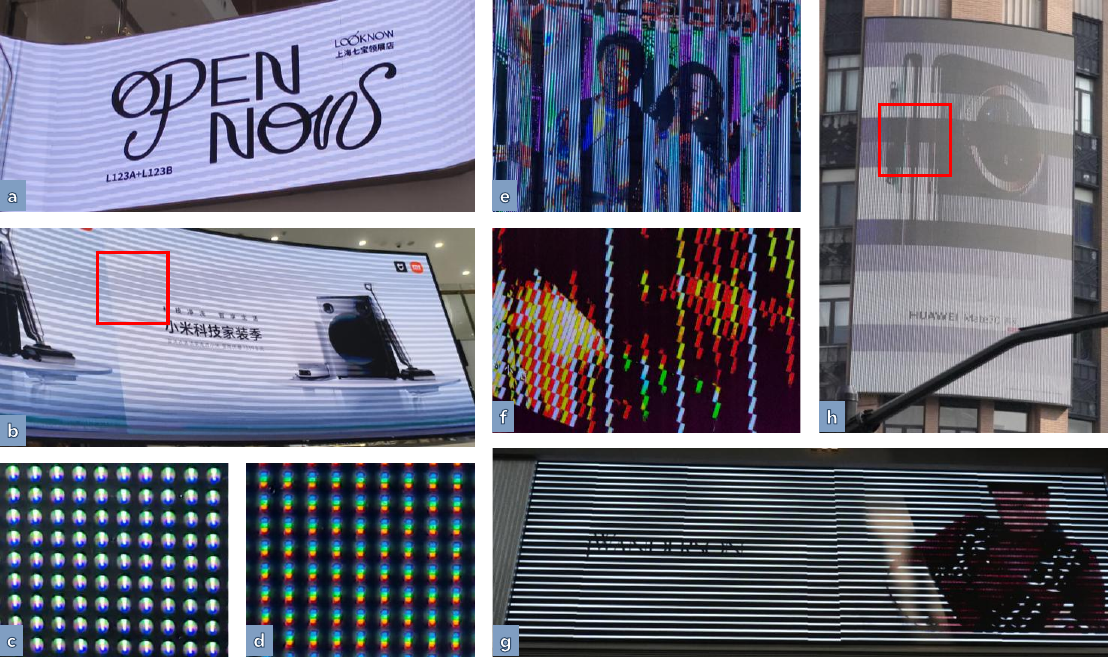}
\caption{Flicker-banding patterns on LED matrix displays with different scanning refresh mechanisms captured under various conditions. \textbf{b)} Thickening of dark stripes, likely due to changes in PWM duty cycle. \textbf{c-d)} Different LED subpixel arrangements. \textbf{e-f)} Atypical banding patterns. \textbf{h)} Complex banding patterns caused by per-block scanning refresh.}
\label{Fig.pattern-led}
\end{figure*}

This subsection surveys LED, OLED, CRT, and projector mechanisms to explain the broader origins of screen-capture banding; it does not imply that RIFLE covers all four technologies. RIFLE is designed and validated only for dark FB in OLED and LED display captures. The surveyed chromatic, complex, per-block, and other atypical patterns illustrate mechanisms beyond the validated scope. FB from CRTs, projectors, fluorescent lamps, ambient LED light sources, and other scenarios is outside the current scope. We distinguish this mechanism survey from the real-world benchmark acquisition and supported FB types described in Sec.~\ref{sec:methods} and Appendix~\ref{app:acquisition}.

\subsubsection{LED Matrix Displays}

LED matrix displays often use a scanning refresh scheme, where rows or columns of pixels are activated in succession. The matrix is often refreshed in blocks or chunks of rows, leading to more complex banding patterns when filmed. PWM is also commonly used in LED matrices, further contributing to more complex banding effects.

\subsubsection{OLED Displays}

OLED displays typically use a combination of PWM and current modulation to achieve smooth brightness control \citep{geffroy2006organic}. In most cases, the duty cycle is same across the entire panel at a given brightness level, resulting in uniform banding patterns when filmed with a camera. However, some OLED panels may reduce the duty cycle in very dark scenes to enhance contrast, leading to non-uniform banding artifacts. Color shifts may also occur on edges of banding stripes due to phase differences in RGB subpixel driving.

\subsubsection{CRT Displays}

In cathode-ray tube (CRT) monitors, stripe artifacts are primarily related to the scanning process of the electron beam. The resulting banding patterns differ from those in OLED and LED displays, as factors including different scanning methods (e.g., interlaced vs. progressive) and phosphor persistence characteristics come into play.

\subsubsection{Projectors}

Projectors, especially those using digital light processing (DLP), can also exhibit banding artifacts when captured on camera \citep{han2014fast}. These systems often employ spinning color wheels to generate full-color images, which can interact with the camera's rolling shutter to produce unique colored banding patterns. Brightness modulation is achieved through rapid micromirror switching, which can introduce non-uniform banding under certain filming conditions.

Across the surveyed display families, flicker-banding emerges from different timing mechanisms: row/column or per-block scanning and PWM dimming on LED matrix panels, a mixture of PWM and current-driven modulation with subpixel-phase offsets on OLEDs, raster scanning with phosphor persistence on CRTs, and time-sequential illumination (e.g., color wheels) in projectors. These mechanisms create bands that can be straight or curved (due to rolling-shutter readout), globally periodic or piecewise-structured (under per-block scanning/local modulation), and purely luminance-based or chromatic (due to subpixel phase shifts or color wheels).

Such diversity makes restoration challenging: the banding is frequently non-stationary, its frequency and pattern are strongly device-specific, and its appearance can vary with scene content, exposure settings, and viewing angle. Additionally, ISP processing and sensor noise further blur the boundary between true scene detail and artifact. This calls for future work to develop more advanced, physics-informed models and datasets that reflect this variety.

\begin{figure*}[t]
\centering
\includegraphics[width=0.99\textwidth]{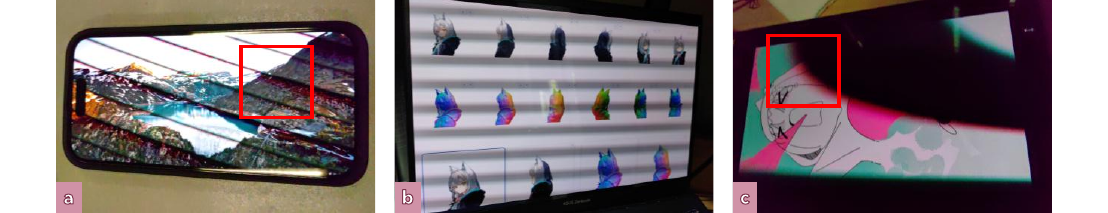}
\caption{Flicker-banding patterns on OLED displays with PWM driving. \textbf{a)} Non-uniform banding in darker areas on an iPhone OLED screen. \textbf{b)} Horizontal gray banding on laptop OLED. \textbf{c)} Extremely bending and color shifts on banding edges.}
\label{Fig.pattern-oled}
\end{figure*}

\begin{figure*}[t]
\centering
\includegraphics[width=0.99\textwidth]{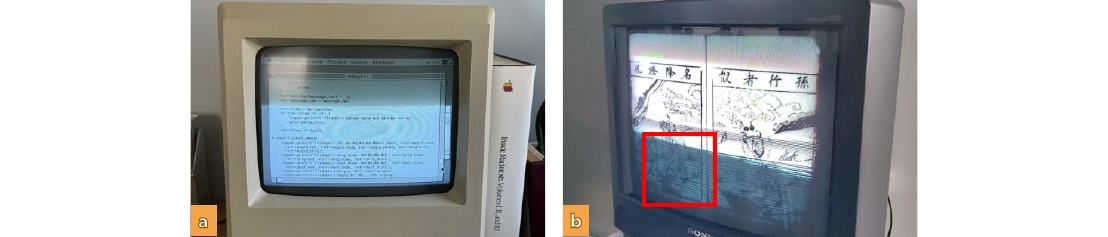}
\caption{Flicker-banding patterns on CRT monitors with electron beam scanning. \textbf{b)} Phosphor persistence effect causing trailing artifacts, leaving a ghosting effect.}
\label{Fig.pattern-crt}
\end{figure*}

\begin{figure*}[t]
\centering
\includegraphics[width=0.99\textwidth]{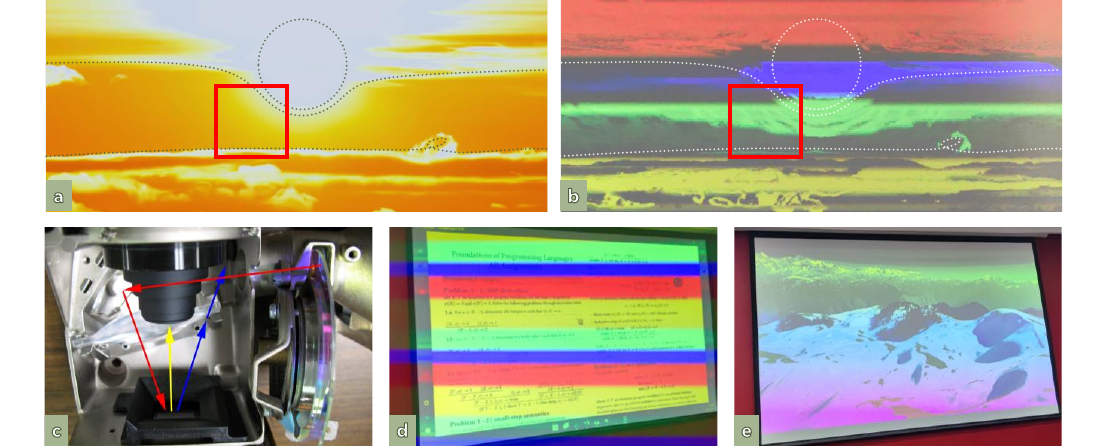}
\caption{Flicker-banding patterns on DLP projectors with color wheel and micromirror switching. \textbf{a-b)} Comparison between original and projected images, highlighting the width of color stripes changes with brightness levels. \textbf{c)} Internal components adapted from DMahalko (CC BY-SA 3.0).}
\label{Fig.pattern-proj}
\end{figure*}
\newcommand{\TabCellBg}{cyan!0}

\begin{figure*}[t]
  \centering
  \setlength{\tabcolsep}{2pt}
  \renewcommand{\arraystretch}{1.0}

  \begin{tabular}{@{}*{4}{>{\centering\arraybackslash}m{0.23\linewidth}}@{}}
    \includegraphics[width=\linewidth]{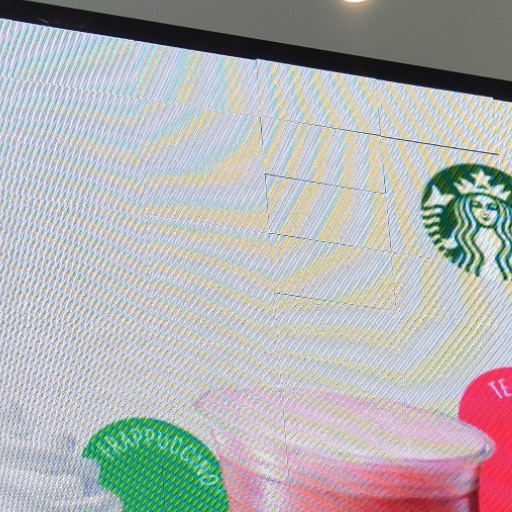} &
    \includegraphics[width=\linewidth]{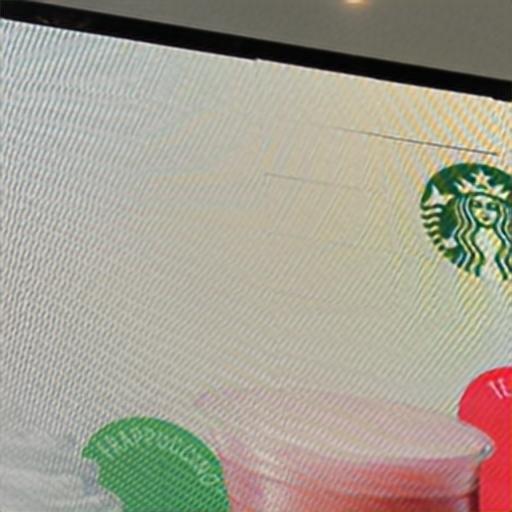} &
    \includegraphics[width=\linewidth]{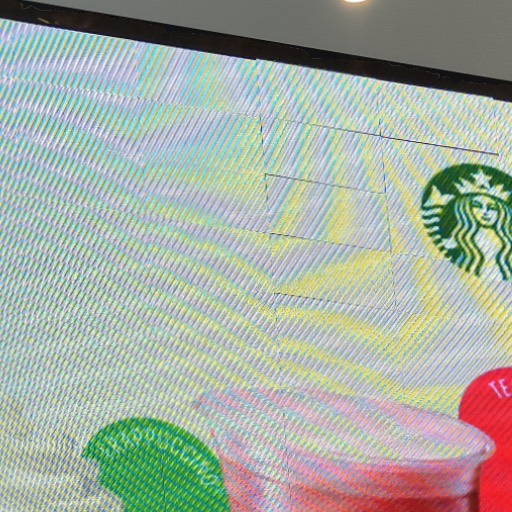} &
    \includegraphics[width=\linewidth]{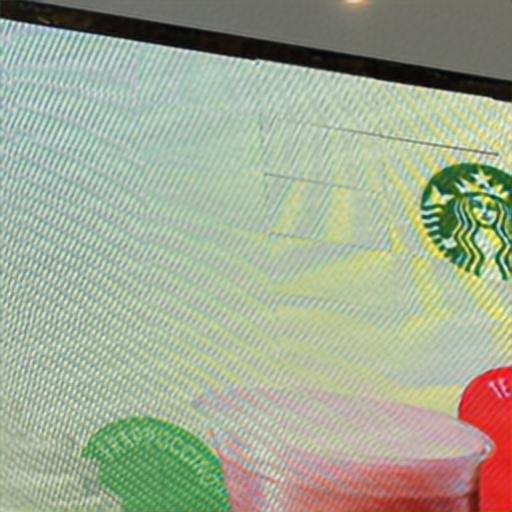} \\
    \includegraphics[width=\linewidth]{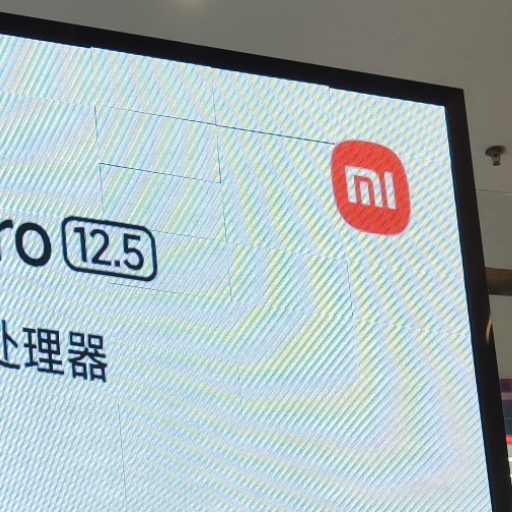} &
    \includegraphics[width=\linewidth]{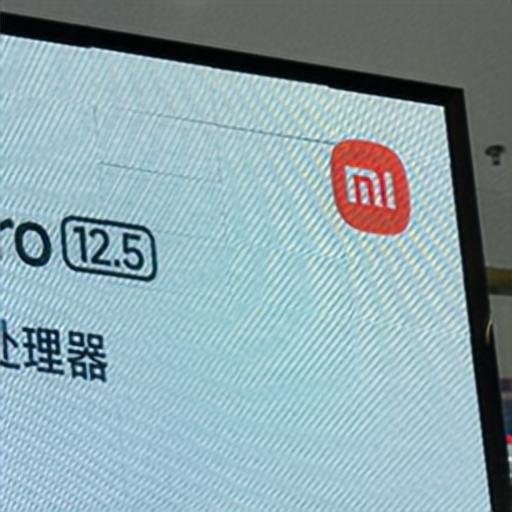} &
    \includegraphics[width=\linewidth]{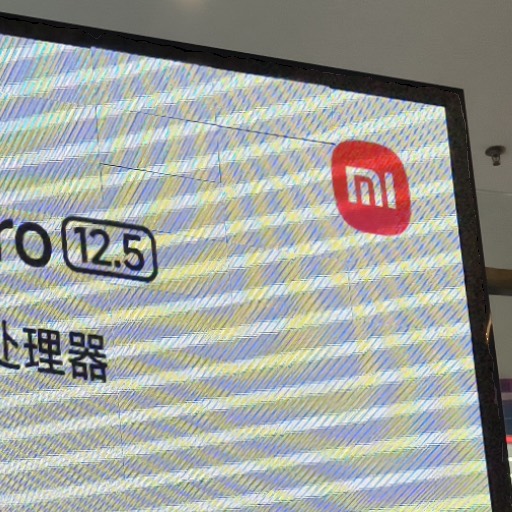} &
    \includegraphics[width=\linewidth]{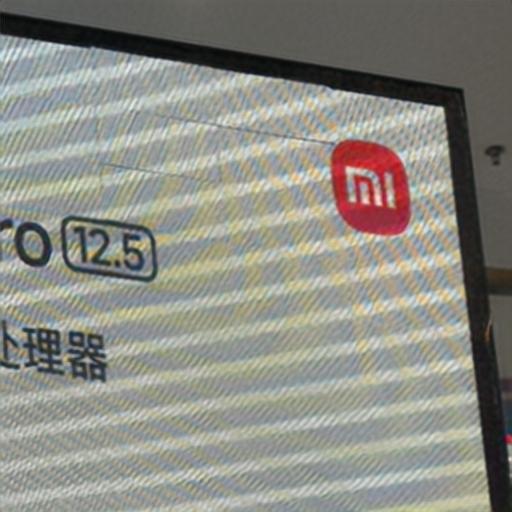} \\
    \small No FB  & \small No FB & \small FB & FB \\
    \small before demoiré  & \small after demoiré & \small before demoiré  & \small after demoiré \\
  \end{tabular}

  \caption{The demoiré performance comparison between no FB scene and FB scene.}
  \label{fig:demoire test}
\end{figure*}

\subsection{Additional Visual Comparison}
\label{app:additional-visual}
\begin{figure*}[t]
  \scriptsize
  \centering
  \begin{tabular}{@{}cc@{}}
    \hspace{-0.35cm}
    \begin{adjustbox}{valign=t}
    \begin{tabular}{c}
      \includegraphics[width=0.26\textwidth]{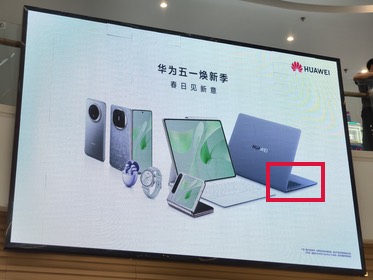}\\
      Real-Flicker: Banding07\_2
    \end{tabular}
    \end{adjustbox}
    \hspace{-0.30cm}
    &
    \begin{adjustbox}{valign=t}
    \begin{tabular}{ccccc}
      \includegraphics[width=0.13\textwidth]{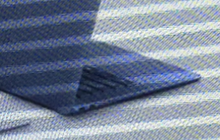}\hspace{-3mm} &
      \includegraphics[width=0.13\textwidth]{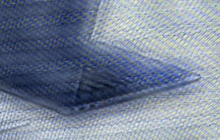}\hspace{-3mm} &
      \includegraphics[width=0.13\textwidth]{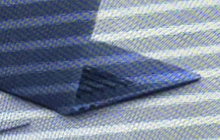}\hspace{-3mm} &
      \includegraphics[width=0.13\textwidth]{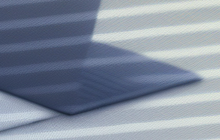}\hspace{-3mm} &
      \includegraphics[width=0.13\textwidth]{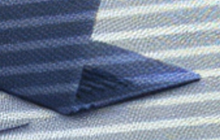}\hspace{-3mm} \\
      LQ \hspace{-3mm} & Notch Filter \hspace{-3mm} & MAT \hspace{-3mm} & InvSR \hspace{-3mm} & Step1X \hspace{-3mm} \\
      \includegraphics[width=0.13\textwidth]{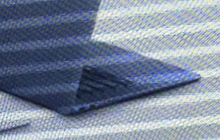}\hspace{-3mm} &
      \includegraphics[width=0.13\textwidth]{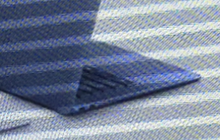}\hspace{-3mm} &
      \includegraphics[width=0.13\textwidth]{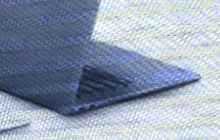}\hspace{-3mm} &
      \includegraphics[width=0.13\textwidth]{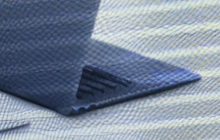}\hspace{-3mm} &
      \includegraphics[width=0.13\textwidth]{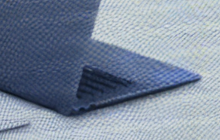}\hspace{-3mm} \\
      ESDNet \hspace{-3mm} & FlickerFormer* \hspace{-3mm} & FlickerFormer \hspace{-3mm} & PiSA-SR \hspace{-3mm} & RIFLE \hspace{-3mm}
    \end{tabular}
    \end{adjustbox}
  \end{tabular}
  \vspace{-0.8mm}

  \begin{tabular}{@{}cc@{}}
    \hspace{-0.35cm}
    \begin{adjustbox}{valign=t}
    \begin{tabular}{c}
      \includegraphics[width=0.26\textwidth]{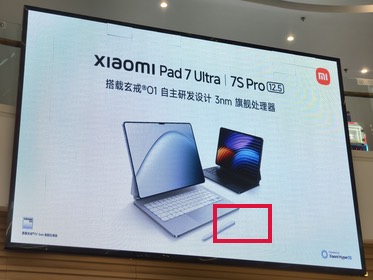}\\
      Real-Flicker: Banding15\_2
    \end{tabular}
    \end{adjustbox}
    \hspace{-0.30cm}
    &
    \begin{adjustbox}{valign=t}
    \begin{tabular}{ccccc}
      \includegraphics[width=0.13\textwidth]{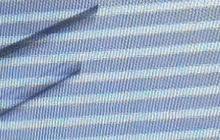}\hspace{-3mm} &
      \includegraphics[width=0.13\textwidth]{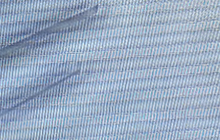}\hspace{-3mm} &
      \includegraphics[width=0.13\textwidth]{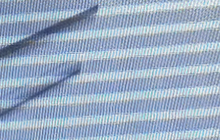}\hspace{-3mm} &
      \includegraphics[width=0.13\textwidth]{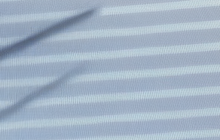}\hspace{-3mm} &
      \includegraphics[width=0.13\textwidth]{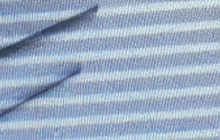}\hspace{-3mm} \\
      LQ \hspace{-3mm} & Notch Filter \hspace{-3mm} & MAT \hspace{-3mm} & InvSR \hspace{-3mm} & Step1X \hspace{-3mm} \\
      \includegraphics[width=0.13\textwidth]{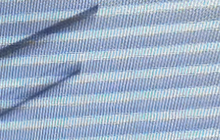}\hspace{-3mm} &
      \includegraphics[width=0.13\textwidth]{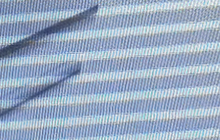}\hspace{-3mm} &
      \includegraphics[width=0.13\textwidth]{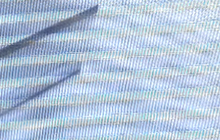}\hspace{-3mm} &
      \includegraphics[width=0.13\textwidth]{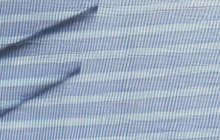}\hspace{-3mm} &
      \includegraphics[width=0.13\textwidth]{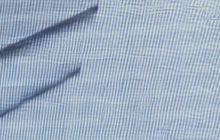}\hspace{-3mm} \\
      ESDNet \hspace{-3mm} & FlickerFormer* \hspace{-3mm} & FlickerFormer \hspace{-3mm} & PiSA-SR \hspace{-3mm} & RIFLE \hspace{-3mm}
    \end{tabular}
    \end{adjustbox}
  \end{tabular}
  \vspace{-0.8mm}

  \begin{tabular}{@{}cc@{}}
    \hspace{-0.35cm}
    \begin{adjustbox}{valign=t}
    \begin{tabular}{c}
      \includegraphics[width=0.26\textwidth]{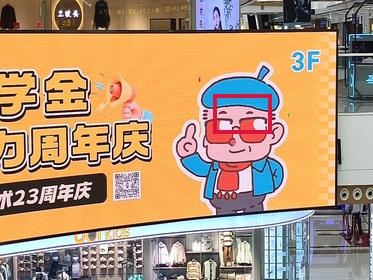}\\
      Real-Flicker: Banding40\_1
    \end{tabular}
    \end{adjustbox}
    \hspace{-0.30cm}
    &
    \begin{adjustbox}{valign=t}
    \begin{tabular}{ccccc}
      \includegraphics[width=0.13\textwidth]{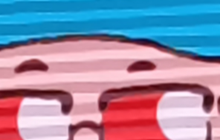}\hspace{-3mm} &
      \includegraphics[width=0.13\textwidth]{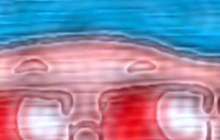}\hspace{-3mm} &
      \includegraphics[width=0.13\textwidth]{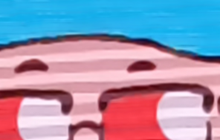}\hspace{-3mm} &
      \includegraphics[width=0.13\textwidth]{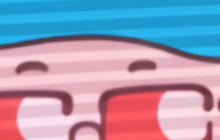}\hspace{-3mm} &
      \includegraphics[width=0.13\textwidth]{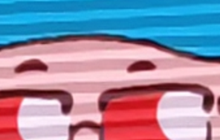}\hspace{-3mm} \\
      LQ \hspace{-3mm} & Notch Filter \hspace{-3mm} & MAT \hspace{-3mm} & InvSR \hspace{-3mm} & Step1X \hspace{-3mm} \\
      \includegraphics[width=0.13\textwidth]{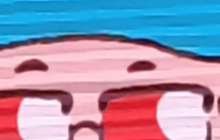}\hspace{-3mm} &
      \includegraphics[width=0.13\textwidth]{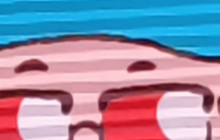}\hspace{-3mm} &
      \includegraphics[width=0.13\textwidth]{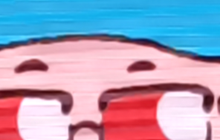}\hspace{-3mm} &
      \includegraphics[width=0.13\textwidth]{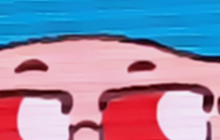}\hspace{-3mm} &
      \includegraphics[width=0.13\textwidth]{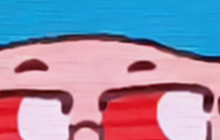}\hspace{-3mm} \\
      ESDNet \hspace{-3mm} & FlickerFormer* \hspace{-3mm} & FlickerFormer \hspace{-3mm} & PiSA-SR \hspace{-3mm} & RIFLE \hspace{-3mm}
    \end{tabular}
    \end{adjustbox}
  \end{tabular}
  \vspace{-0.8mm}

  \begin{tabular}{@{}cc@{}}
    \hspace{-0.35cm}
    \begin{adjustbox}{valign=t}
    \begin{tabular}{c}
      \includegraphics[width=0.26\textwidth]{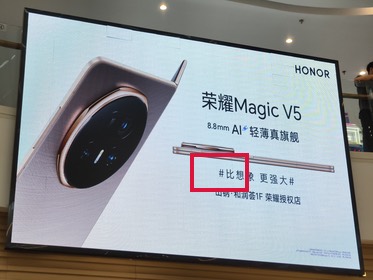}\\
      Real-Flicker: Banding03\_1
    \end{tabular}
    \end{adjustbox}
    \hspace{-0.30cm}
    &
    \begin{adjustbox}{valign=t}
    \begin{tabular}{ccccc}
      \includegraphics[width=0.13\textwidth]{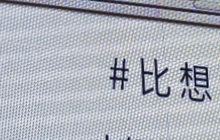}\hspace{-3mm} &
      \includegraphics[width=0.13\textwidth]{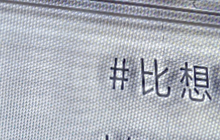}\hspace{-3mm} &
      \includegraphics[width=0.13\textwidth]{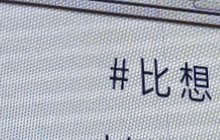}\hspace{-3mm} &
      \includegraphics[width=0.13\textwidth]{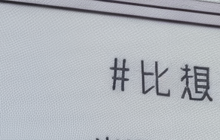}\hspace{-3mm} &
      \includegraphics[width=0.13\textwidth]{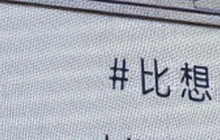}\hspace{-3mm} \\
      LQ \hspace{-3mm} & Notch Filter \hspace{-3mm} & MAT \hspace{-3mm} & InvSR \hspace{-3mm} & Step1X \hspace{-3mm} \\
      \includegraphics[width=0.13\textwidth]{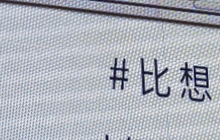}\hspace{-3mm} &
      \includegraphics[width=0.13\textwidth]{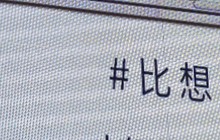}\hspace{-3mm} &
      \includegraphics[width=0.13\textwidth]{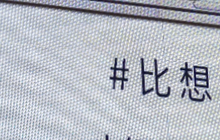}\hspace{-3mm} &
      \includegraphics[width=0.13\textwidth]{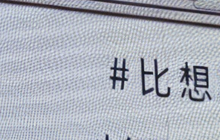}\hspace{-3mm} &
      \includegraphics[width=0.13\textwidth]{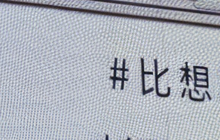}\hspace{-3mm} \\
      ESDNet \hspace{-3mm} & FlickerFormer* \hspace{-3mm} & FlickerFormer \hspace{-3mm} & PiSA-SR \hspace{-3mm} & RIFLE \hspace{-3mm}
    \end{tabular}
    \end{adjustbox}
  \end{tabular}
  \vspace{-0.8mm}

  \begin{tabular}{@{}cc@{}}
    \hspace{-0.35cm}
    \begin{adjustbox}{valign=t}
    \begin{tabular}{c}
      \includegraphics[width=0.26\textwidth]{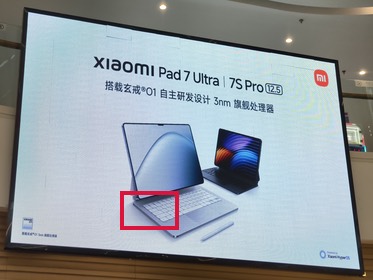}\\
      Real-Flicker: Banding15\_1
    \end{tabular}
    \end{adjustbox}
    \hspace{-0.30cm}
    &
    \begin{adjustbox}{valign=t}
    \begin{tabular}{ccccc}
      \includegraphics[width=0.13\textwidth]{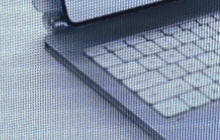}\hspace{-3mm} &
      \includegraphics[width=0.13\textwidth]{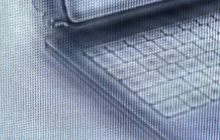}\hspace{-3mm} &
      \includegraphics[width=0.13\textwidth]{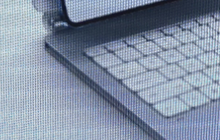}\hspace{-3mm} &
      \includegraphics[width=0.13\textwidth]{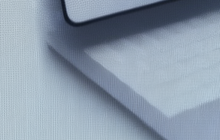}\hspace{-3mm} &
      \includegraphics[width=0.13\textwidth]{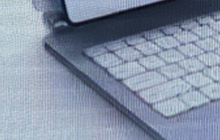}\hspace{-3mm} \\
      LQ \hspace{-3mm} & Notch Filter \hspace{-3mm} & MAT \hspace{-3mm} & InvSR \hspace{-3mm} & Step1X \hspace{-3mm} \\
      \includegraphics[width=0.13\textwidth]{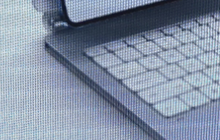}\hspace{-3mm} &
      \includegraphics[width=0.13\textwidth]{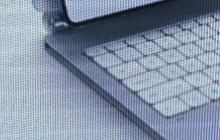}\hspace{-3mm} &
      \includegraphics[width=0.13\textwidth]{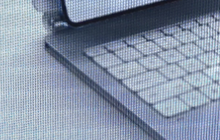}\hspace{-3mm} &
      \includegraphics[width=0.13\textwidth]{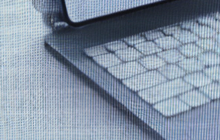}\hspace{-3mm} &
      \includegraphics[width=0.13\textwidth]{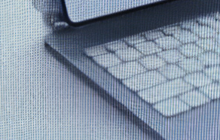}\hspace{-3mm} \\
      ESDNet \hspace{-3mm} & FlickerFormer* \hspace{-3mm} & FlickerFormer \hspace{-3mm} & PiSA-SR \hspace{-3mm} & RIFLE \hspace{-3mm}
    \end{tabular}
    \end{adjustbox}
  \end{tabular}
  \vspace{-0.8mm}

  \begin{tabular}{@{}cc@{}}
    \hspace{-0.35cm}
    \begin{adjustbox}{valign=t}
    \begin{tabular}{c}
      \includegraphics[width=0.26\textwidth]{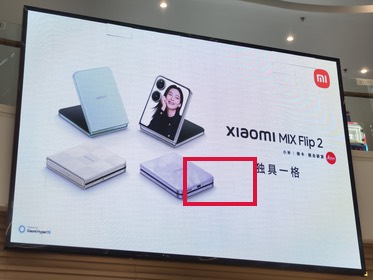}\\
      Real-Flicker: Banding17\_2
    \end{tabular}
    \end{adjustbox}
    \hspace{-0.30cm}
    &
    \begin{adjustbox}{valign=t}
    \begin{tabular}{ccccc}
      \includegraphics[width=0.13\textwidth]{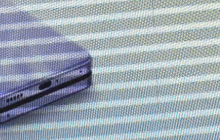}\hspace{-3mm} &
      \includegraphics[width=0.13\textwidth]{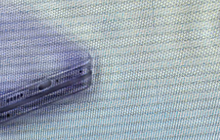}\hspace{-3mm} &
      \includegraphics[width=0.13\textwidth]{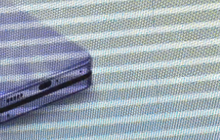}\hspace{-3mm} &
      \includegraphics[width=0.13\textwidth]{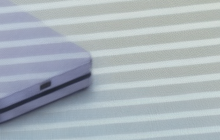}\hspace{-3mm} &
      \includegraphics[width=0.13\textwidth]{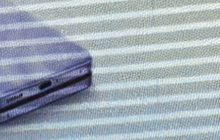}\hspace{-3mm} \\
      LQ \hspace{-3mm} & Notch Filter \hspace{-3mm} & MAT \hspace{-3mm} & InvSR \hspace{-3mm} & Step1X \hspace{-3mm} \\
      \includegraphics[width=0.13\textwidth]{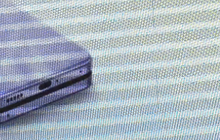}\hspace{-3mm} &
      \includegraphics[width=0.13\textwidth]{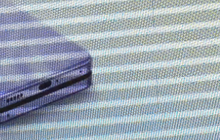}\hspace{-3mm} &
      \includegraphics[width=0.13\textwidth]{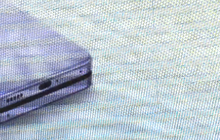}\hspace{-3mm} &
      \includegraphics[width=0.13\textwidth]{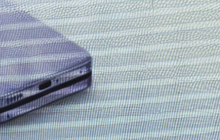}\hspace{-3mm} &
      \includegraphics[width=0.13\textwidth]{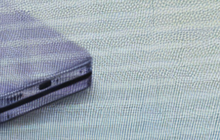}\hspace{-3mm} \\
      ESDNet \hspace{-3mm} & FlickerFormer* \hspace{-3mm} & FlickerFormer \hspace{-3mm} & PiSA-SR \hspace{-3mm} & RIFLE \hspace{-3mm}
    \end{tabular}
    \end{adjustbox}
  \end{tabular}
  \vspace{2pt}
  \caption{Additional visual comparison on the remaining Section-1 cases that are not used in Fig.~\ref{fig:main-visual-comparison}.}
  \label{fig:Additional-Visual-Comparison}
\end{figure*}

We provide additional visual comparison for different methods on our Real-Flicker dataset in Fig.~\ref{fig:Additional-Visual-Comparison}. More results demonstrate our RIFLE's outstanding performance compared to recent image reconstruction methods and show strong application potential in real-world scenarios.

\subsection{Demoir\'e model performance in the case of Flicker-banding}
\label{app:demoire-fb}

In addition to the flikcer-banding (FB), moiré patterns are also bring great damage to the image quality when shooting screen. We find that the performance of demoiré model largely decrease when both of the FB and moiré patterns exist, as shown in Fig.~\ref{fig:demoire test}. It indicates the difference between FB and moiré patterns and removing them simultaneously is also a valuable issue to be solved.

\subsection{User Study}
\label{app:user-study}

\begin{figure}[t]
    \centering
    \includegraphics[width=0.95\linewidth]{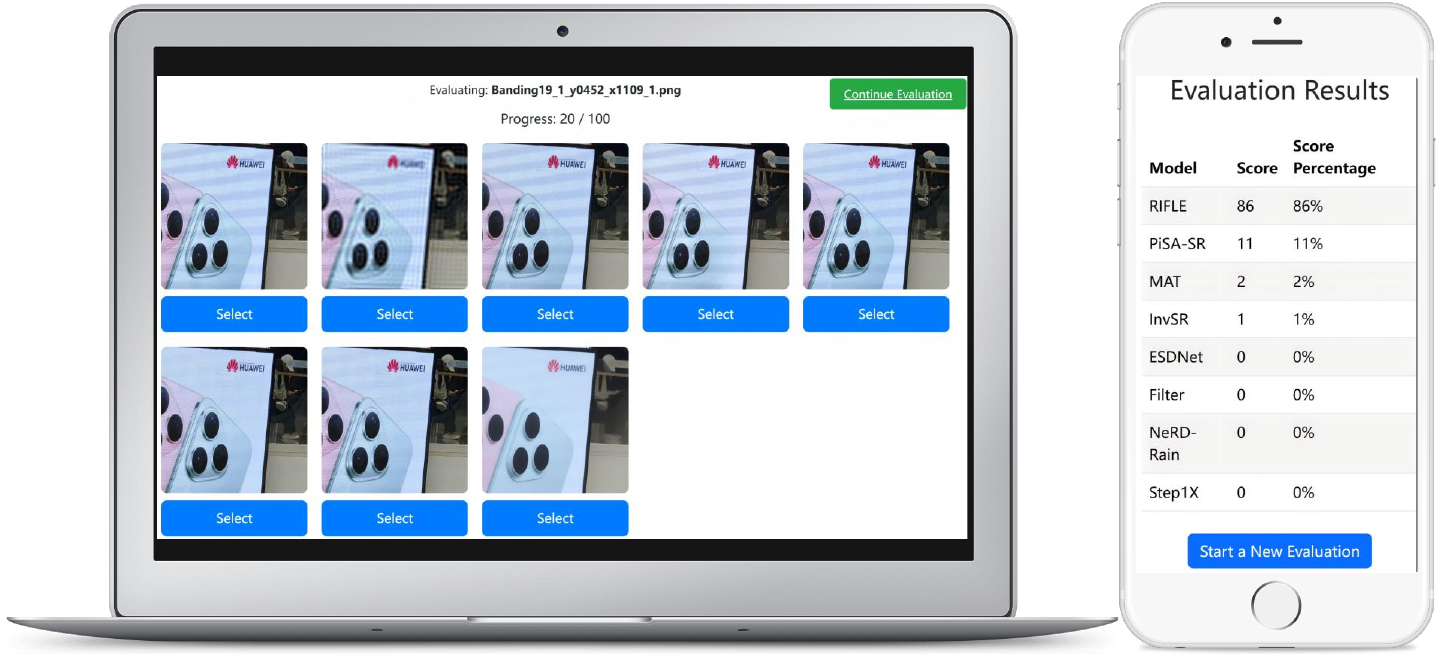}
    \caption{The operating system of the User Study.}
    \label{fig:evaluation panel}
\end{figure}

\begin{table*}[t]
\centering
\scriptsize
\caption{The results of User Study.}
\vspace{-3mm}
\resizebox{\textwidth}{!}
{
\begin{tabular}{c |  c c  c c c c c c}
\toprule
\rowcolor{iclrcolor!30}
{Methods} & RIFLE & PiSA-SR & MAT & InvSR & Step1X & ESDNet & FlickerFormer & Notch Filter \\
\midrule
Score/(\%) & 82.4 & 14.7 & 2.7 & 0.02 & 0 & 0.1 & 0.08 & 0\\
\bottomrule
\end{tabular}
}
\label{tab: User Study}
\end{table*}

Standard image-quality metrics do not directly measure FB severity, so we complement them with a single-choice preference study. Figure~\ref{fig:evaluation panel} shows the evaluation interface. All participants used the same interface under consistent viewing conditions; each trial displayed the eight outputs for the same test case simultaneously at the same size and layout, with method names hidden. Before the study, participants received the same explanation of the evaluation objective and selection criterion, together with practice guidance on operating the interface.

Each of the 50 participants completed one session of 100 trials, selecting the result with the best FB removal performance before proceeding to the next case. The study therefore produced 5,000 valid decisions across 100 test cases and eight methods. Scores in Table~\ref{tab: User Study} are the number of selections for each method divided by the total number of valid decisions; they are preference shares, not rankings or numerical quality ratings.

RIFLE received 4,120 selections, corresponding to 82.4\%. At the decision level, the 95\% Wilson confidence interval is [81.32\%, 83.43\%], and a one-sided exact binomial test against chance selection among eight methods ($p_0=1/8$) gives $p<0.001$. These statistics summarize the observed decisions; they do not model dependence across repeated decisions from the same participant or test case.

\subsection{The out-of-distribution test results.}
\label{app:ood-results}

\begin{figure*}[t]
  \centering
  \setlength{\tabcolsep}{2pt}
  \renewcommand{\arraystretch}{1.0}

  \begin{tabular}{@{}*{2}{>{\centering\arraybackslash}m{0.48\linewidth}}@{}}
    \includegraphics[width=\linewidth]{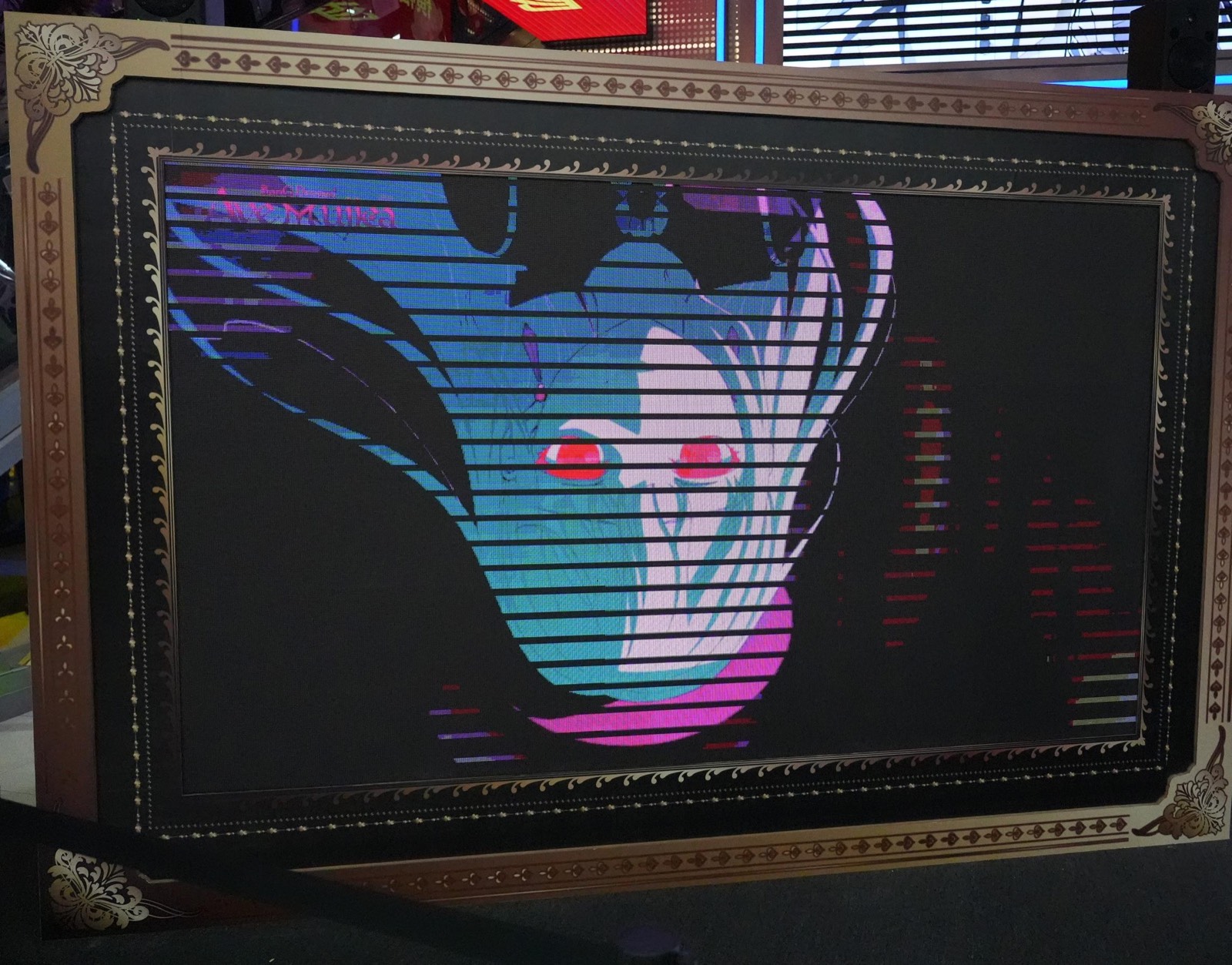} &
    \includegraphics[width=\linewidth]{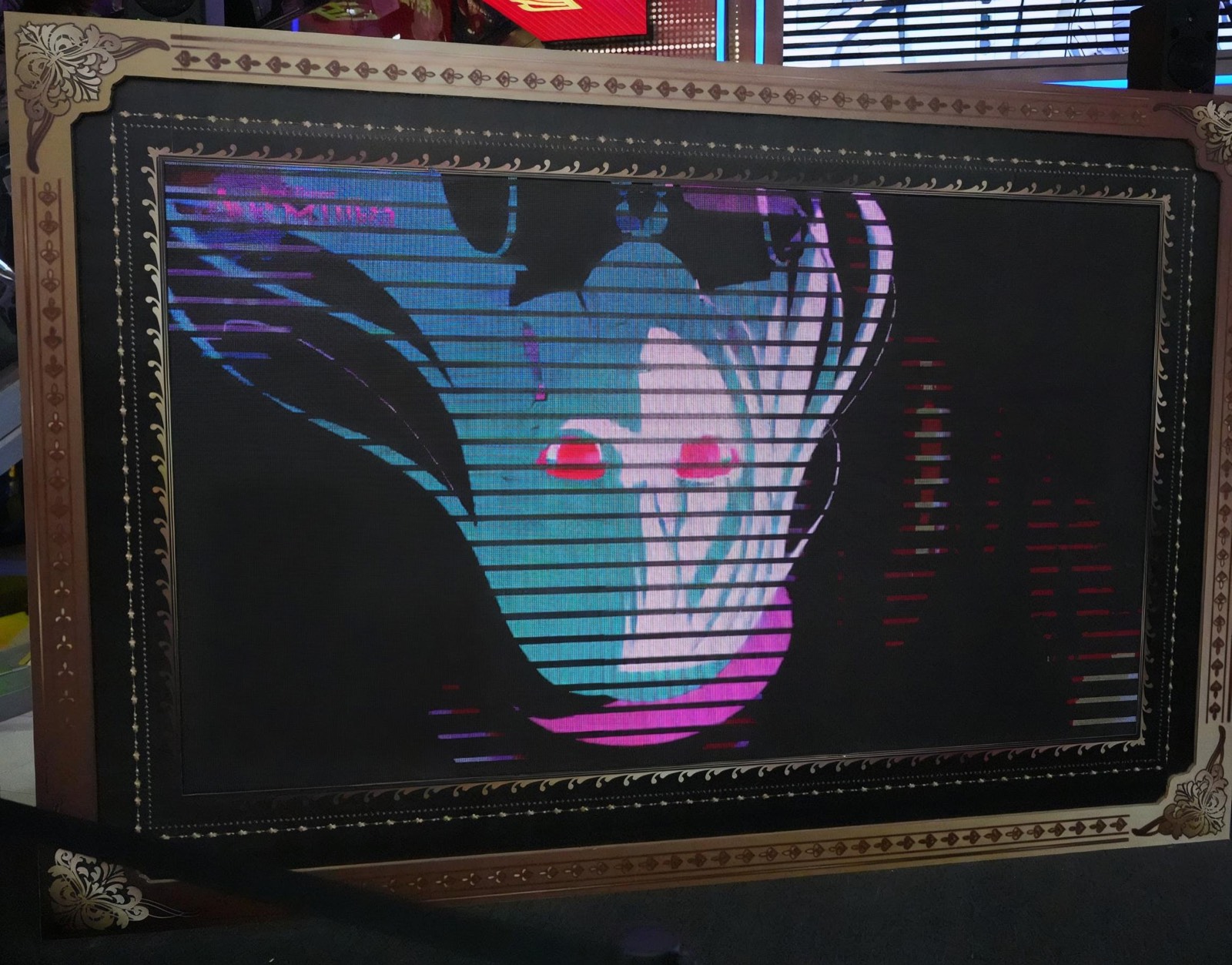} \\
    \includegraphics[width=\linewidth]{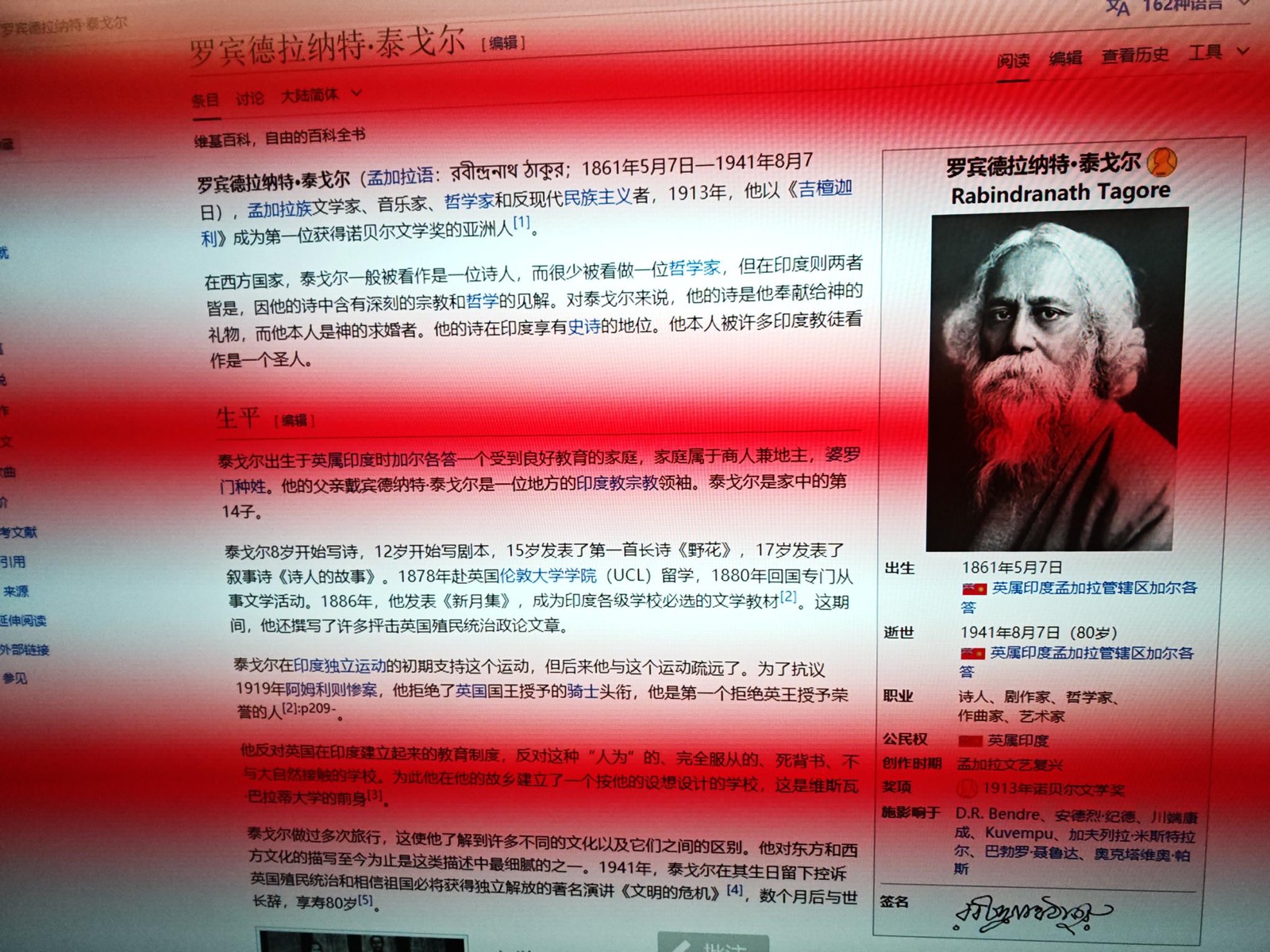} &
    \includegraphics[width=\linewidth]{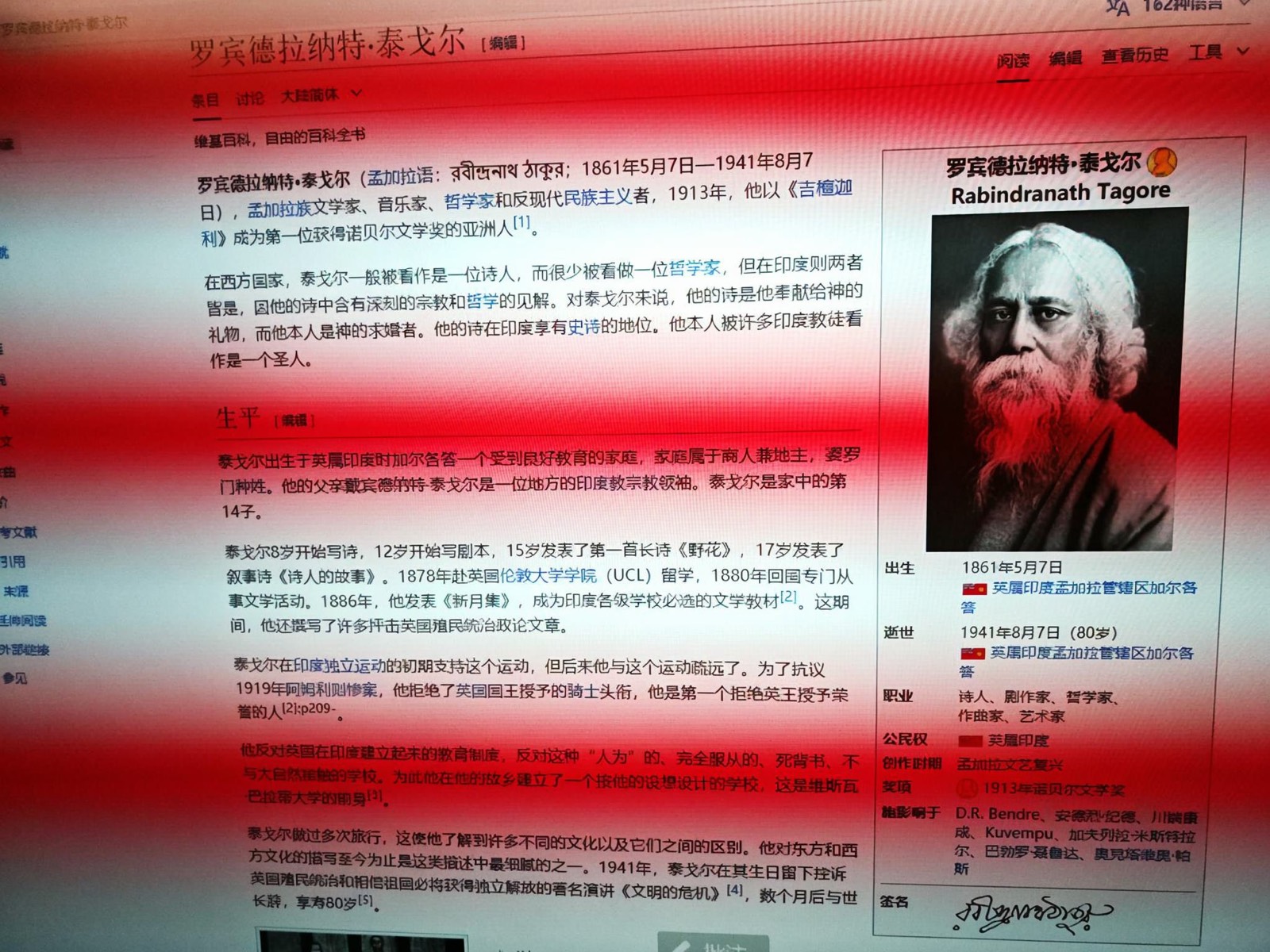} \\
    \includegraphics[width=\linewidth]{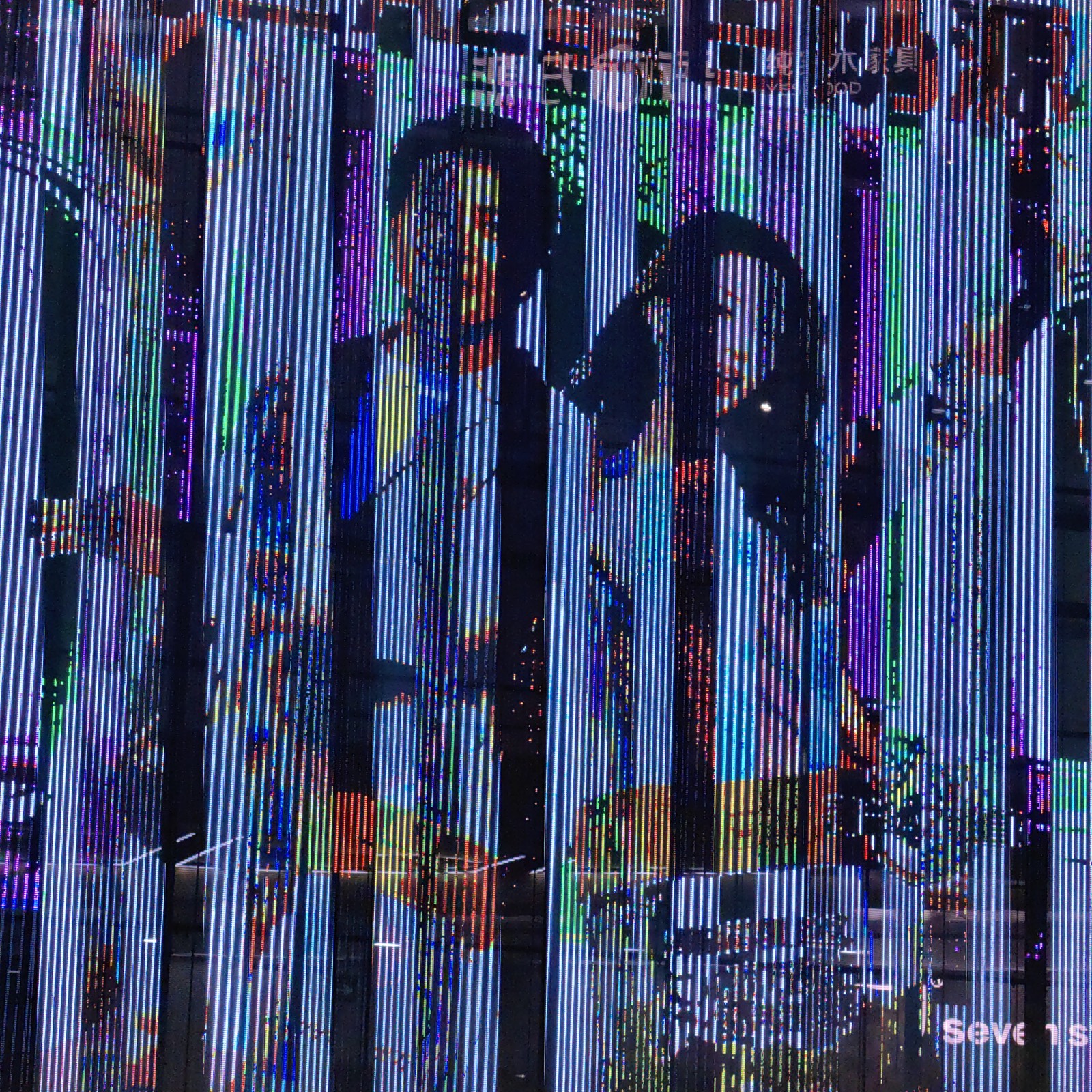} &
    \includegraphics[width=\linewidth]{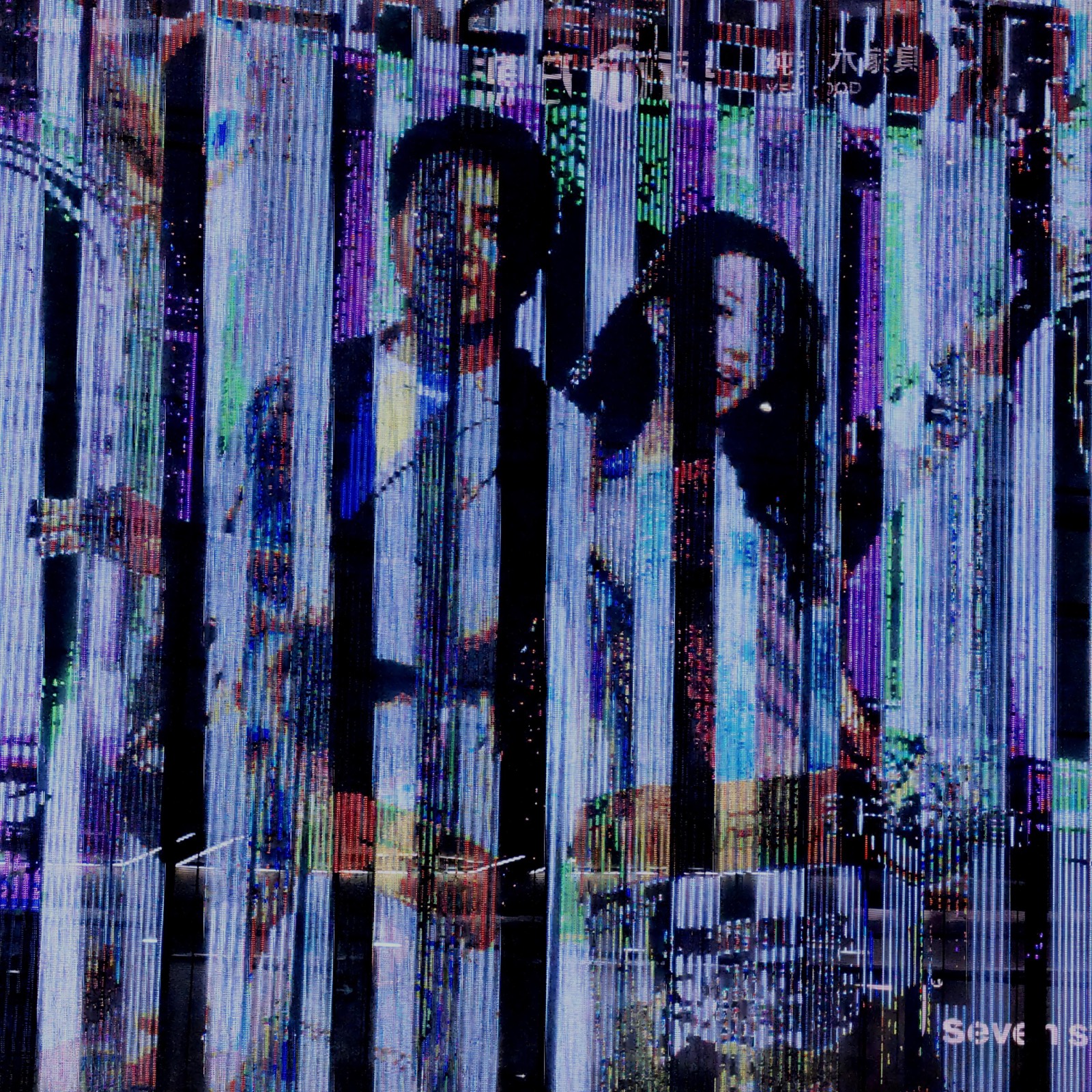} \\
    \small Input  & \small Reconstructed \\
  \end{tabular}
  \caption{The out-of-distribution test results.}
  \vspace{-4mm}
  \label{fig:failure test}
\end{figure*}

We provide the test results on other severe flicker-banding (FB) scenes in Fig.~\ref{fig:failure test}, not only on the dark straight FB scenes. The first row indicates that the black stripes are hard to recover and our model only played a minor role in filling in the gaps. The second row indicates that our model makes no difference to the stripes with severe color shifts. The third row implies that there is still a long way to go for our model to reconstruct an image whose details are completely destroyed.

\subsection{Analysis on the Flicker-banding simulation}
\label{app:simulation-analysis}

\begin{figure*}[t]
  \centering
  \setlength{\tabcolsep}{2pt}
  \renewcommand{\arraystretch}{1.0}

  \begin{tabular}{@{}m{0.03\linewidth}*{3}{>{\centering\arraybackslash}m{0.315\linewidth}}@{}}
    \small GT &
    \includegraphics[width=\linewidth]{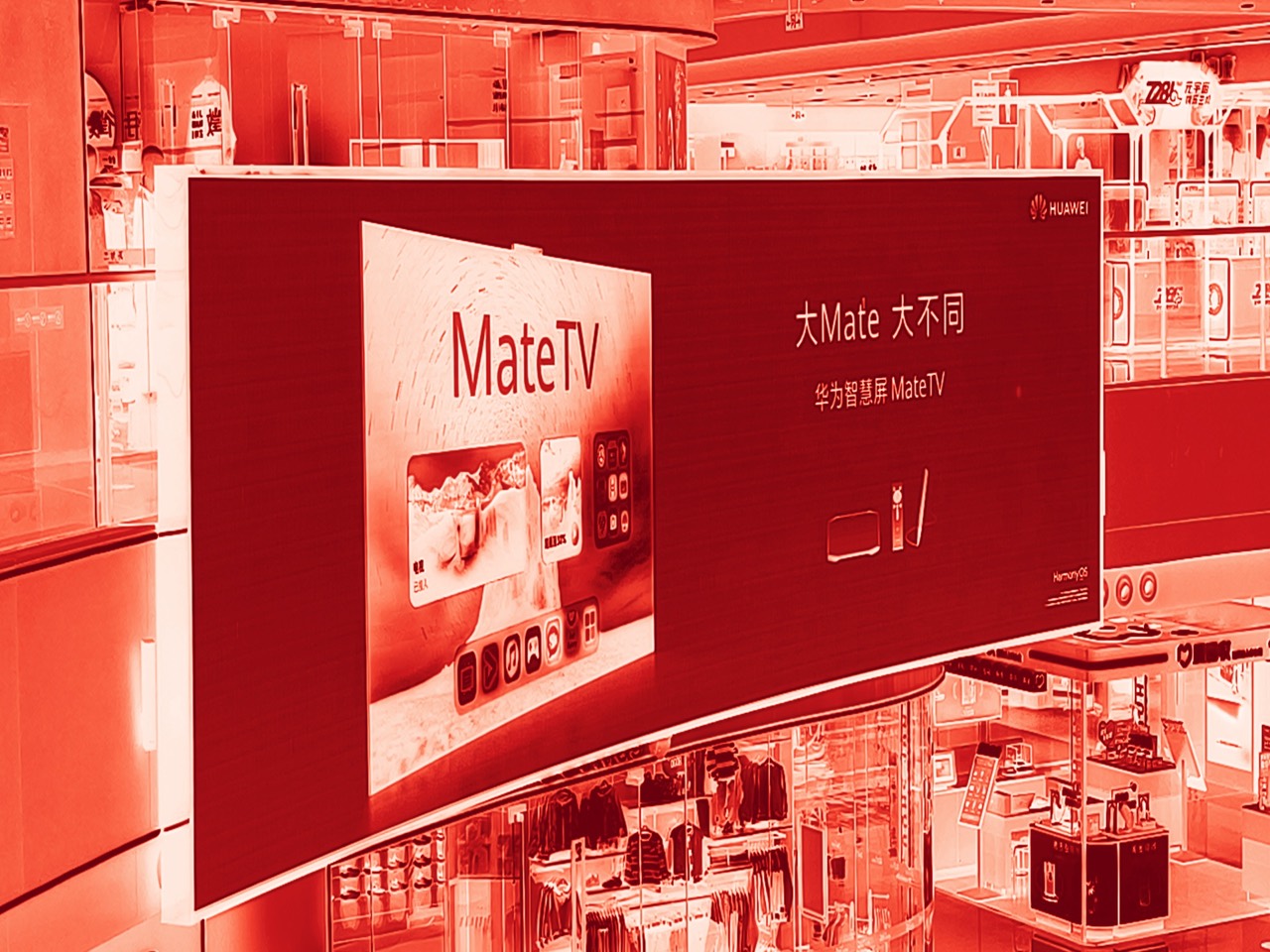} &
    \includegraphics[width=\linewidth]{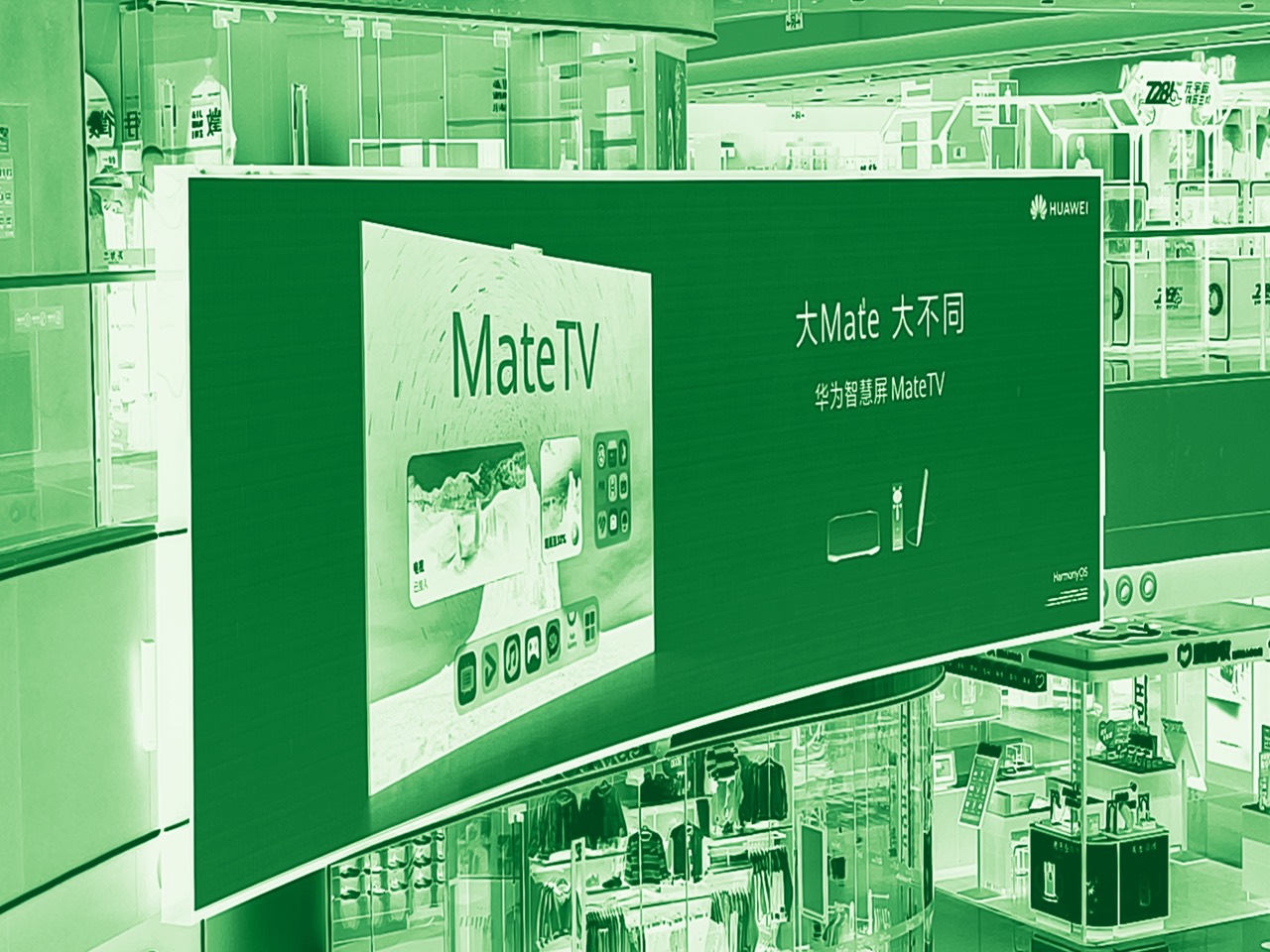} &
    \includegraphics[width=\linewidth]{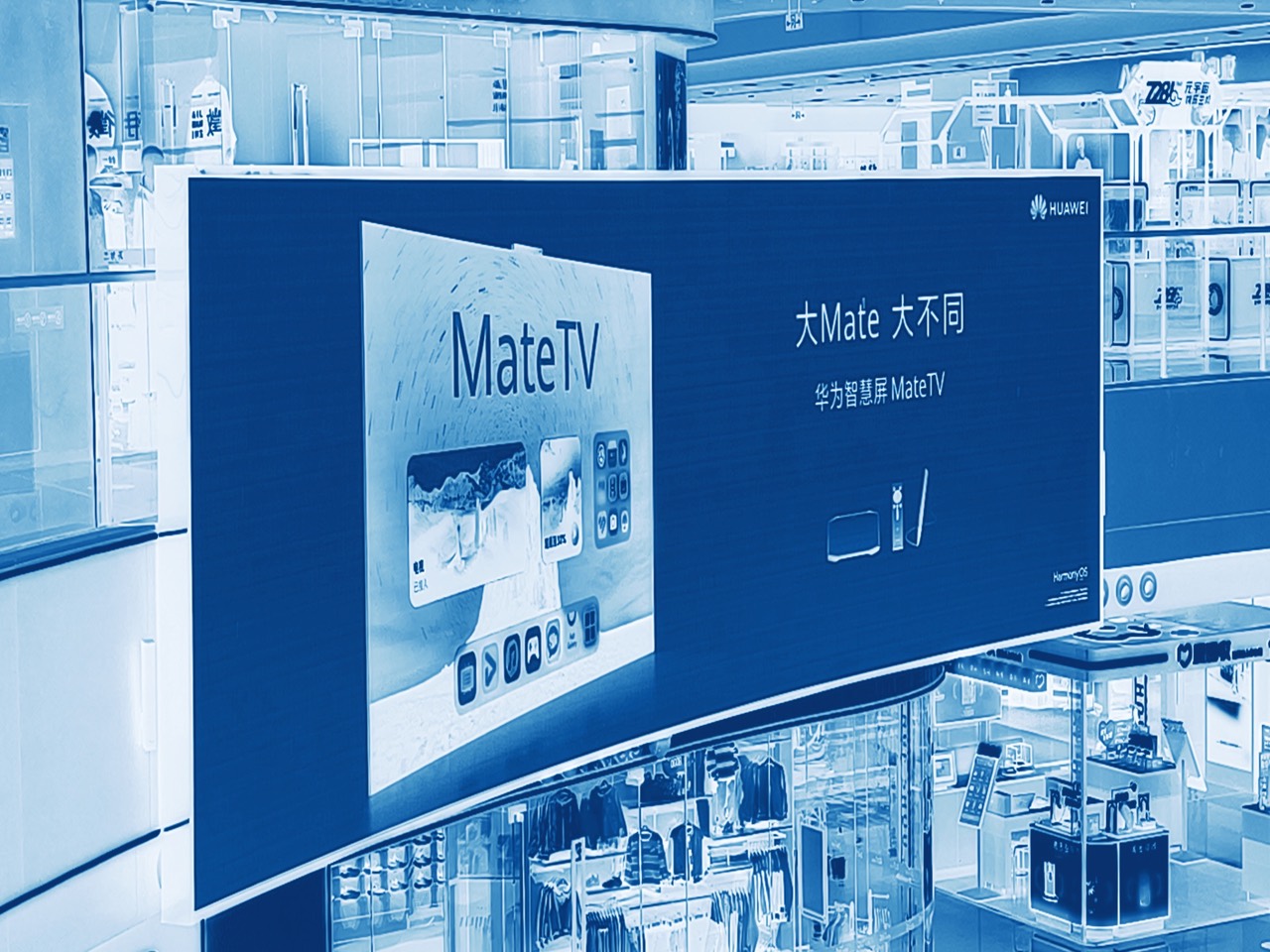} \\
    \small LQ &
    \includegraphics[width=\linewidth]{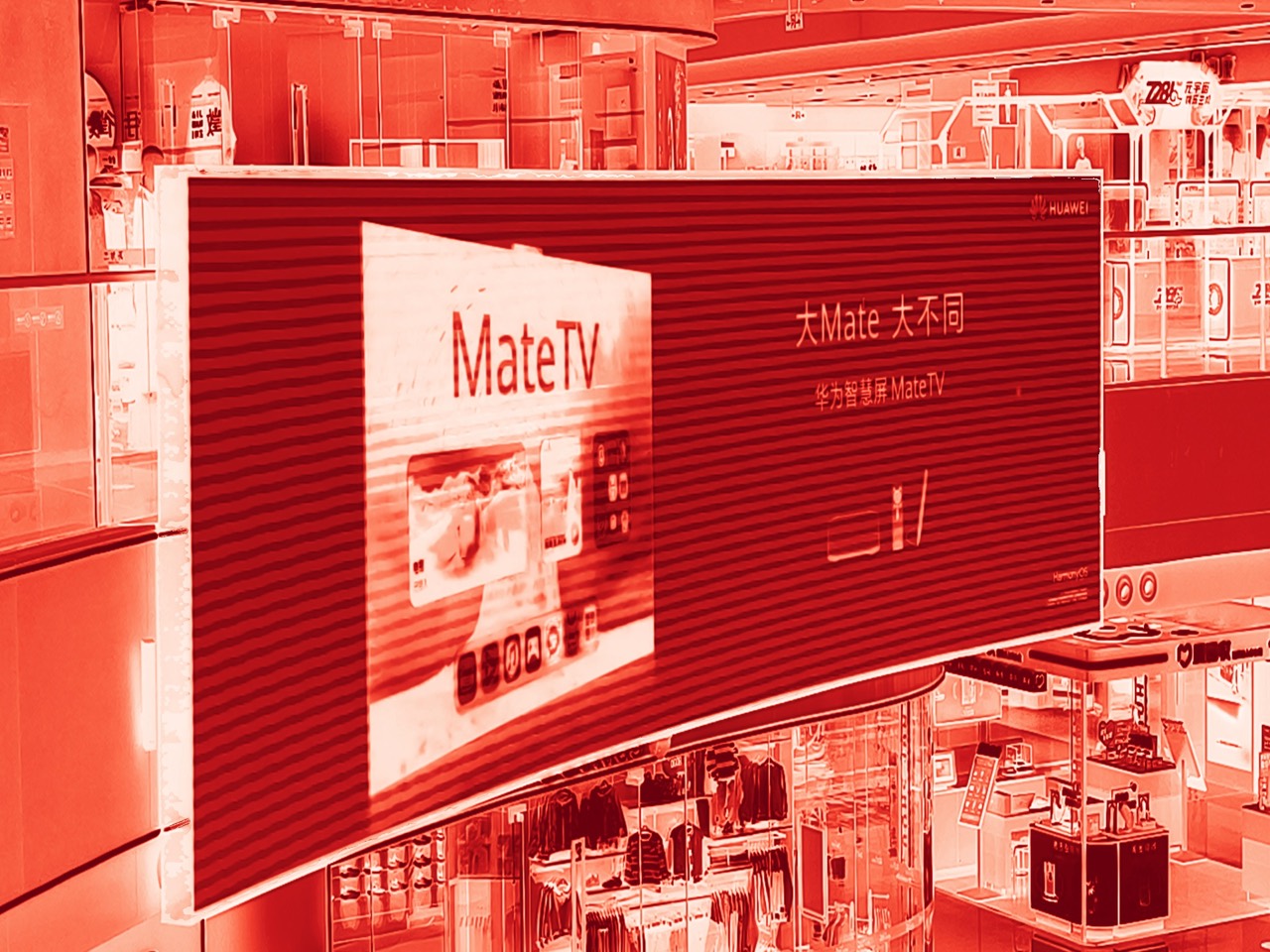} &
    \includegraphics[width=\linewidth]{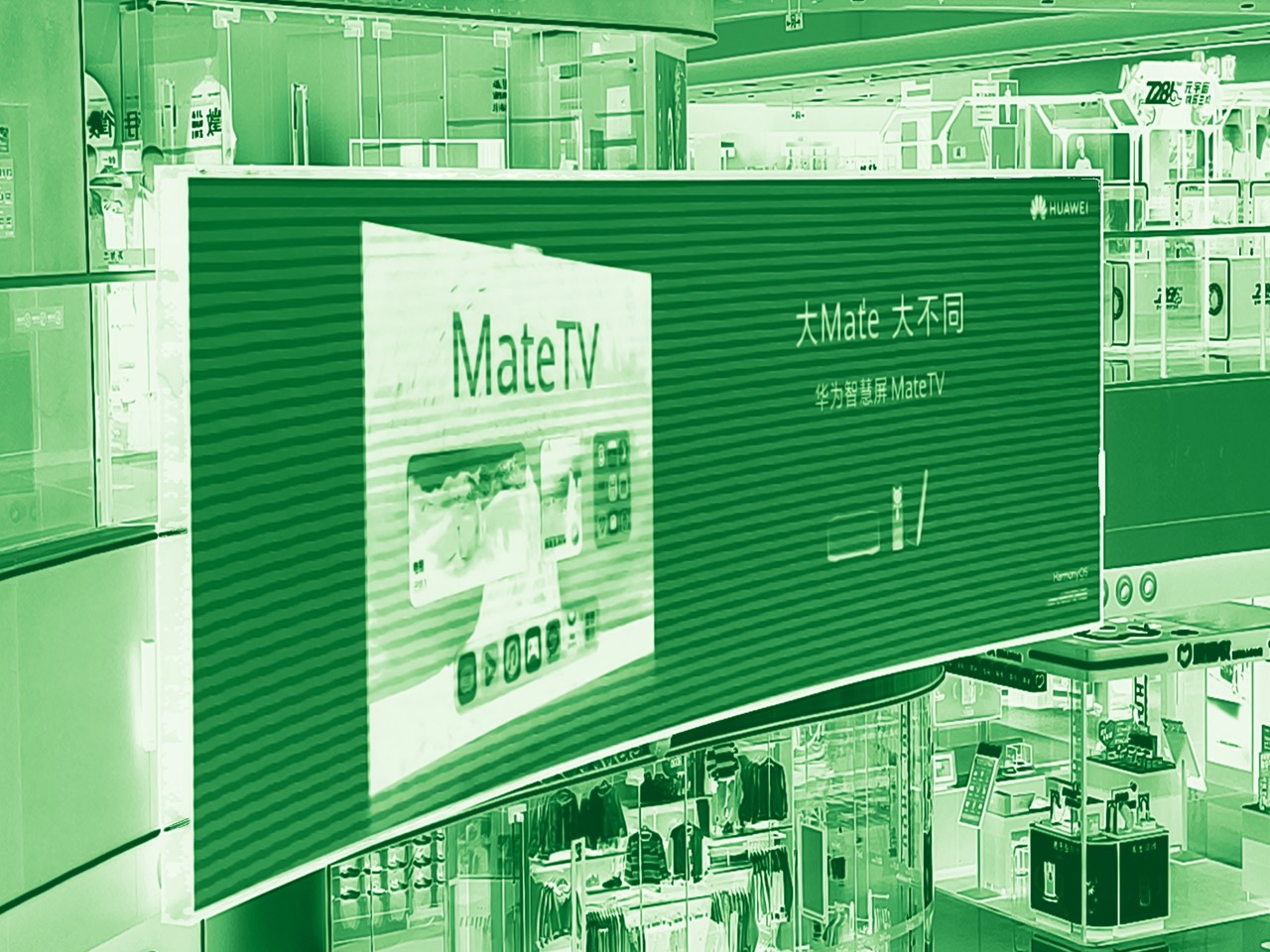} &
    \includegraphics[width=\linewidth]{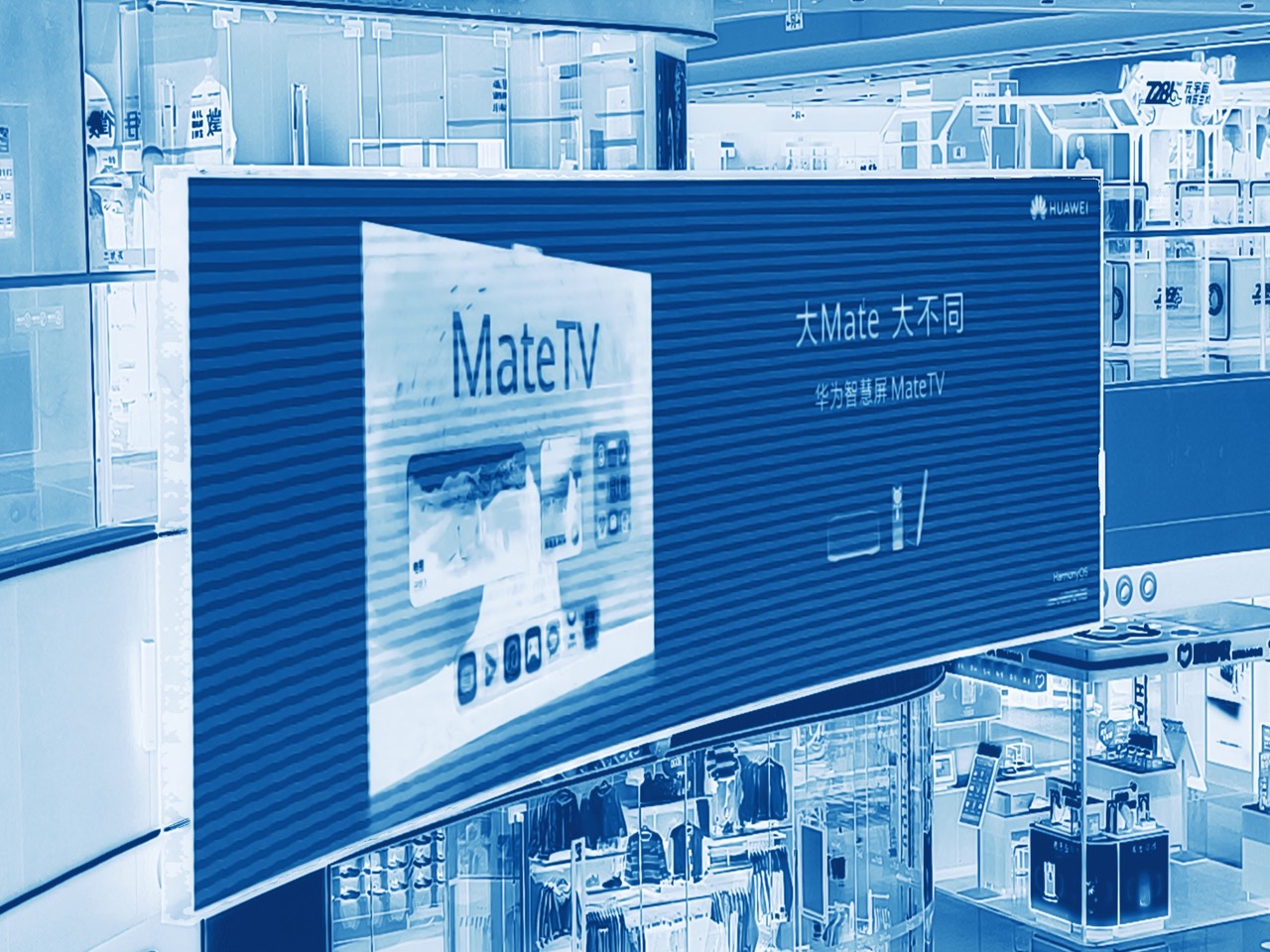} \\
    & \small R & \small G & \small B \\
  \end{tabular}

  \caption{Per-channel visualization of banding-free images (GT) and flicker-banding images (LQ) in RGB color space. Flicker-banding artifacts appear in all three channels of the LQ images.}
  \label{fig:RGB visualization}
\end{figure*}

\begin{figure*}[t]
  \centering
  \setlength{\tabcolsep}{2pt}
  \renewcommand{\arraystretch}{1.0}

  \begin{tabular}{@{}m{0.03\linewidth}*{3}{>{\centering\arraybackslash}m{0.315\linewidth}}@{}}
    \small GT &
    \includegraphics[width=\linewidth]{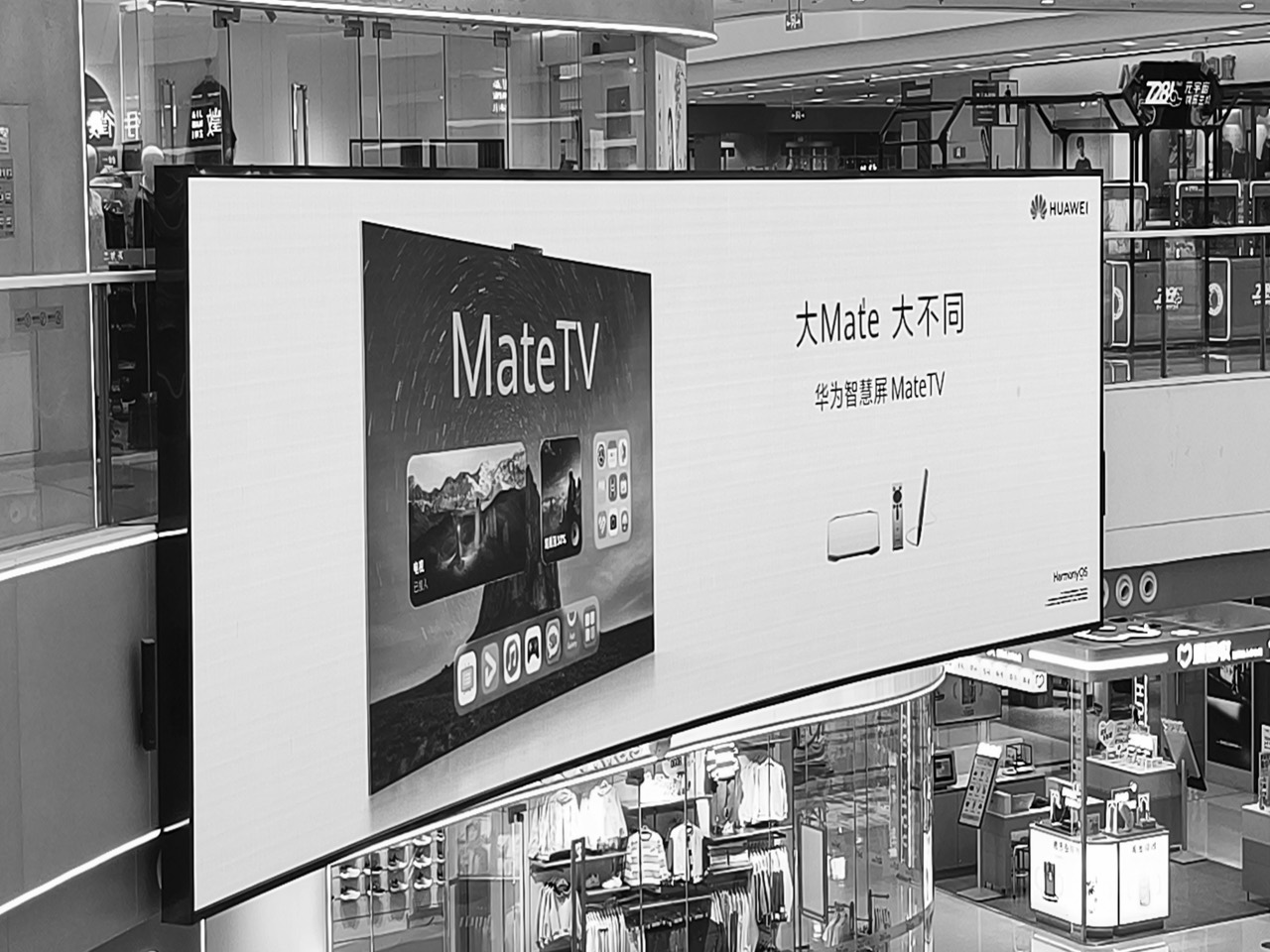} &
    \includegraphics[width=\linewidth]{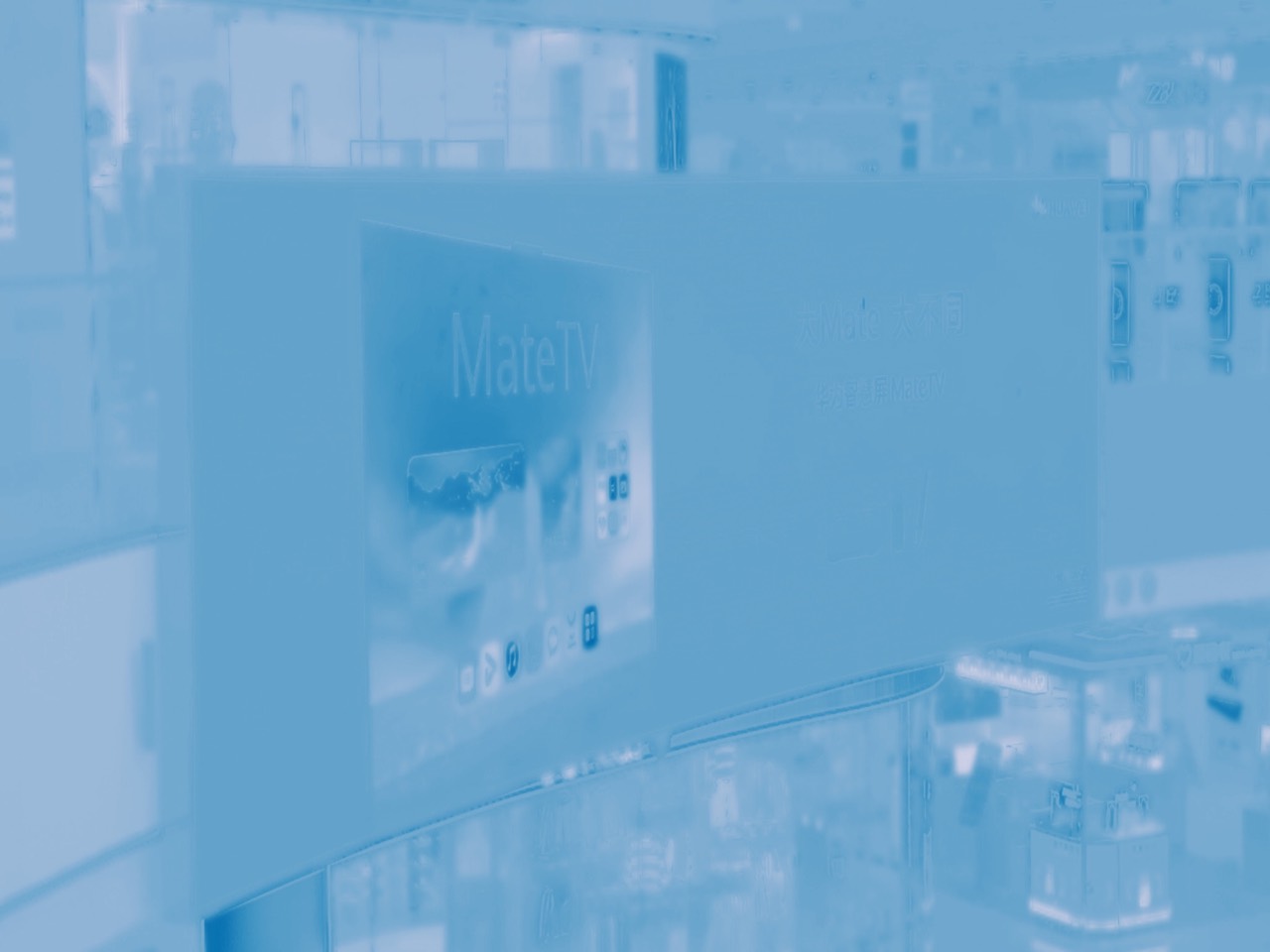} &
    \includegraphics[width=\linewidth]{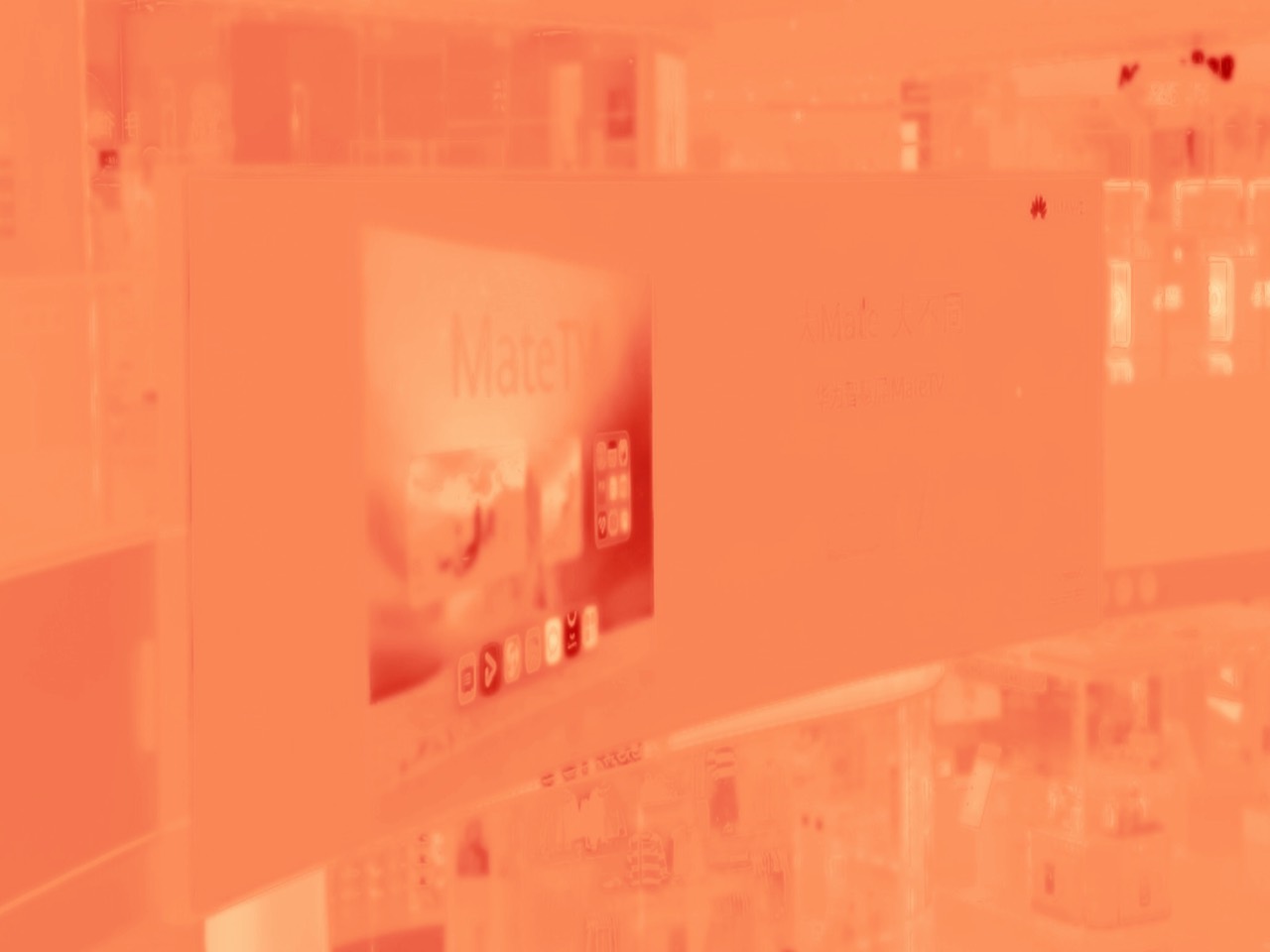} \\
    \small LQ &
    \includegraphics[width=\linewidth]{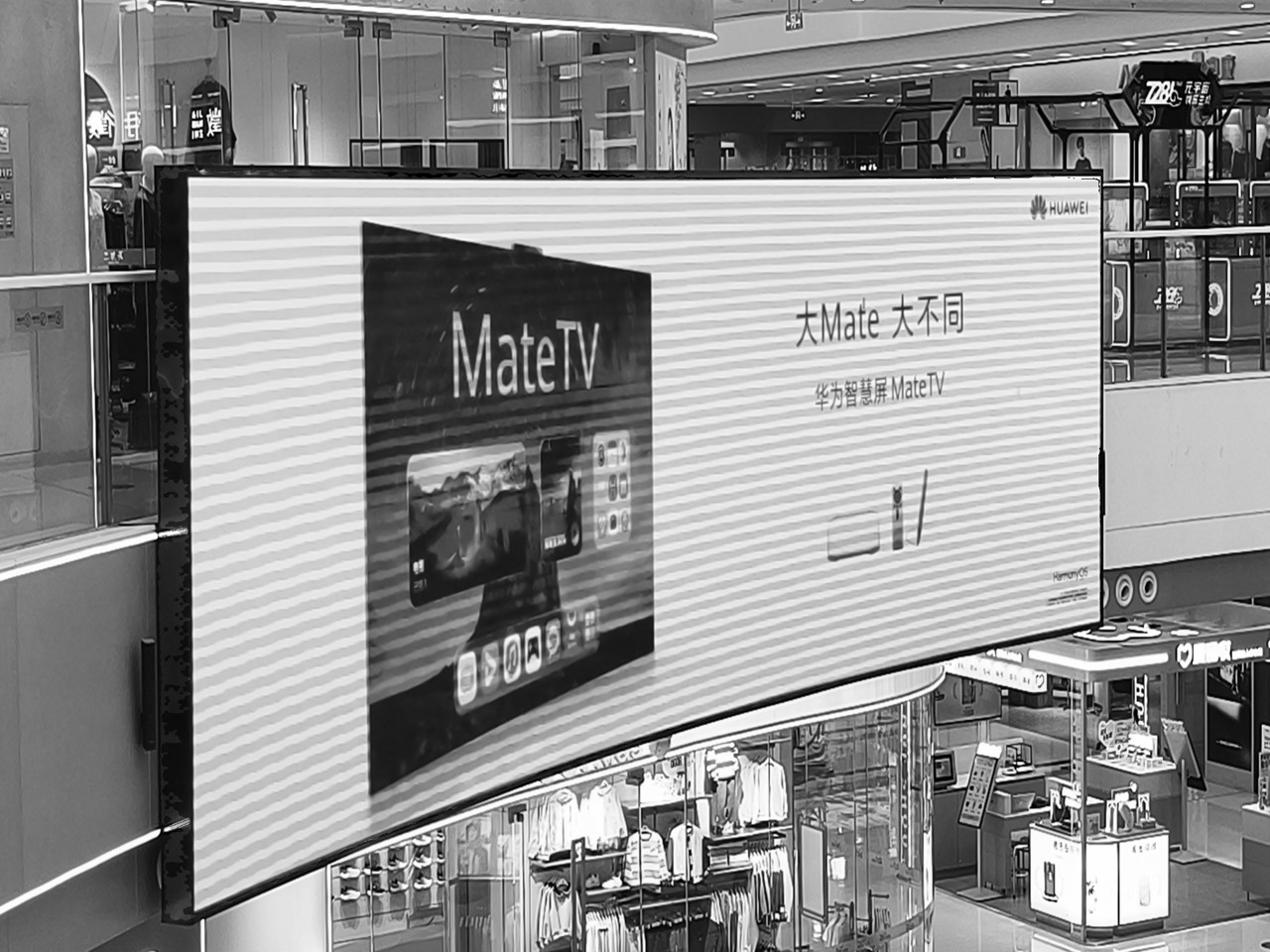} &
    \includegraphics[width=\linewidth]{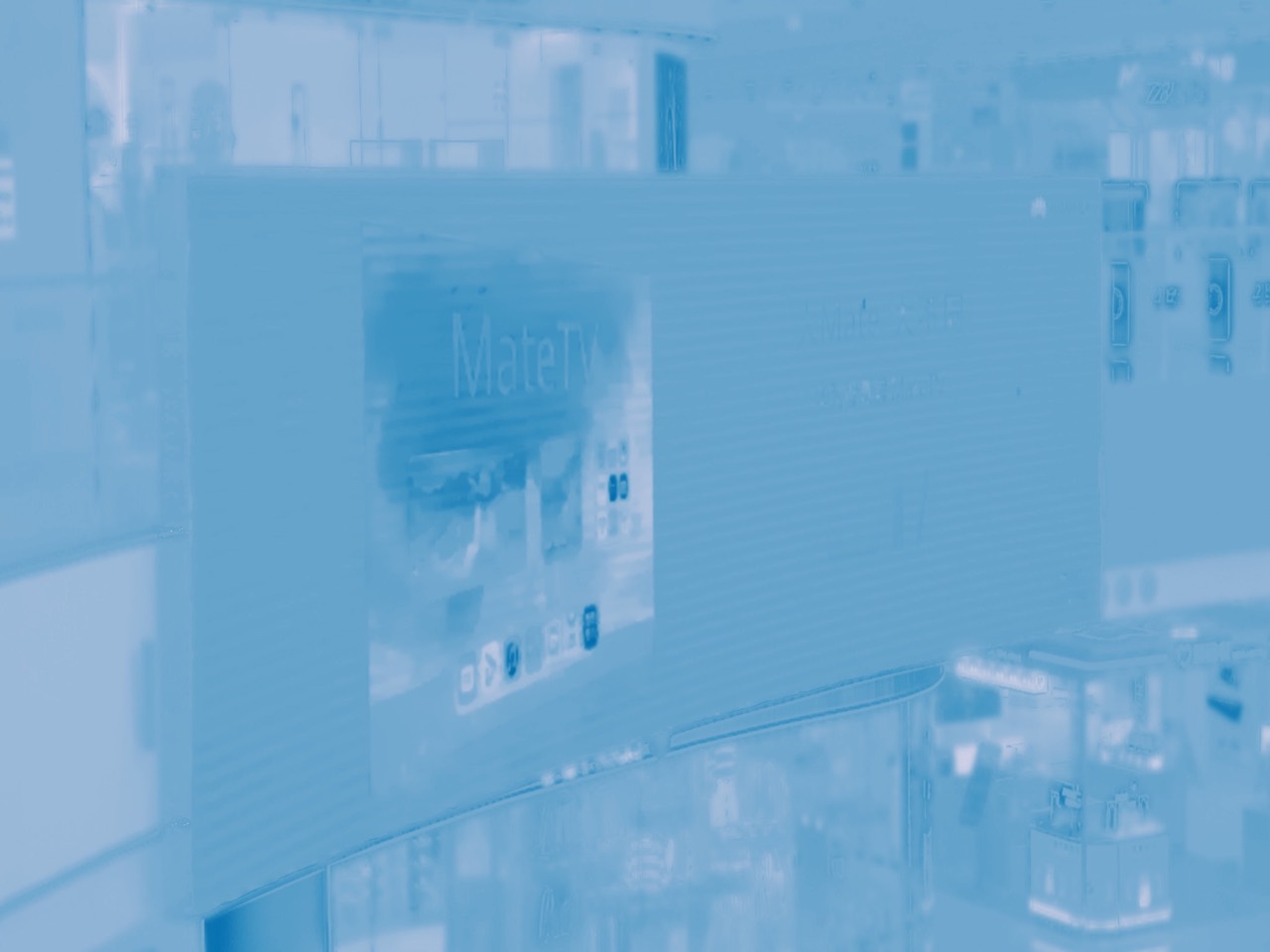} &
    \includegraphics[width=\linewidth]{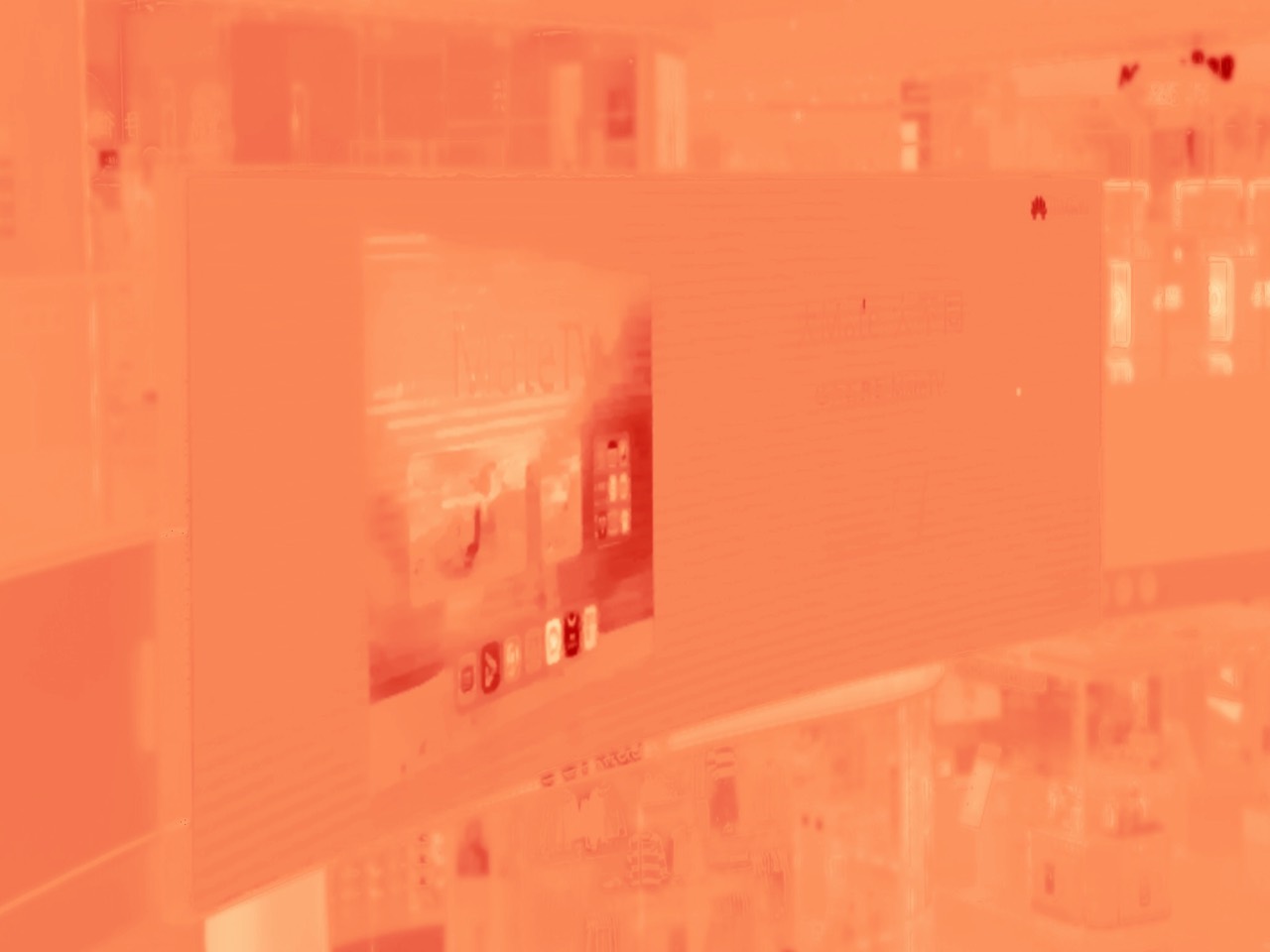} \\
    & \small Y & \small $\text{C}_{\text{b}}$ & \small $\text{C}_{\text{r}}$ \\
  \end{tabular}

  \caption{Per-channel visualization of banding-free images (GT) and flicker-banding images (LQ) in $\text{Y}\text{C}_{\text{b}}\text{C}_{\text{r}}$ color space. Dark flicker-banding is concentrated mainly in Y, with substantially weaker effects in the chrominance channels.}
  \label{fig:YCbCr visualization}
\end{figure*}

Figures~\ref{fig:RGB visualization} and~\ref{fig:YCbCr visualization} compare GT and LQ images in RGB and YCbCr. Real FB is visible in all three RGB channels, but this does not imply that an identical stripe mask should be applied independently to R, G, and B. Under RGB--YCbCr conversion, modifying Y while preserving Cb/Cr changes all three reconstructed RGB channels. The luminance relation constrains the effective RGB masks to be channel-dependent rather than identical. Thus, the RGB observations and our Y-only simulation are consistent.

For the dark FB considered here, the chrominance channels Cb and Cr are much closer to GT, while Y exhibits pronounced banding. We therefore apply the banding mask to Y. The channel comparison supports this choice, and the simulation comparison below shows that applying the same mask independently to RGB yields no measurable improvement on real-world data, whereas Y-only simulation improves restoration. This evidence concerns the supported dark/luminance FB and does not establish coverage of chromatic banding.

\begin{figure*}[t]
  \centering
  \setlength{\tabcolsep}{2pt}
  \renewcommand{\arraystretch}{1.0}
  \begin{tabular}{@{}m{0.03\linewidth}*{2}{>{\centering\arraybackslash}m{0.37\linewidth}}@{}}
    \centering\small GT &
    \includegraphics[width=\linewidth]{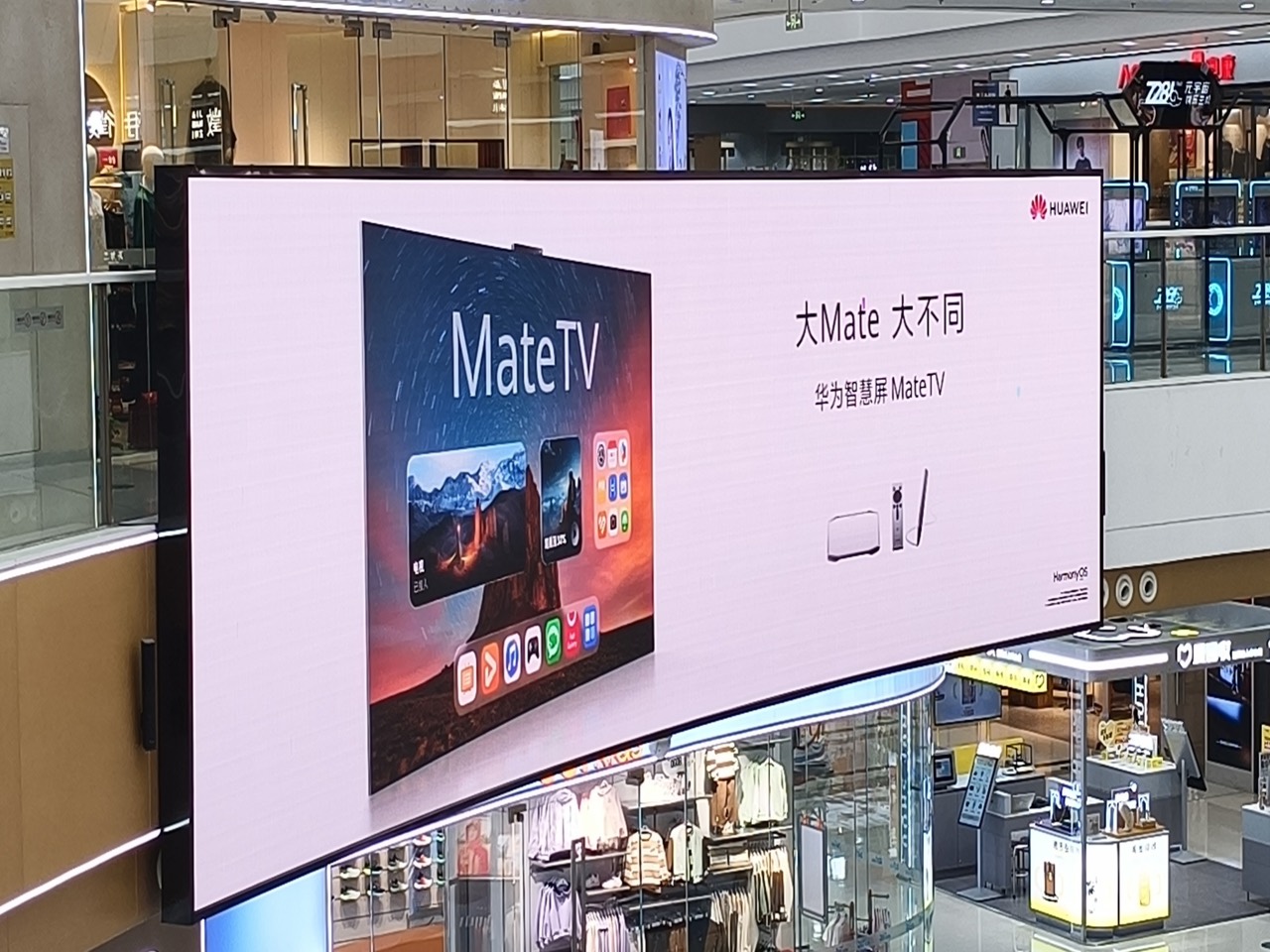} &
    \includegraphics[width=\linewidth]{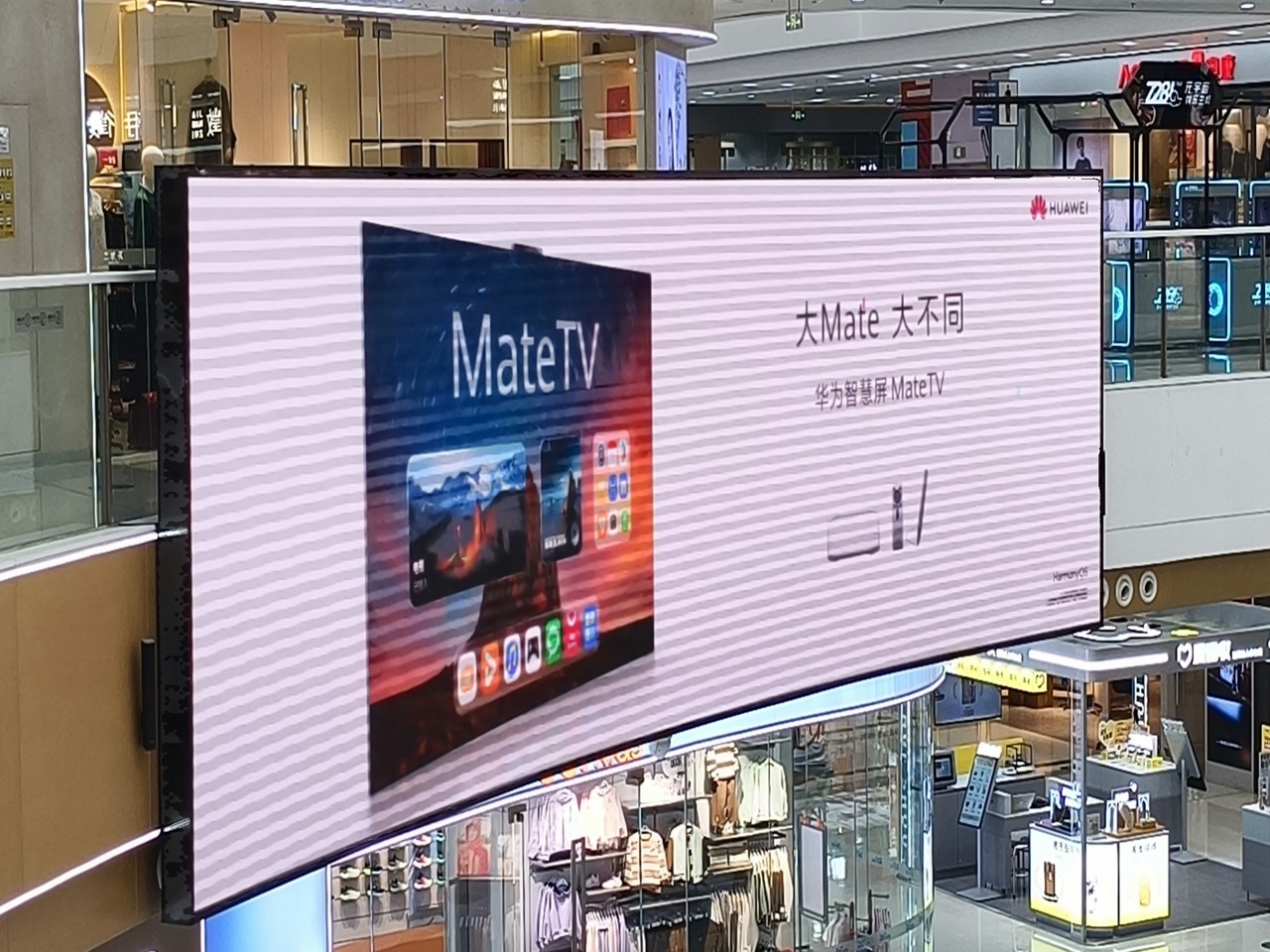} \\[1mm]
    \centering\small LQ &
    \includegraphics[width=\linewidth]{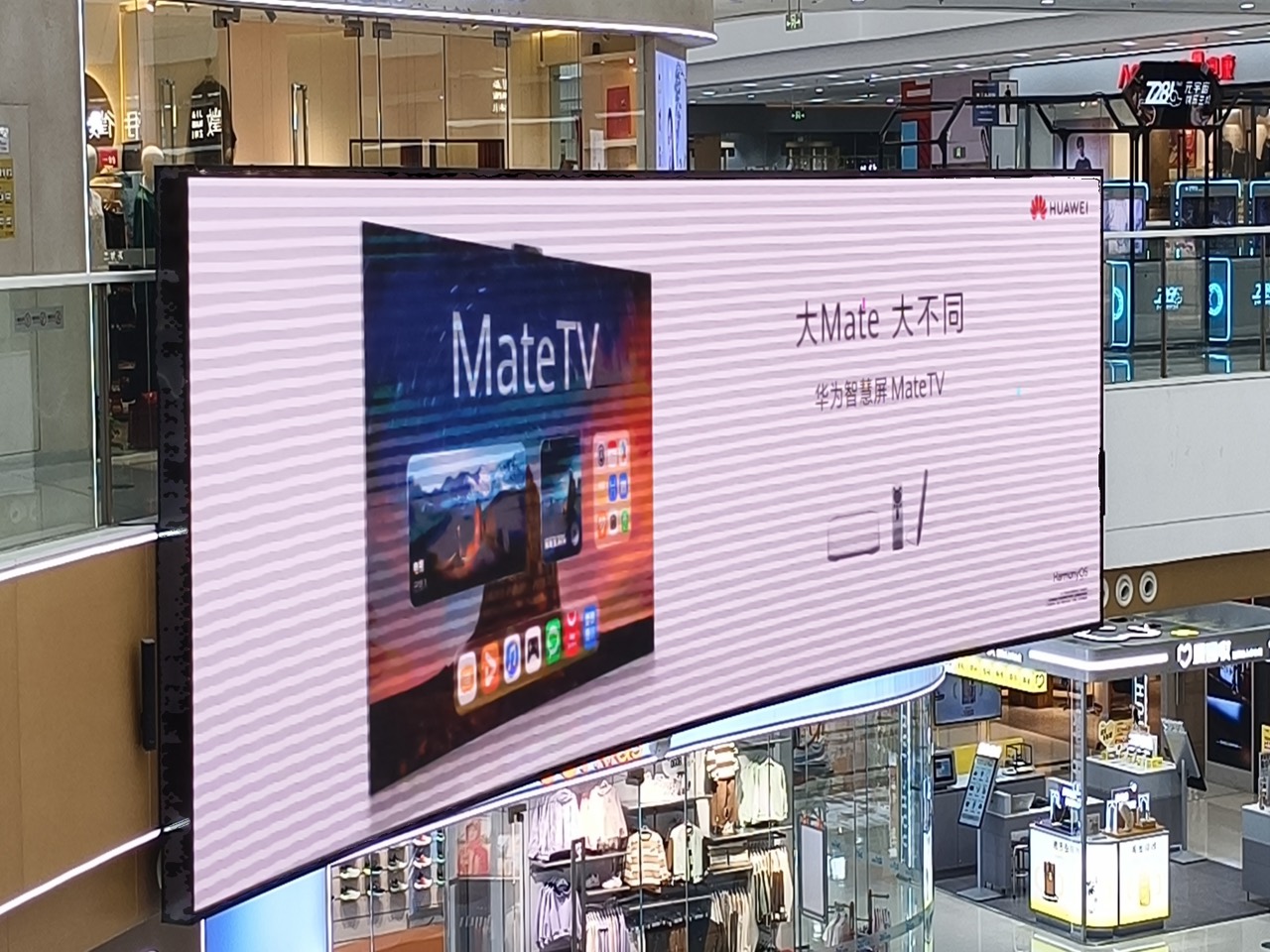} &
    \includegraphics[width=\linewidth]{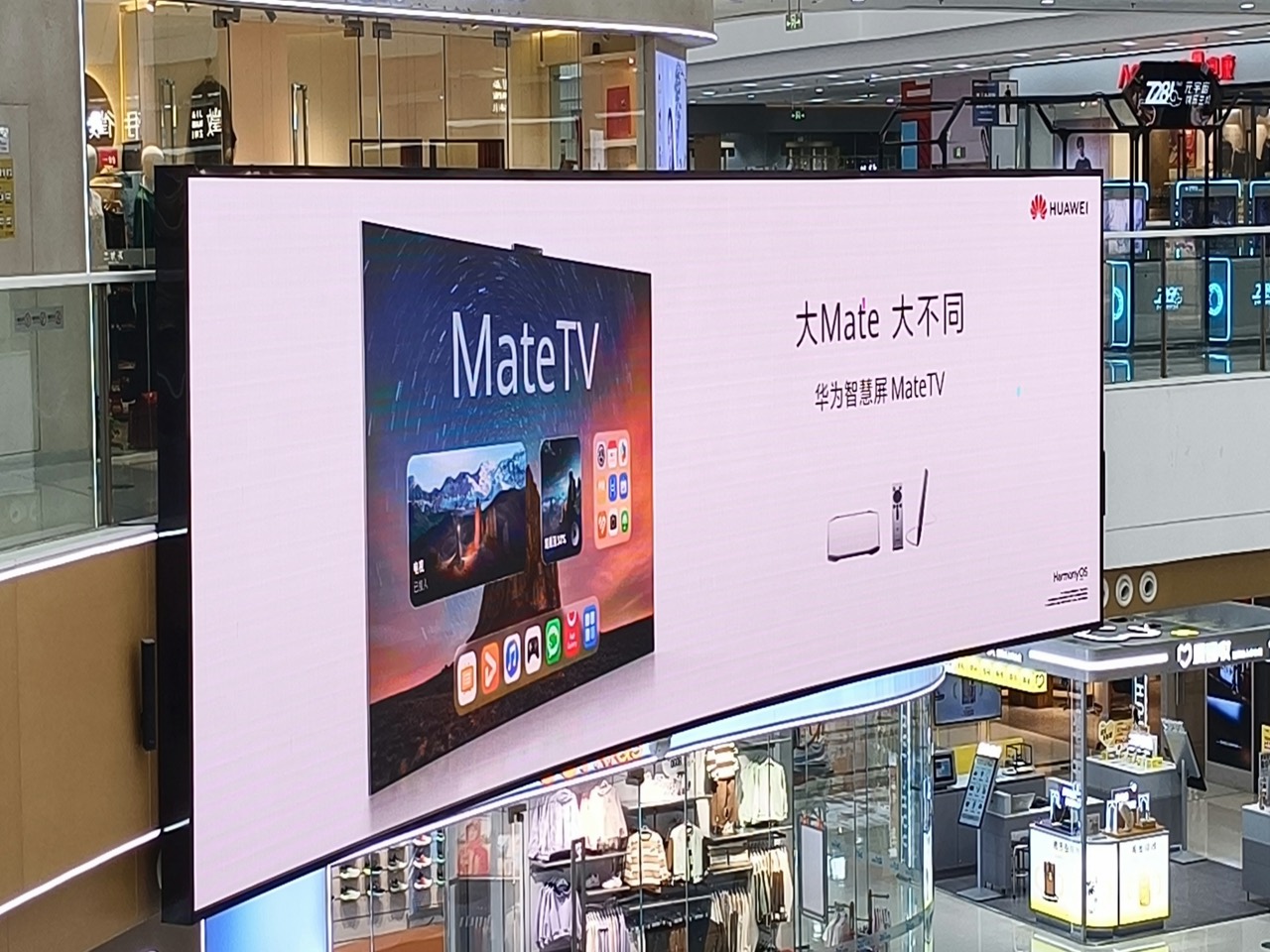} \\[1mm]
    & \small Origin & \small Swapped \\
  \end{tabular}

  \caption{Visualization comparison of Y-channel switching effects. \emph{Origin} shows the original image and \emph{Swapped} shows the image after
exchanging its Y channel with the counterpart.}
  \label{fig:swapped visualization}
\end{figure*}

\begin{figure*}[t]
  \centering
  \setlength{\tabcolsep}{2pt}      
  \renewcommand{\arraystretch}{1.0}

  \begin{tabular}{@{}>{\centering\arraybackslash}m{0.22\linewidth}*{4}{>{\centering\arraybackslash}m{0.11\linewidth}}@{}}
      \includegraphics[width=\linewidth]{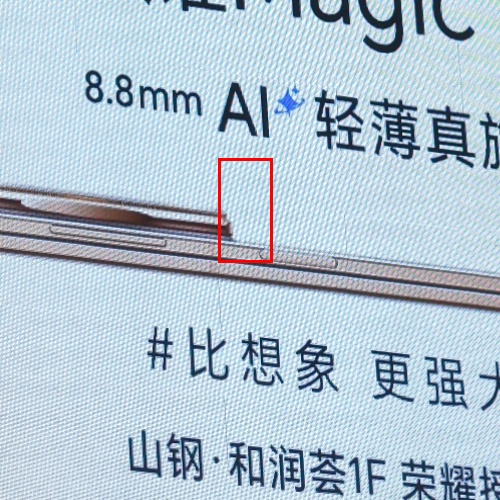} &
      \includegraphics[width=\linewidth]{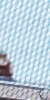} &
      \includegraphics[width=\linewidth]{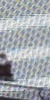} &
      \includegraphics[width=\linewidth]{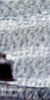} &
      \includegraphics[width=\linewidth]{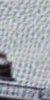} \\
      \includegraphics[width=\linewidth]{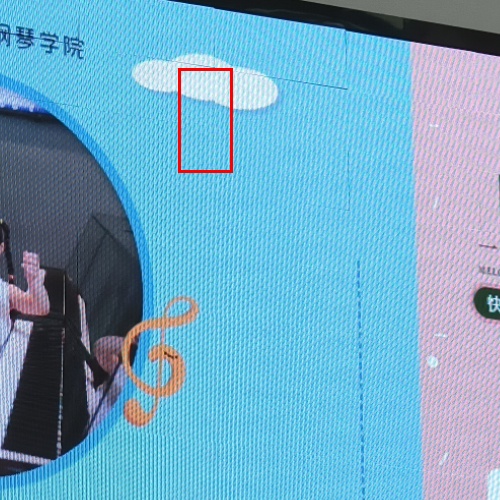} &
      \includegraphics[width=\linewidth]{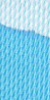} &
      \includegraphics[width=\linewidth]{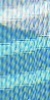} &
      \includegraphics[width=\linewidth]{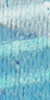} &
      \includegraphics[width=\linewidth]{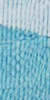} \\
      [-0.5mm]
      \small Real-Flicker & \small GT & \small LQ & \small RGB & \small $\text{Y}\text{C}_{\text{b}}\text{C}_{\text{r}}$
    \end{tabular}%
  \caption{Visual comparison of RIFLE on synthetic datasets based on RGB/$\text{Y}\text{C}_{\text{b}}\text{C}_{\text{r}}$ color space.}
  \label{fig:Simulation experiment results}
\end{figure*}

To further validate this observation, we swapped the Y channels between GT and LQ while keeping their respective $\text{C}{\text{b}}$ and $\text{C}{\text{r}}$ components unchanged. After recomposition (Fig.~\ref{fig:swapped visualization}, the LQ image no longer displays visible banding, whereas the GT image exhibits clear flicker-banding. These results confirm that the flicker-banding phenomenon originates primarily in the luminance component (Y) and exerts negligible influence on the chrominance components ($\text{C}{\text{b}}$ and $\text{C}{\text{r}}$).

Building on this luminance-centric insight, we conducted the experiments in Fig.~\ref{fig:Simulation experiment results}. When a synthetic training set was constructed by superimposing the same mask independently on the R, G, and B channels, the resulting models showed no measurable improvement on real-world data. This outcome underscores that simply applying identical per-channel masks in RGB space is not a faithful approximation of the real phenomenon and introduces a substantial domain gap relative to real-world images. Consequently, a fundamental principle for simulating banding in RGB is that one should avoid superimposing an identical mask across all three channels. Instead, channel-dependent masks must be designed to preserve chrominance while modulating only the luminance (Y).

\subsection{Limitations and Areas for Improvement.}
\label{app:limitations}

\begin{figure*}[t]
  \centering
  \setlength{\tabcolsep}{2pt}
  \renewcommand{\arraystretch}{1.0}
  \begin{tabular}{@{}*{3}{>{\centering\arraybackslash}m{0.3\linewidth}}@{}}
    \includegraphics[width=\linewidth]{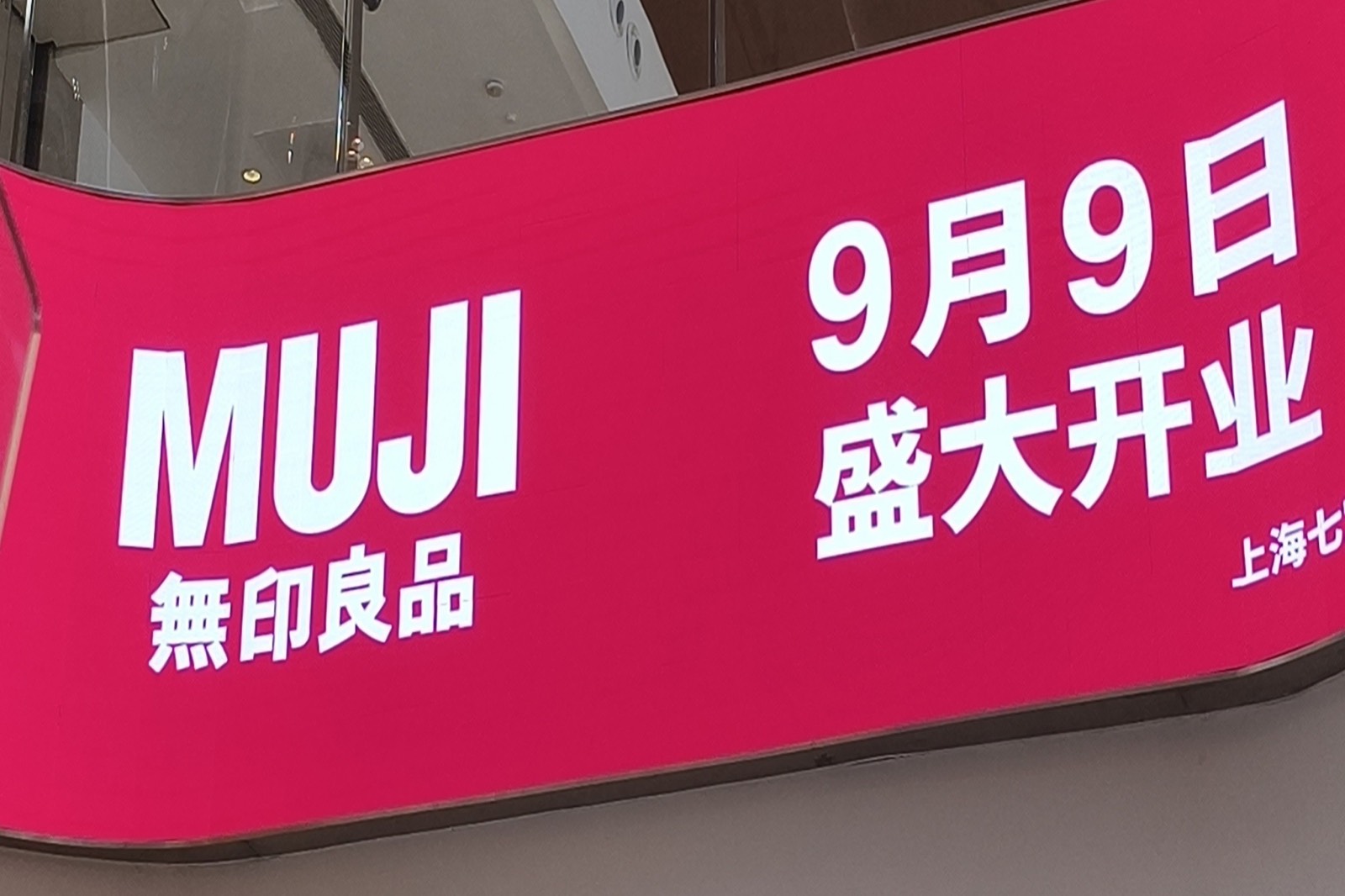} &
    \includegraphics[width=\linewidth]{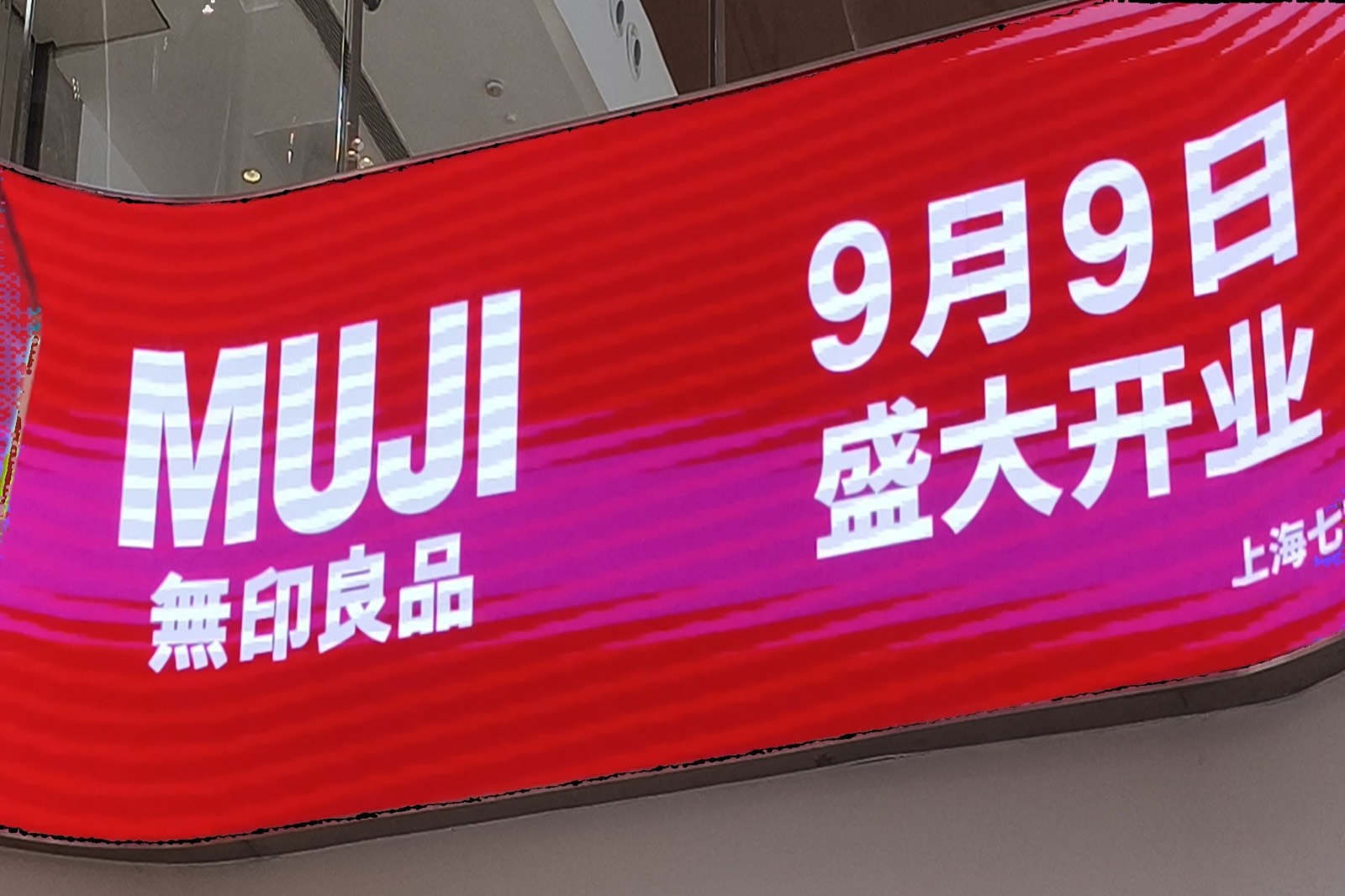} &
    \includegraphics[width=\linewidth]{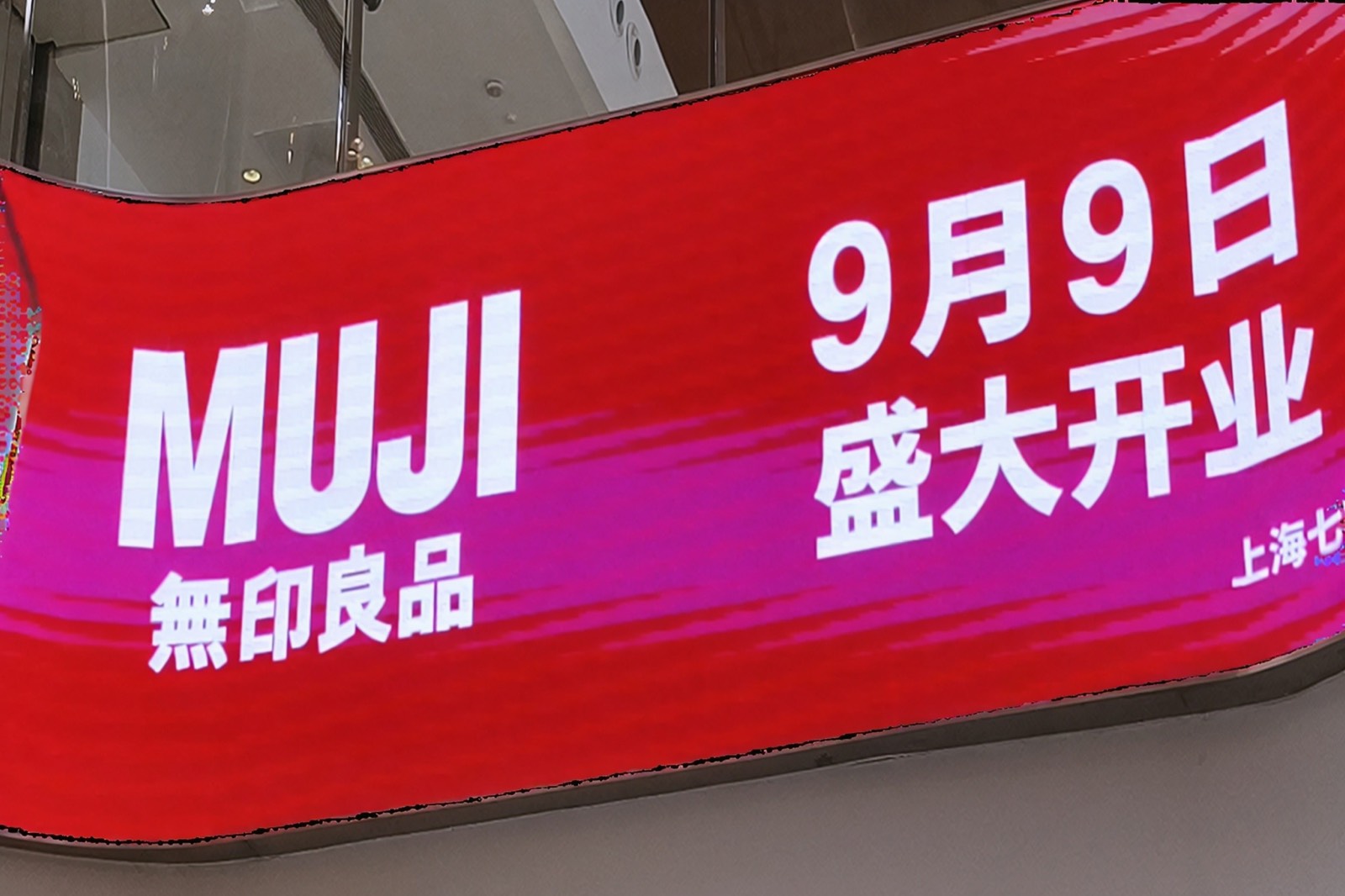} \\
    \small GT & \small LQ & \small RIFLE \\
  \end{tabular}

  \caption{A case where our method leaves residual magenta colored bands.}
  \label{fig:failure-color}
\end{figure*}

Although our proposed method achieves decent performance in flicker-banding removal, there are still some limitations and areas for improvement:

\textbf{Residual Colored Bands.} Some input images may contain extra colored stripes caused by certain display dimming methods. Our model is designed mainly for gray-scale flicker and can leave these colored components visible, as shown in Fig.~\ref{fig:failure-color}.

This limitation arises from the training simulator, which focuses on luminance modulation. Future work will expand the synthetic dataset to cover a wider range of banding types.

\begin{figure*}[t]
  \centering
  \setlength{\tabcolsep}{2pt}
  \renewcommand{\arraystretch}{1.0}
  \begin{minipage}[t]{0.55\linewidth}
    \centering
    \begin{tabular}{@{}*{3}{>{\centering\arraybackslash}m{0.31\linewidth}}@{}}
      \includegraphics[width=\linewidth]{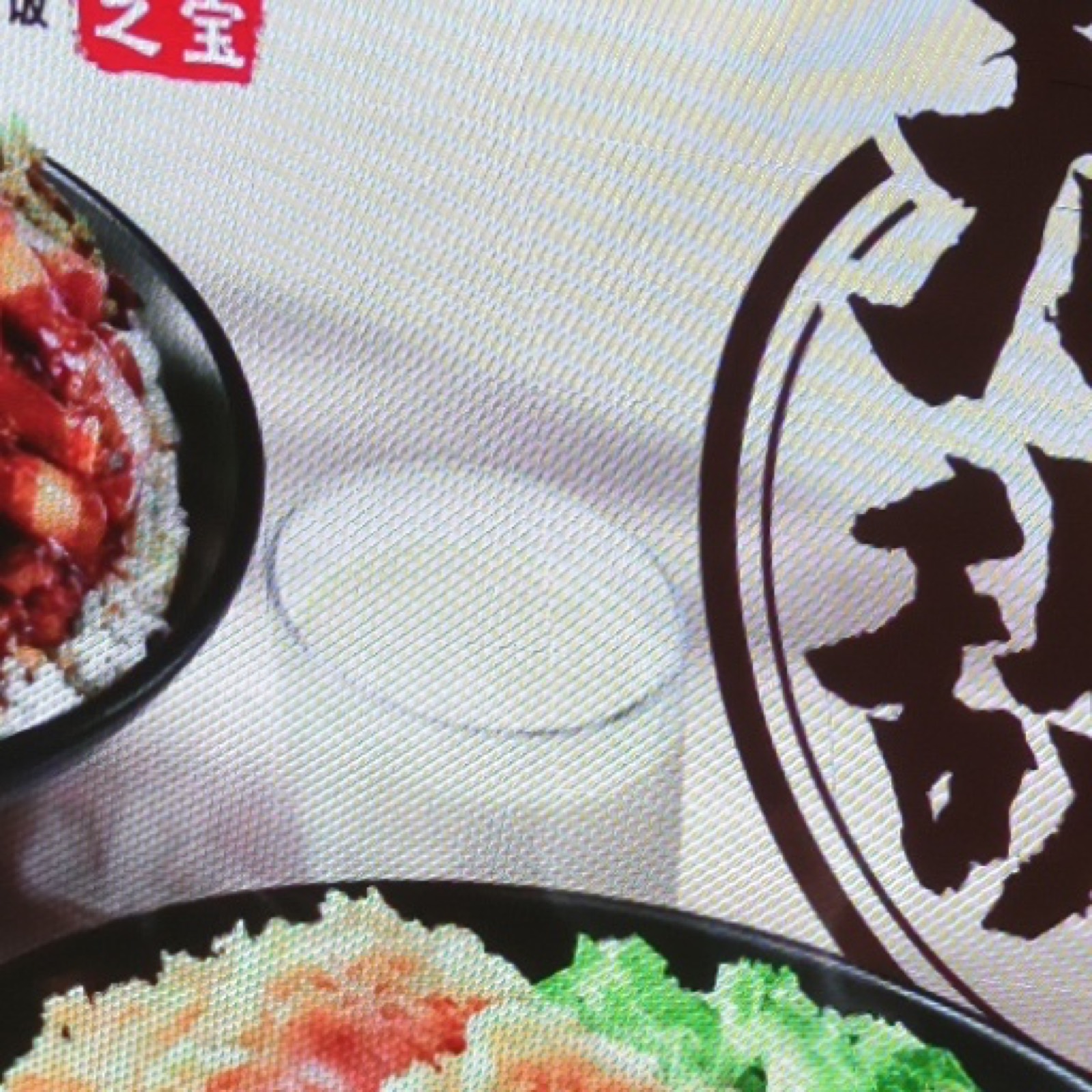} &
      \includegraphics[width=\linewidth]{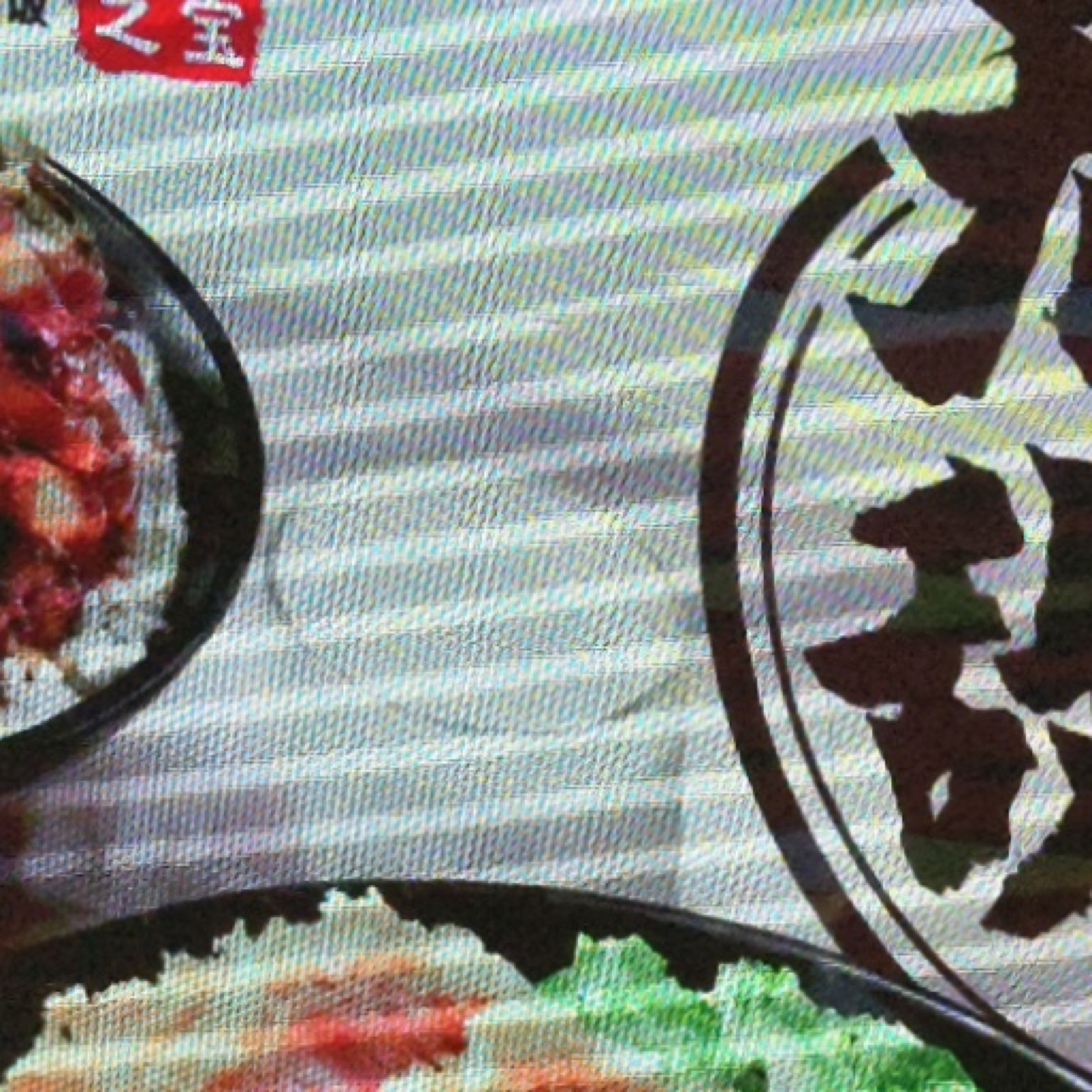} &
      \includegraphics[width=\linewidth]{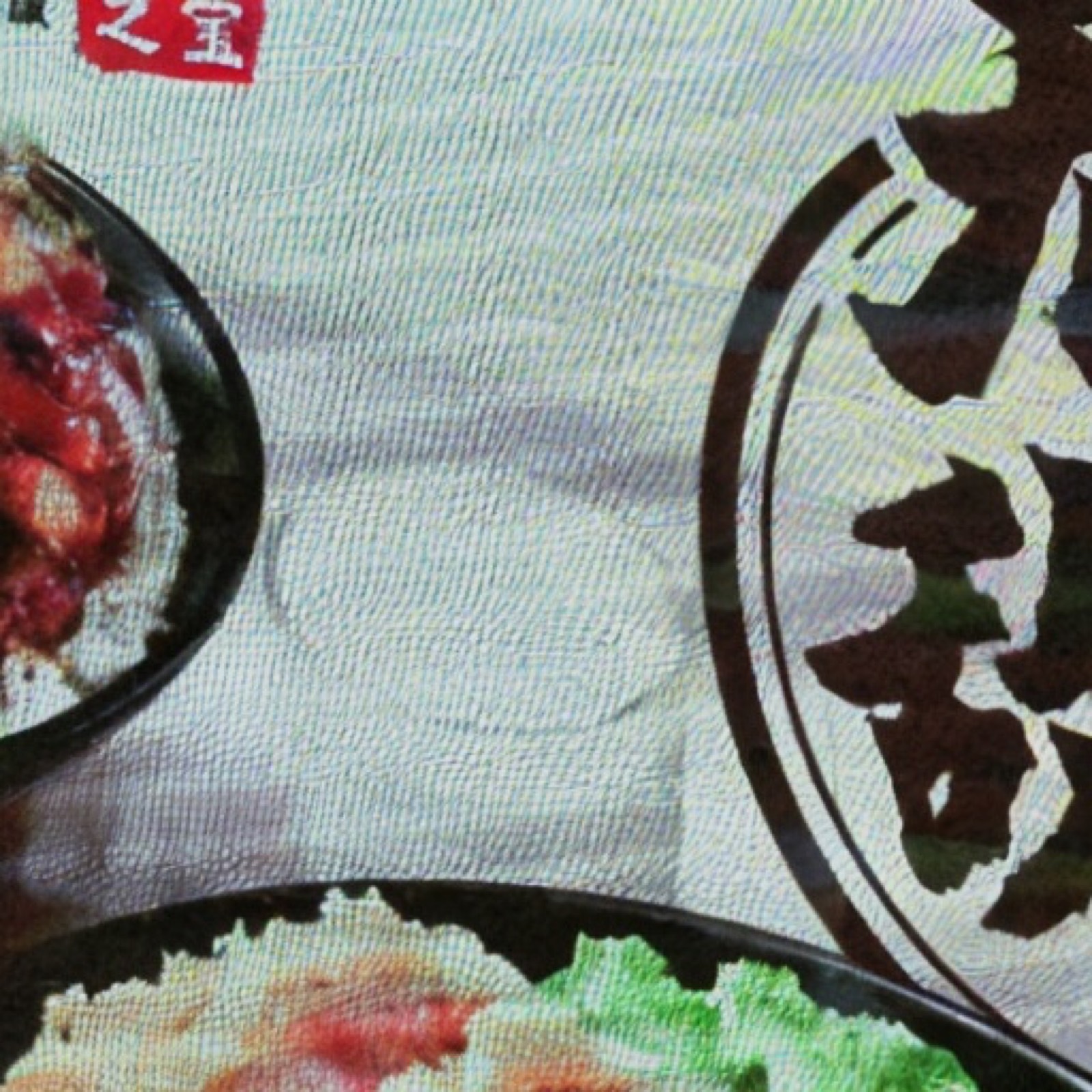} \\
      \small GT & \small LQ & \small RIFLE \\
    \end{tabular}
    \captionof{figure}{RIFLE alters fine texture details.}
    \label{fig:failure-texture}
  \end{minipage}\hfill
  \begin{minipage}[t]{0.38\linewidth}
    \centering
    \begin{tabular}{@{}*{2}{>{\centering\arraybackslash}m{0.46\linewidth}}@{}}
      \includegraphics[width=\linewidth]{iclr/figs/failure/extreme_lq.jpg} &
      \includegraphics[width=\linewidth]{iclr/figs/failure/extreme_our.jpg} \\
      \small LQ & \small RIFLE \\
    \end{tabular}
    \captionof{figure}{RIFLE struggles with extremely complex banding patterns.}
    \label{fig:failure-complex}
  \end{minipage}
\end{figure*}

\textbf{Texture Detail Loss.} Despite using one-step restoration, the generative diffusion prior can alter original textures or introduce fine details that were not present in the source. As shown in Fig.~\ref{fig:failure-texture}, additional details are generated when removing FB.

This is a common challenge in diffusion-based restoration methods. Future improvements could be made by combining diffusion with other techniques to protect true scene details.

\textbf{Text-Sensitive Applications.} RIFLE can improve visual quality, but character-level fidelity is not guaranteed. The OCR evaluation in Appendix~\ref{app:ocr-fidelity} shows better text fidelity than PiSA-SR but lower fidelity than the original LQ input. Text-sensitive outputs should therefore be verified, and improving character-level fidelity while retaining restoration quality remains an important direction.

\textbf{Handling Severe or Complex Patterns.} While our method works well on typical flicker-banding patterns, scenes with highly non-stationary, complex, or curved banding may still pose challenges. Purely black banding regions that completely obscure underlying content can also be difficult to recover. In such cases, the model may failed to restore the lost information.

Improvements could be made by considering multi-frame information, expanding the training data by including more complex banding patterns, or physics-based modeling of display artifacts.

\subsection{Our simulated and real-world dataset visualization.}
\label{app:dataset-visualization}

\begin{figure*}[t]
  \centering
  \setlength{\tabcolsep}{2pt}
  \renewcommand{\arraystretch}{1.0}
  \begin{tabular}{cccc}
    \includegraphics[width=0.23\linewidth,height=0.1725\linewidth]{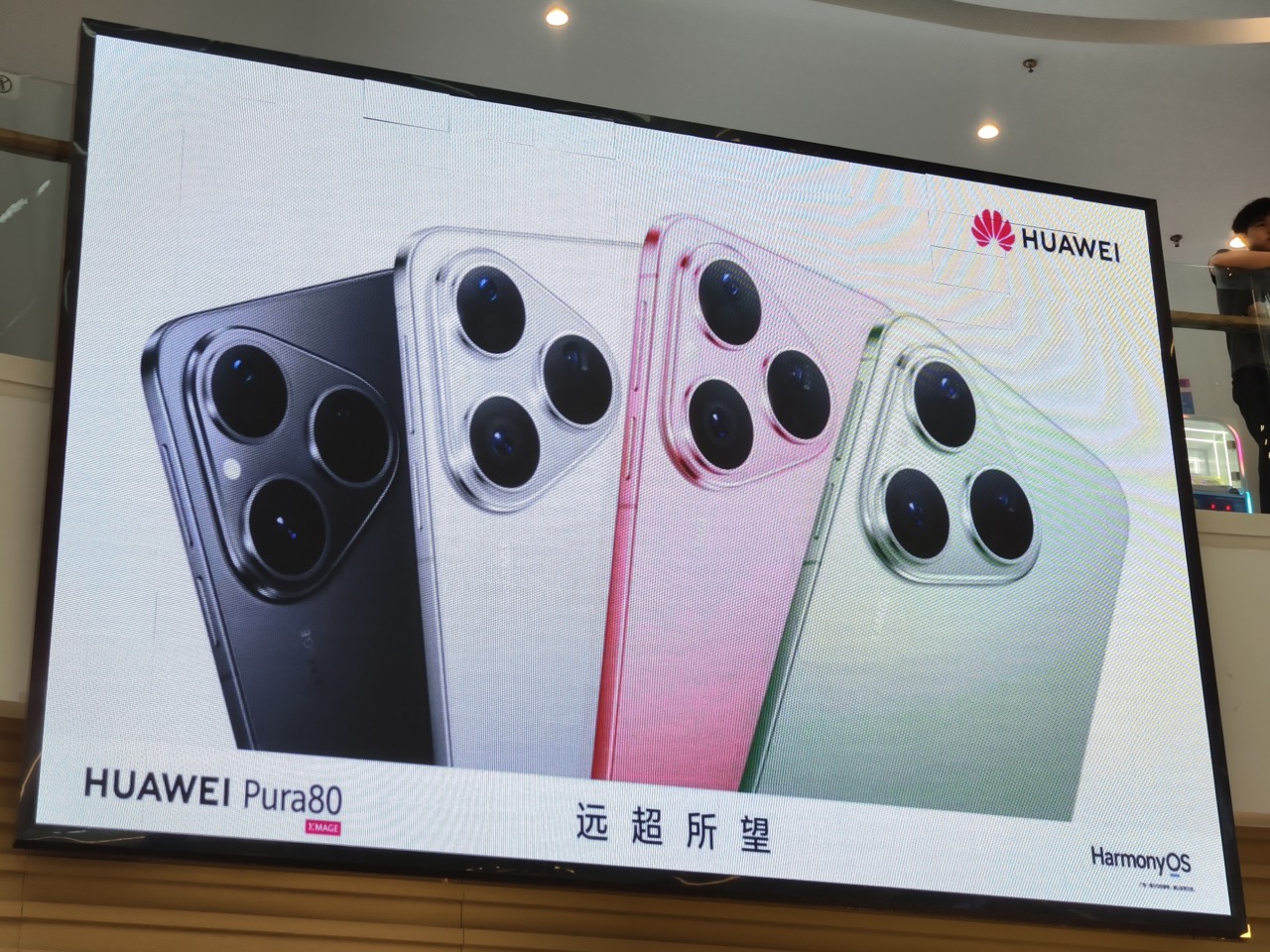} &
    \includegraphics[width=0.23\linewidth,height=0.1725\linewidth]{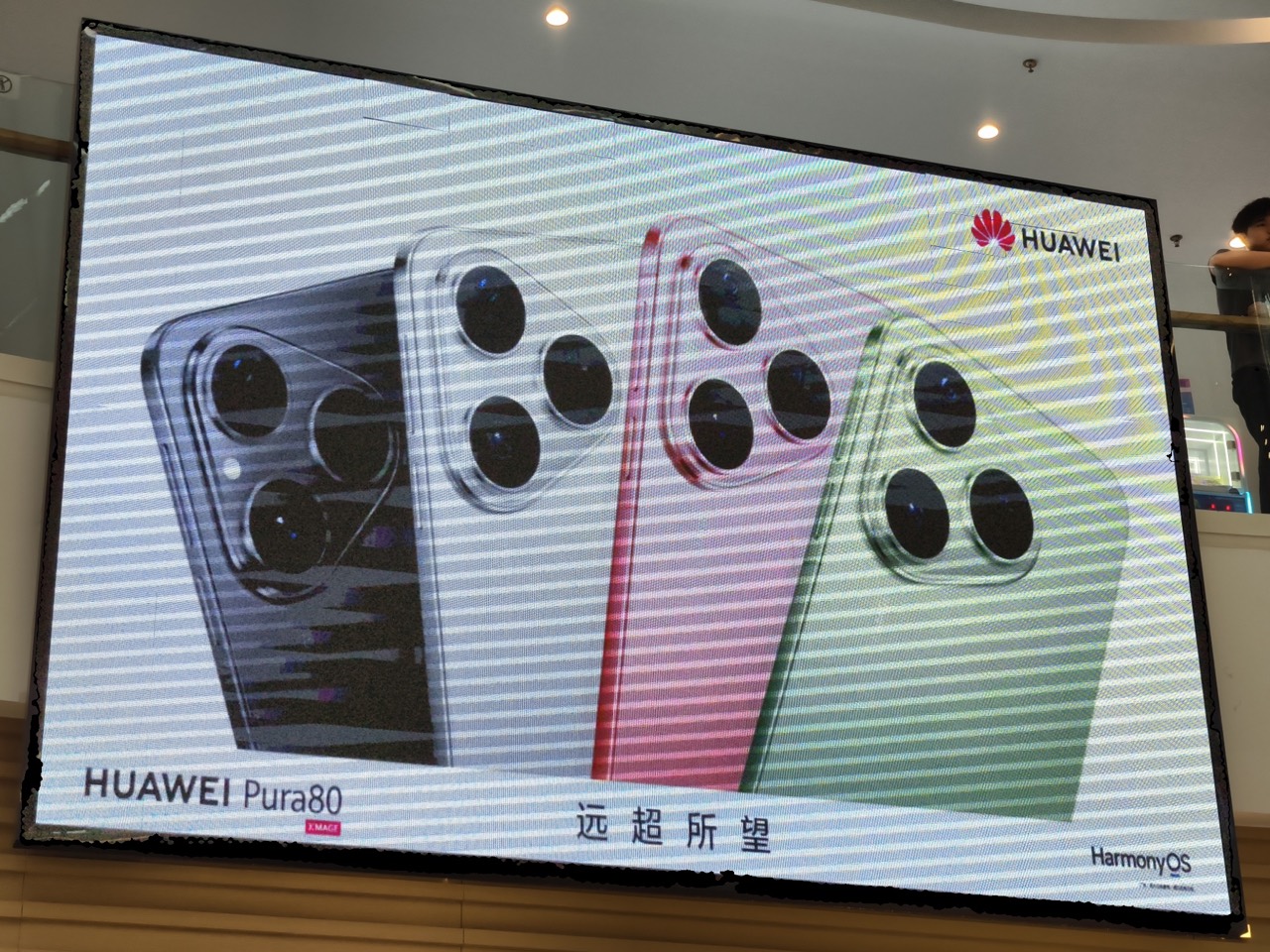} &
    \includegraphics[width=0.23\linewidth,height=0.1725\linewidth]{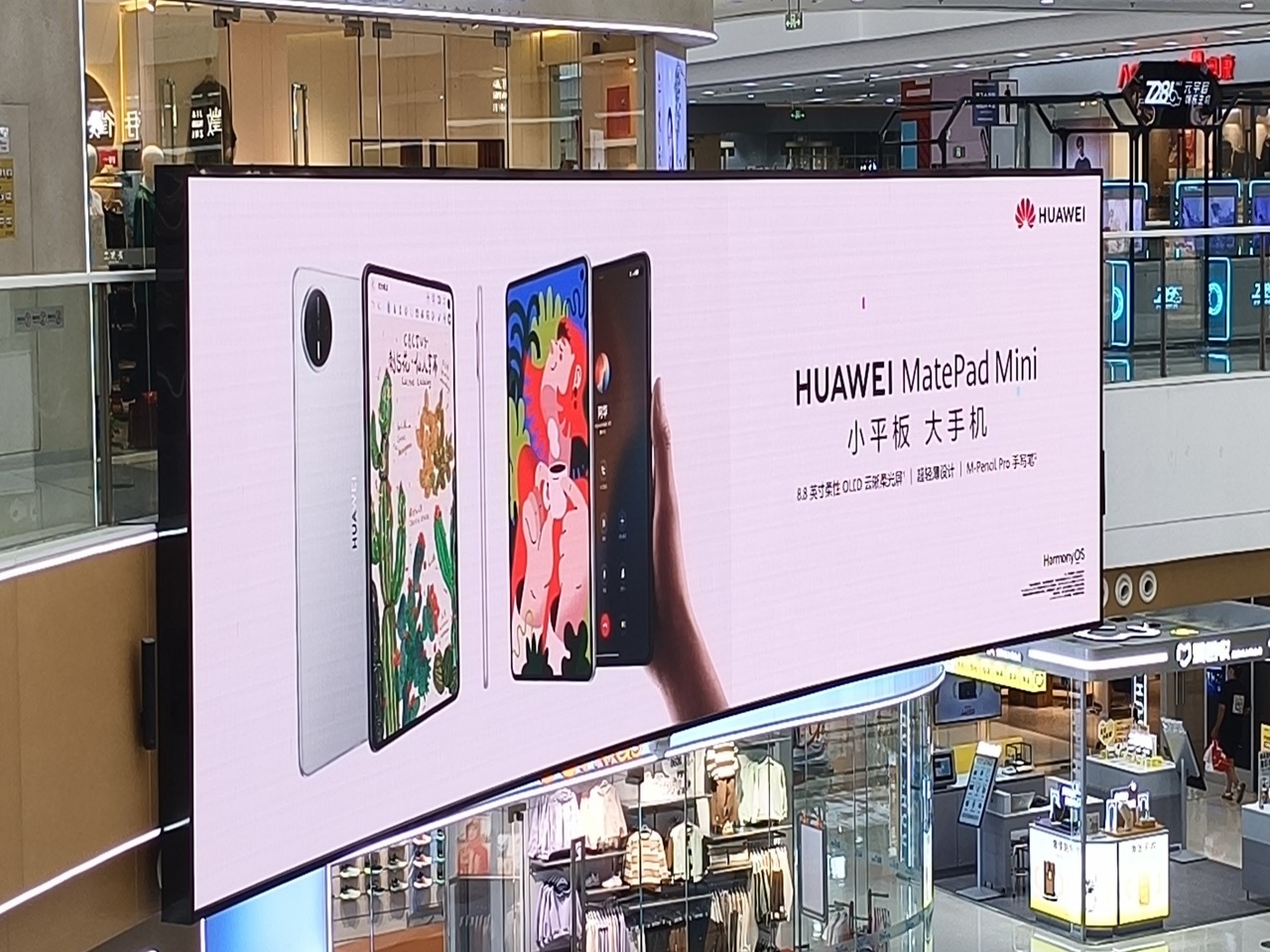} &
    \includegraphics[width=0.23\linewidth,height=0.1725\linewidth]{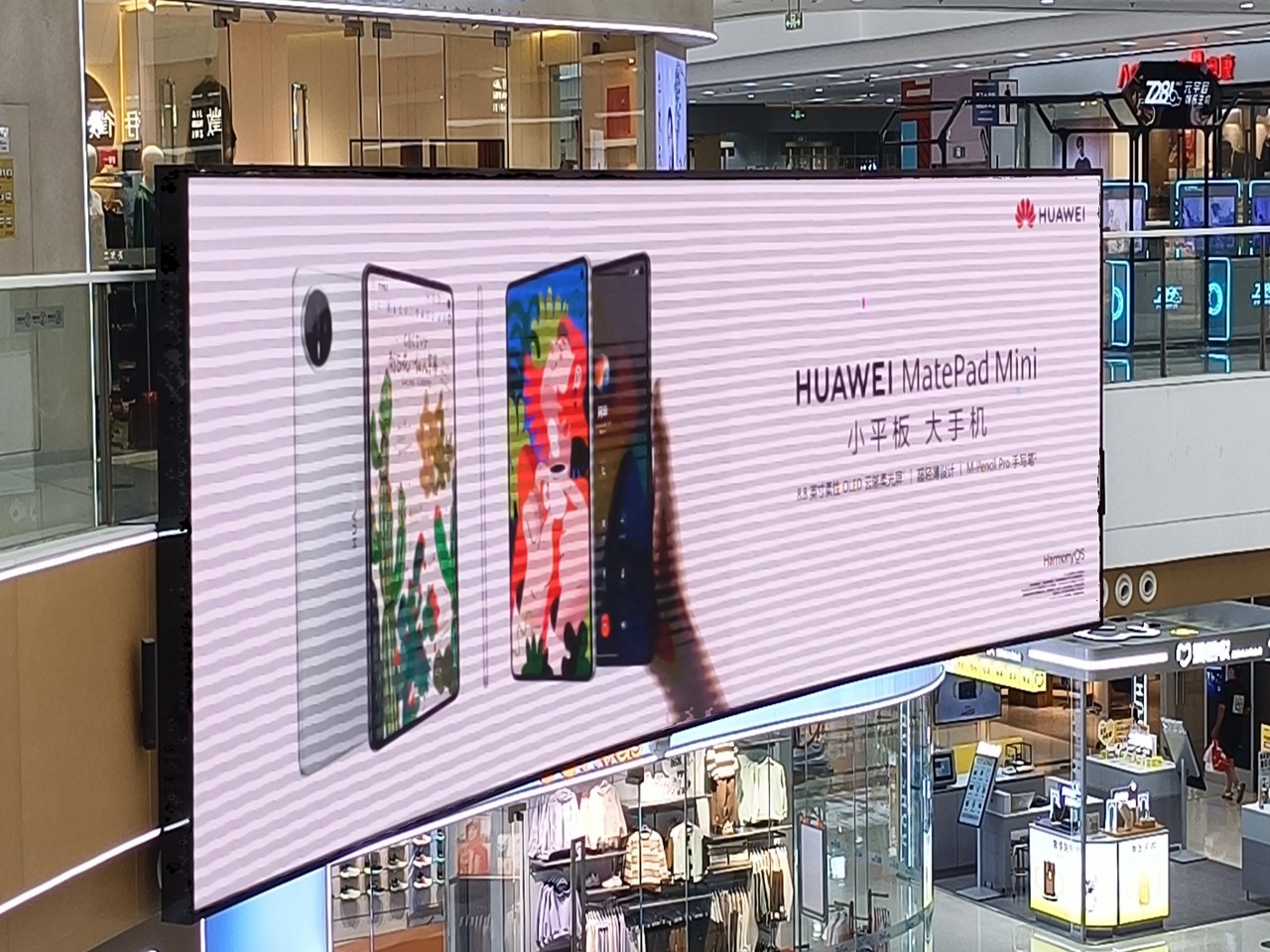} \\
    \small GT & \small LQ & \small GT & \small LQ \\
    \includegraphics[width=0.23\linewidth,height=0.1725\linewidth]{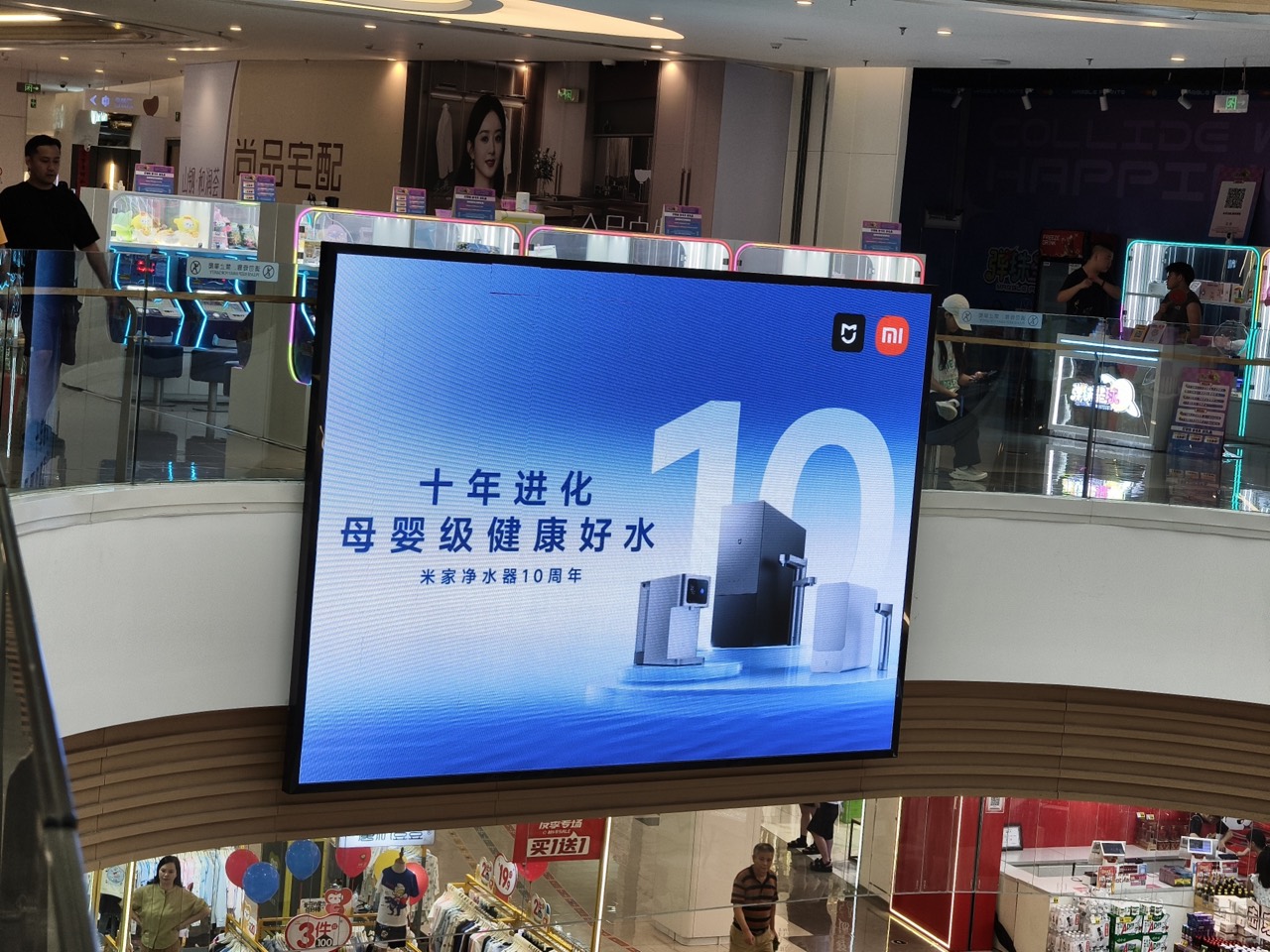} &
    \includegraphics[width=0.23\linewidth,height=0.1725\linewidth]{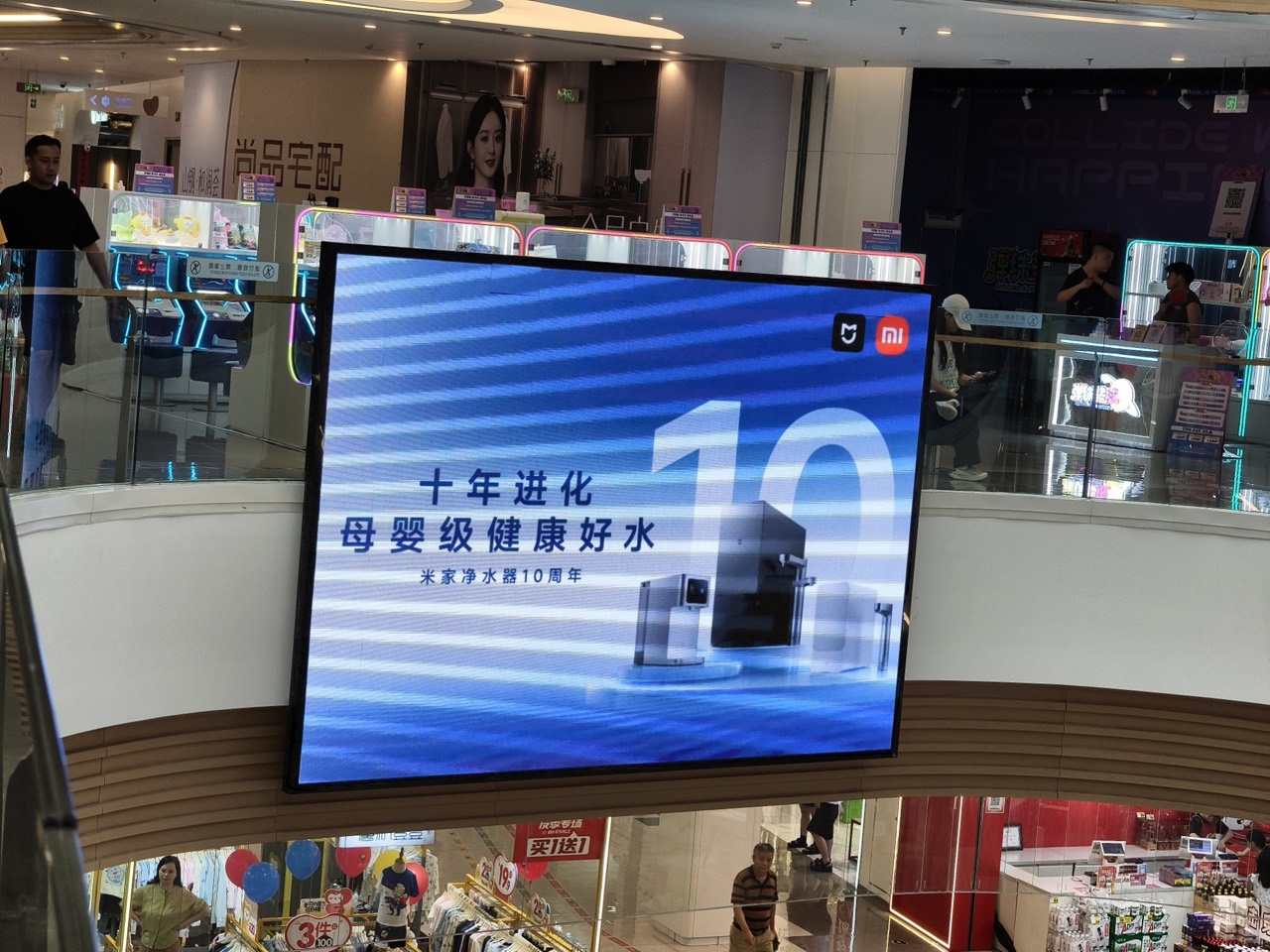} &
    \includegraphics[width=0.23\linewidth,height=0.1725\linewidth]{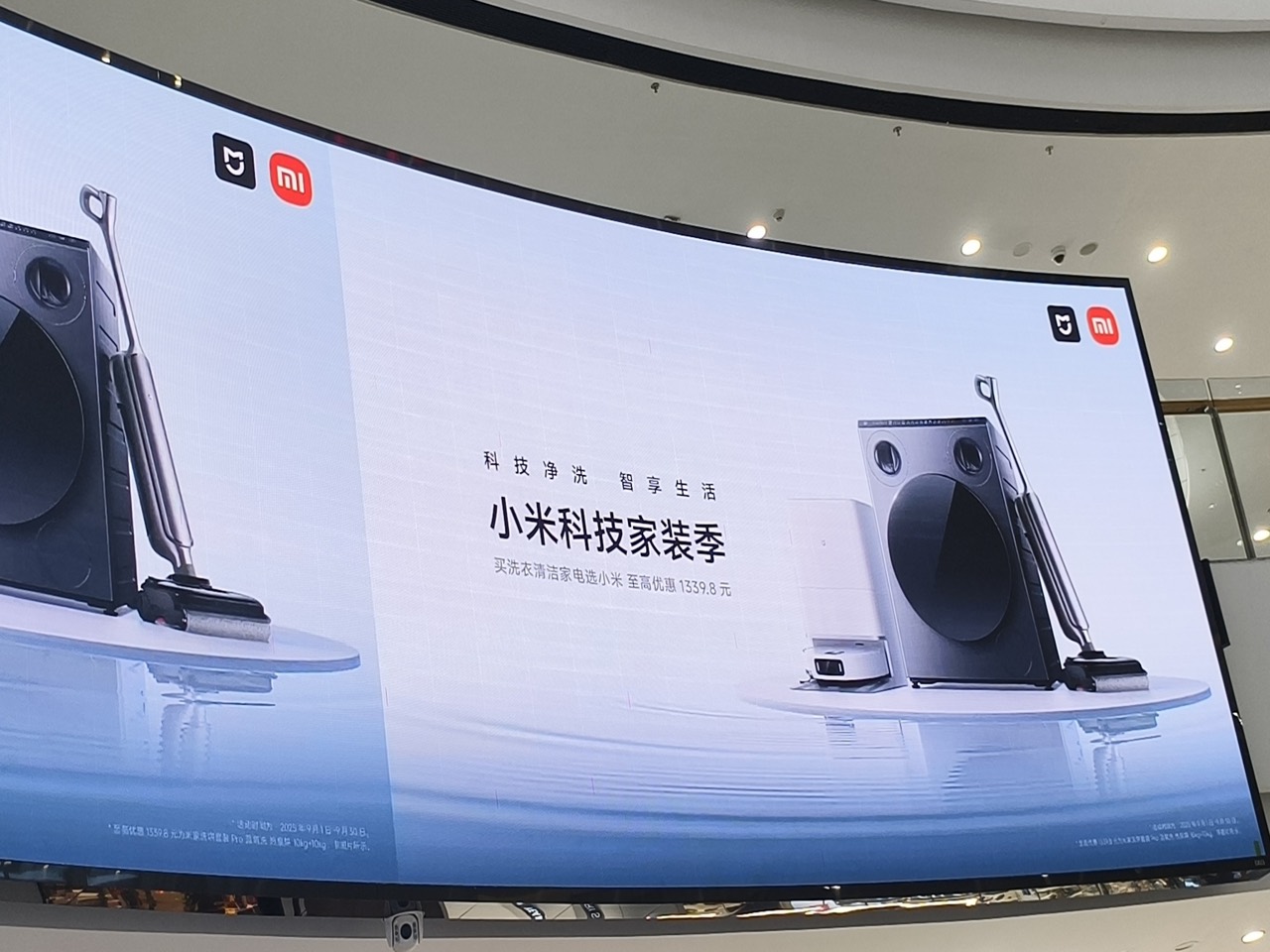} &
    \includegraphics[width=0.23\linewidth,height=0.1725\linewidth]{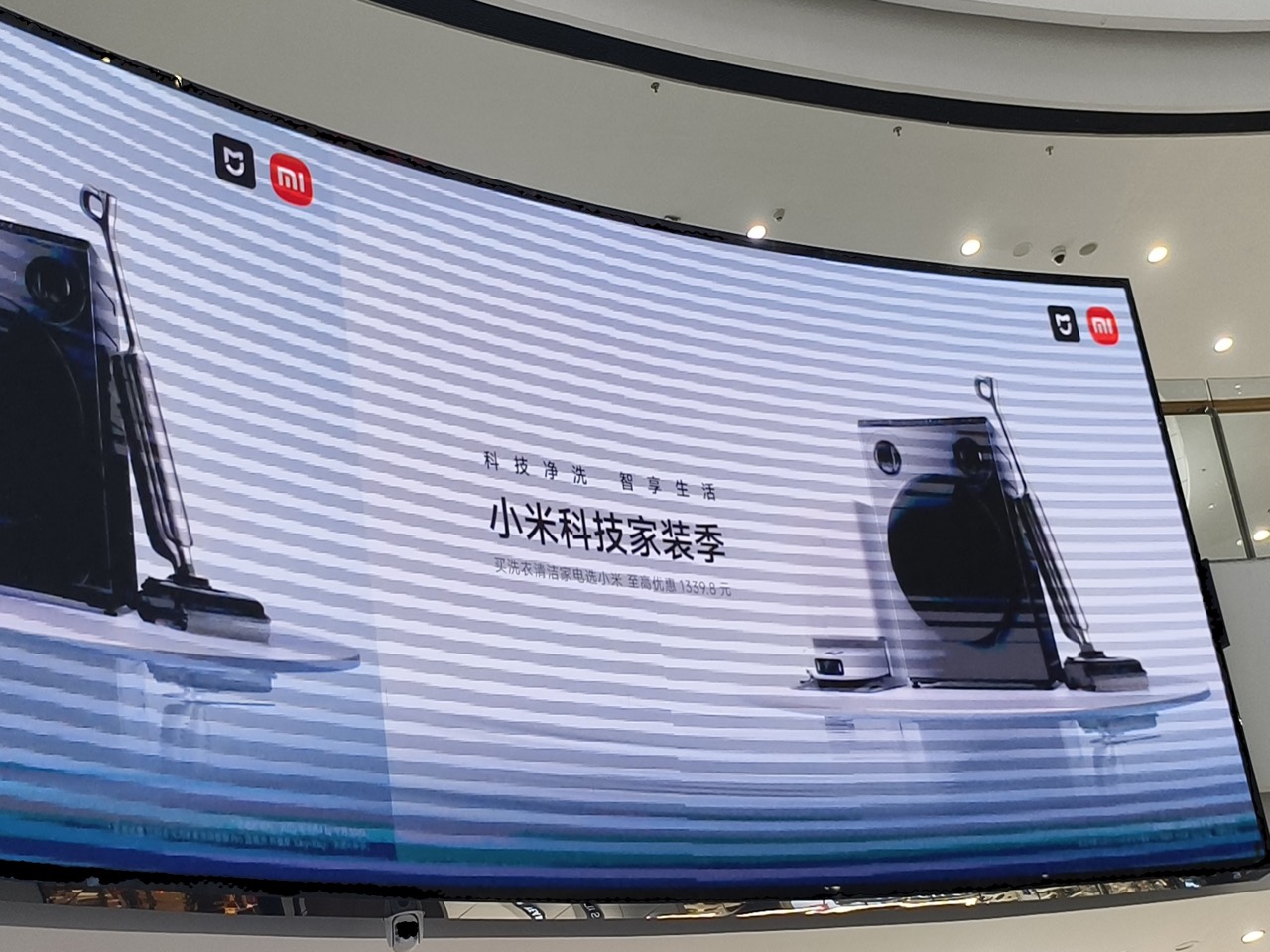} \\
    \small GT & \small LQ & \small GT & \small LQ \\
  \end{tabular}
  \caption{Our real-world dataset partial visualization (part 1).}
  \label{fig:Real-world dataset visualization1}
\end{figure*}

\begin{figure*}[t]
  \centering
  \setlength{\tabcolsep}{2pt}
  \renewcommand{\arraystretch}{1.0}
  \begin{tabular}{cccc}
    \includegraphics[width=0.23\linewidth,height=0.1725\linewidth]{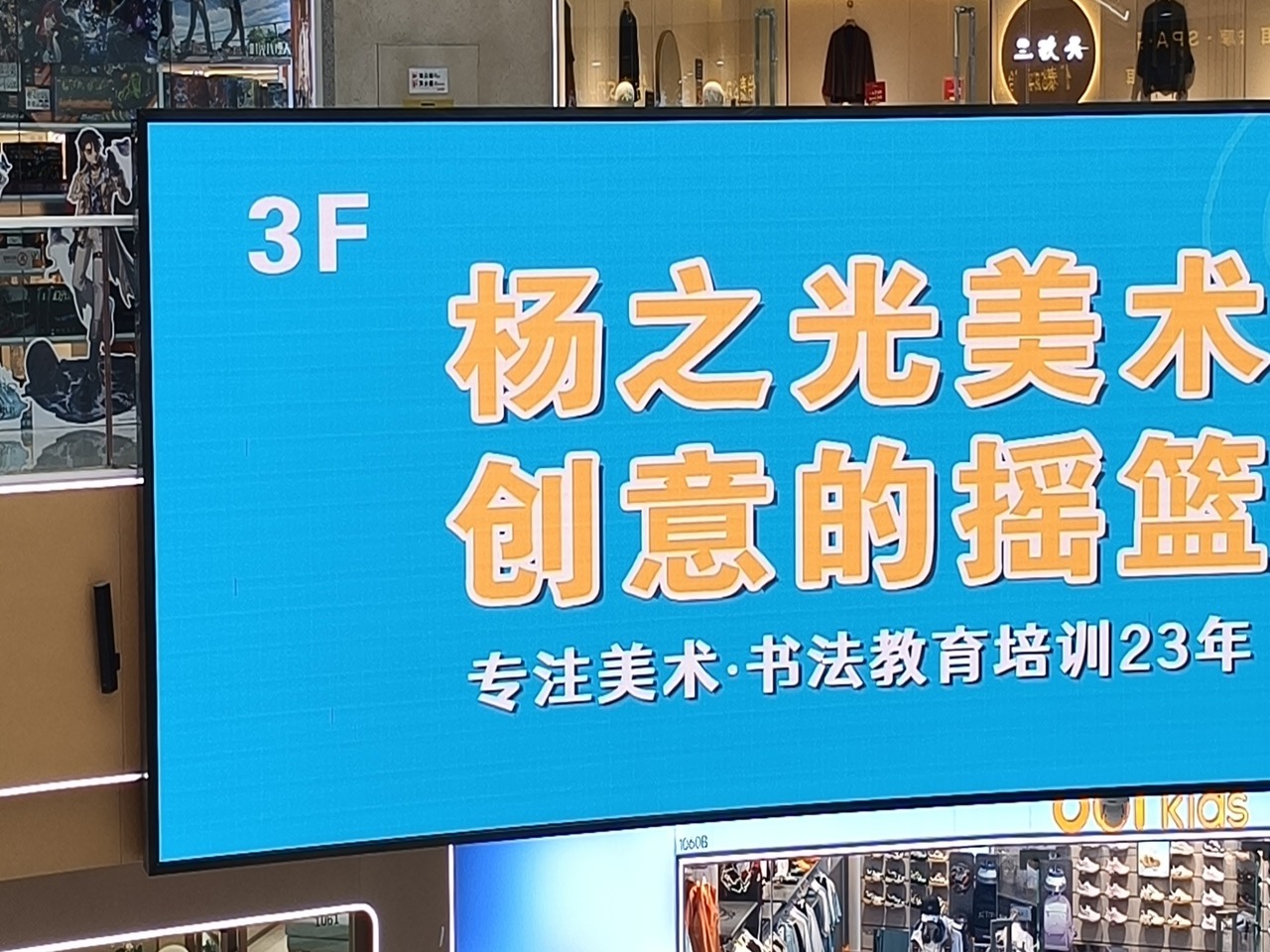} &
    \includegraphics[width=0.23\linewidth,height=0.1725\linewidth]{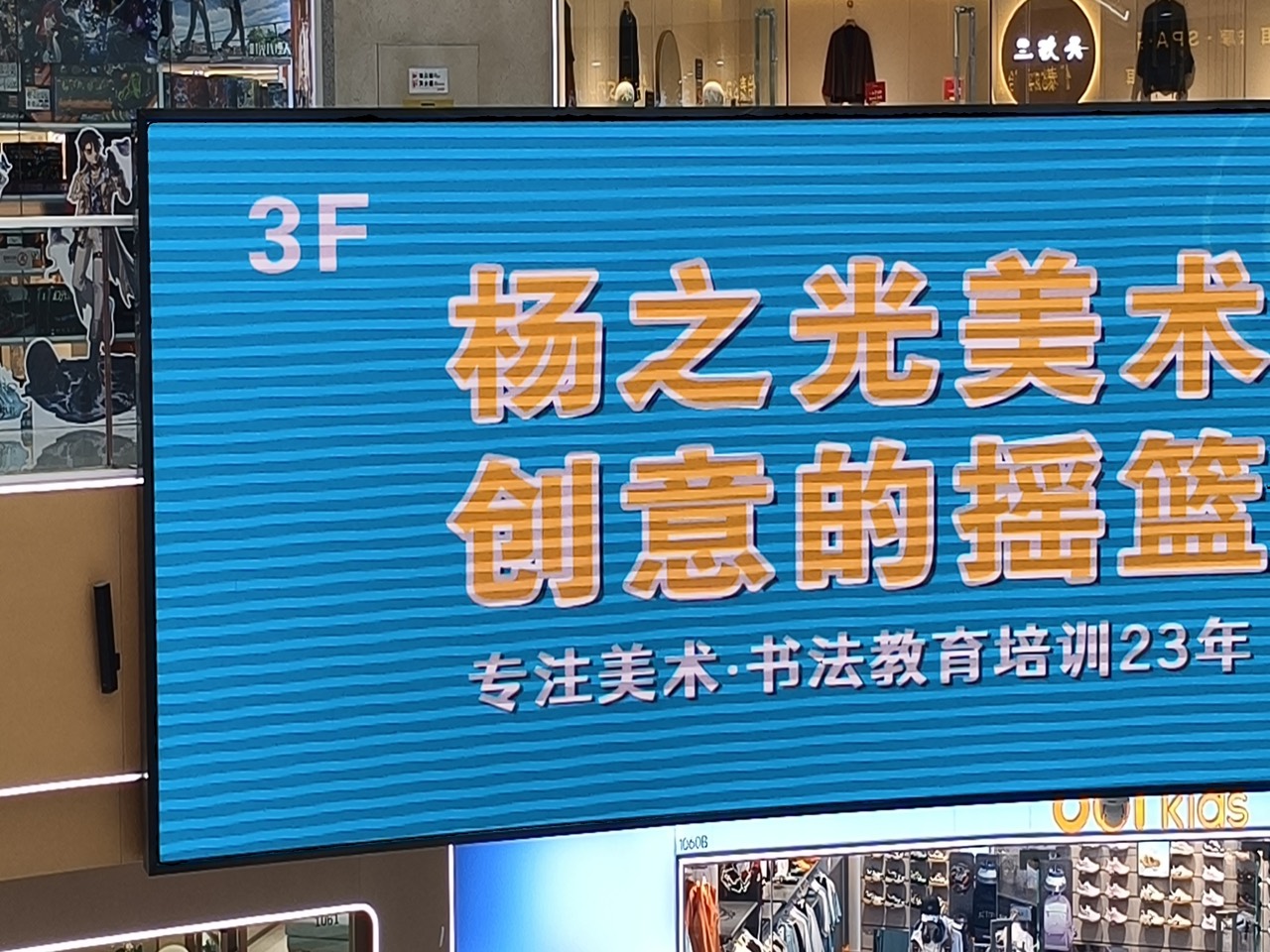} &
    \includegraphics[width=0.23\linewidth,height=0.1725\linewidth]{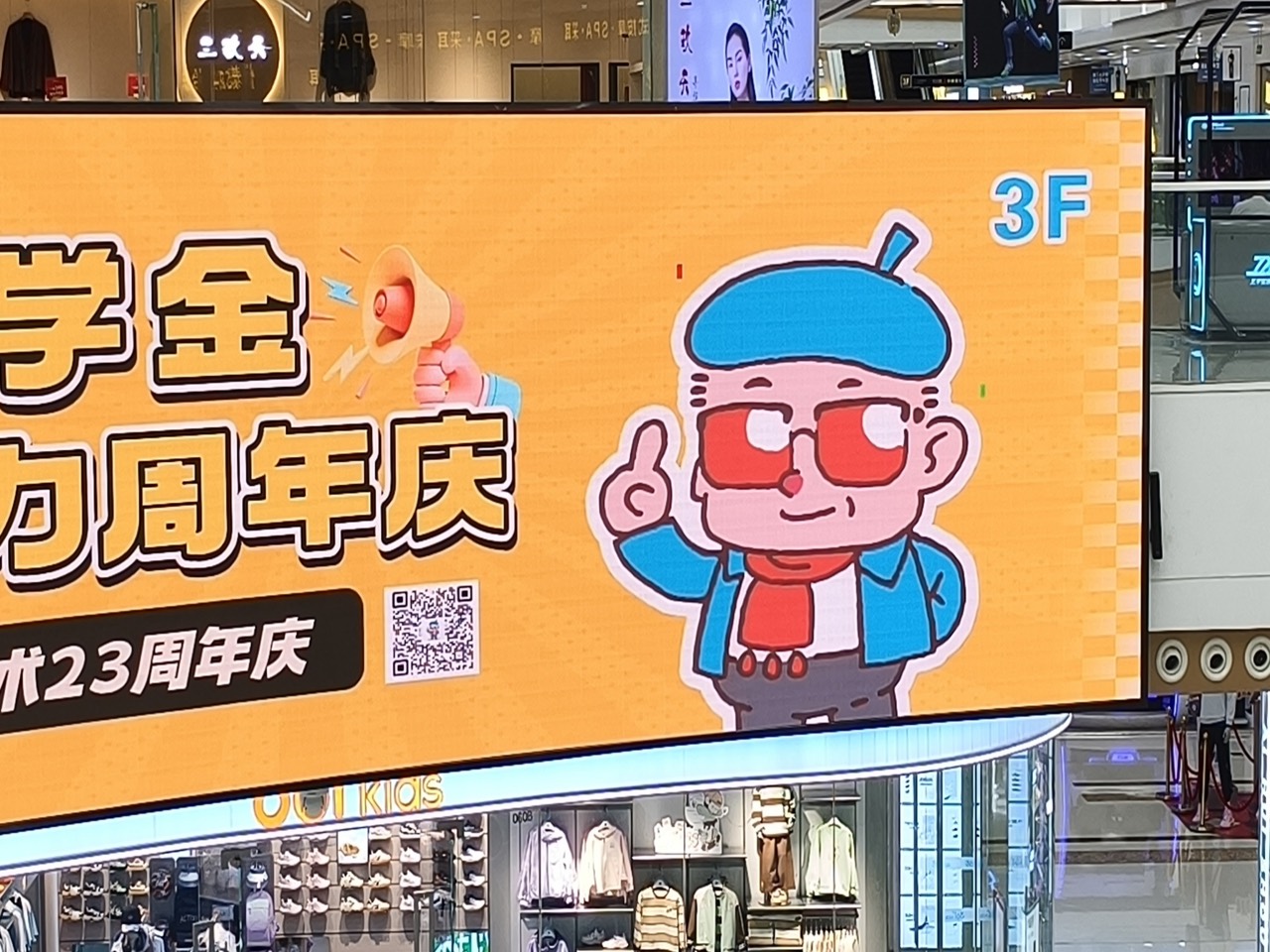} &
    \includegraphics[width=0.23\linewidth,height=0.1725\linewidth]{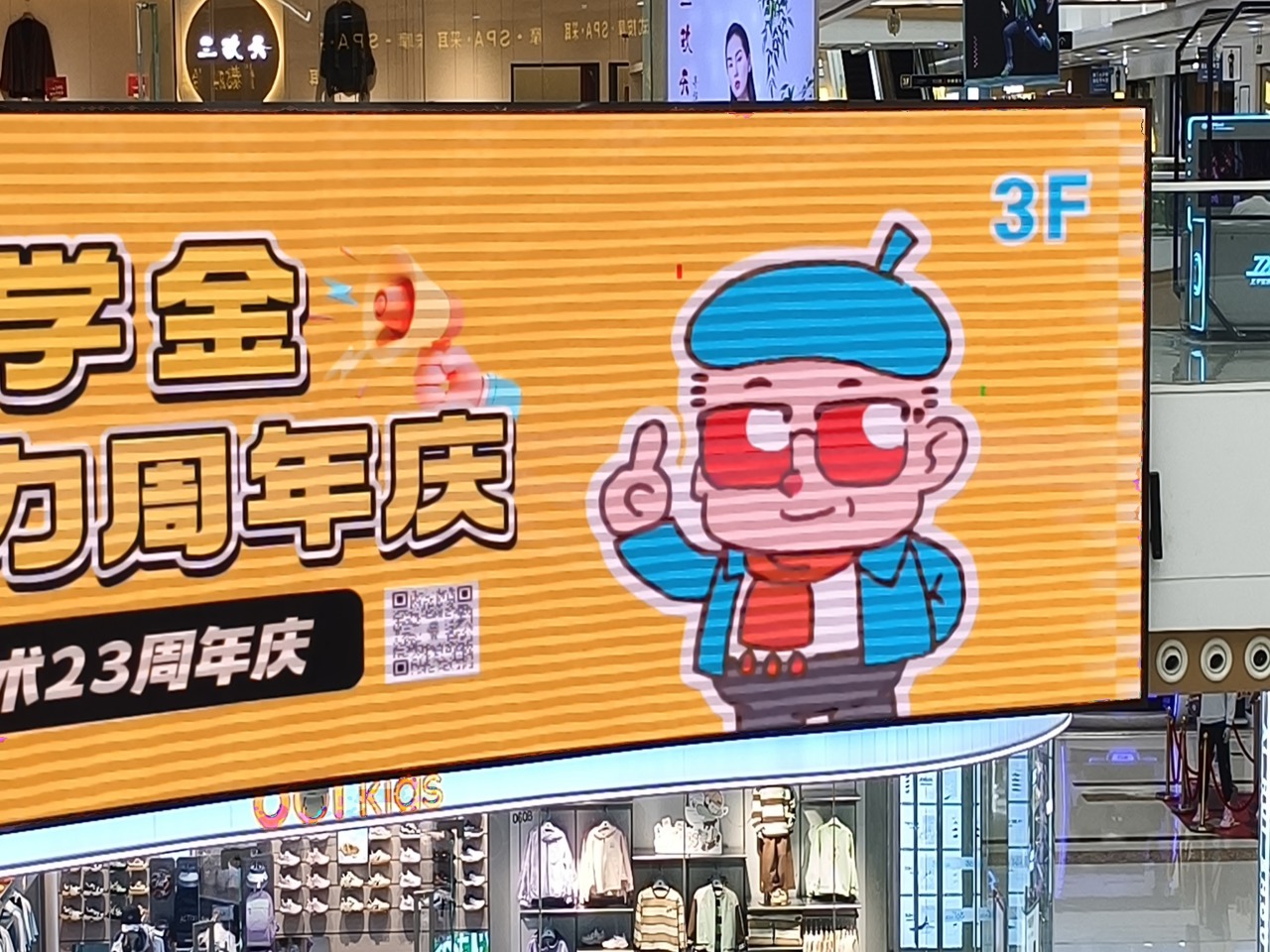} \\
    \small GT & \small LQ & \small GT & \small LQ \\
    \includegraphics[width=0.23\linewidth,height=0.1725\linewidth]{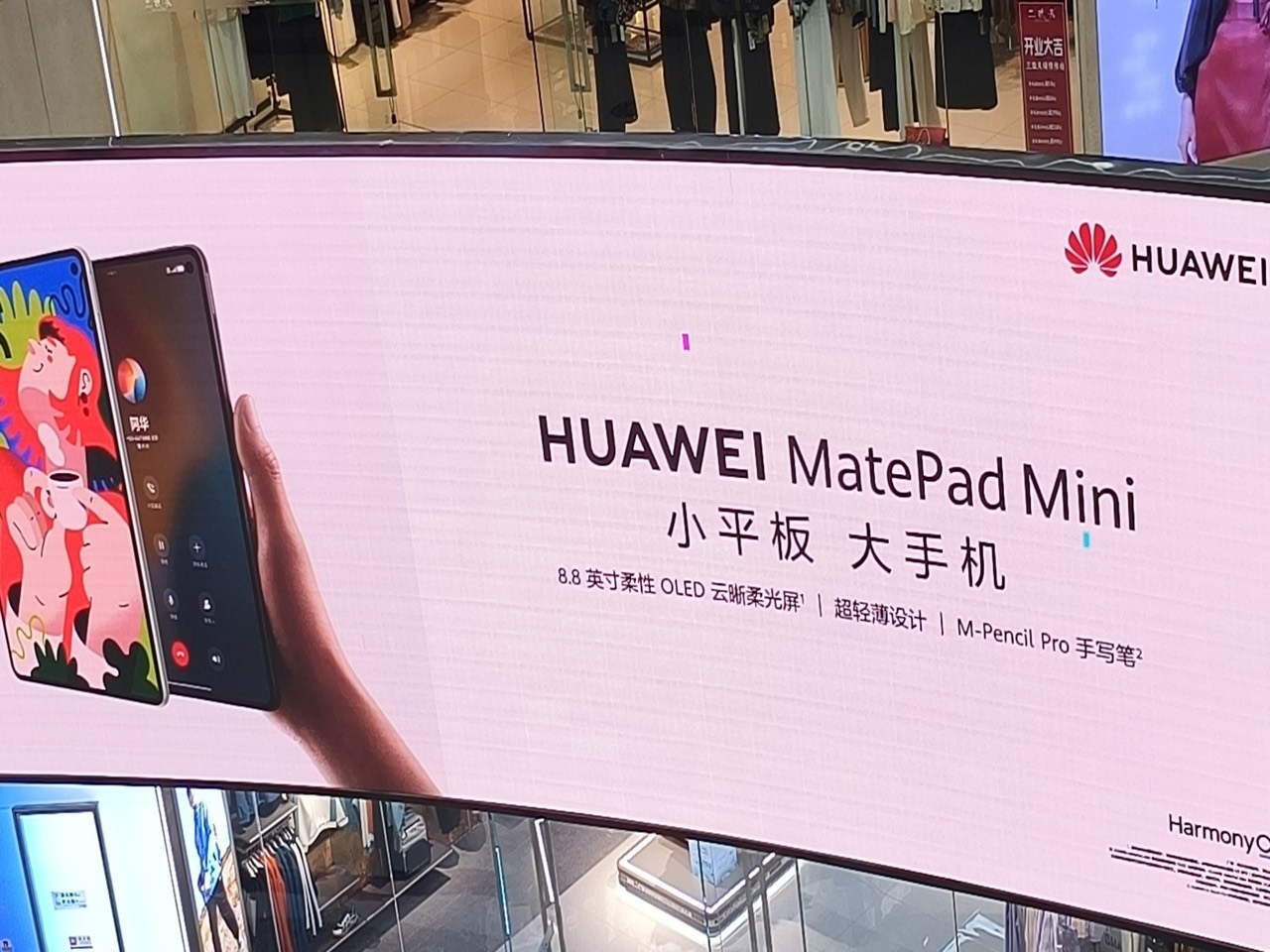} &
    \includegraphics[width=0.23\linewidth,height=0.1725\linewidth]{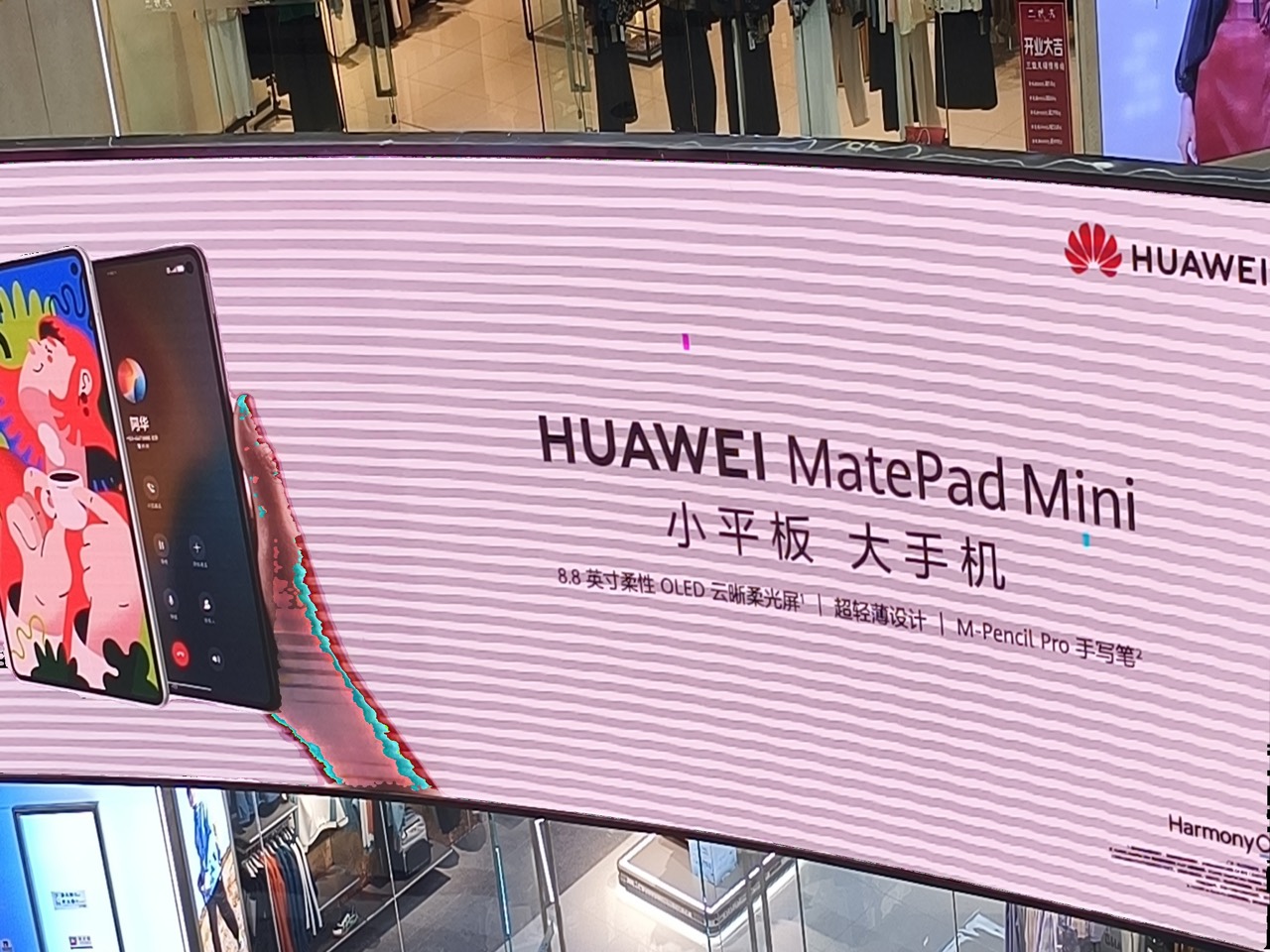} &
    \includegraphics[width=0.23\linewidth,height=0.1725\linewidth]{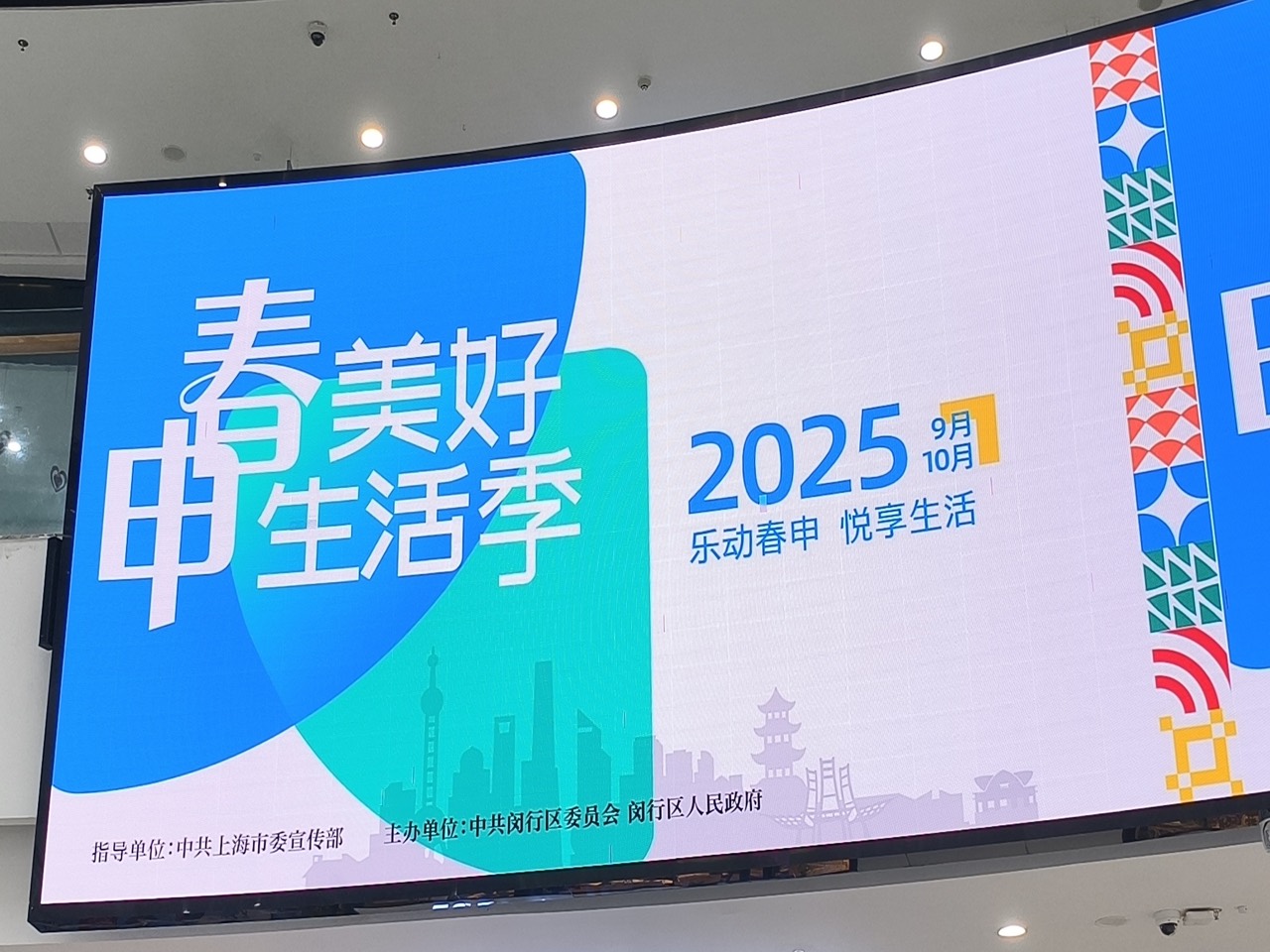} &
    \includegraphics[width=0.23\linewidth,height=0.1725\linewidth]{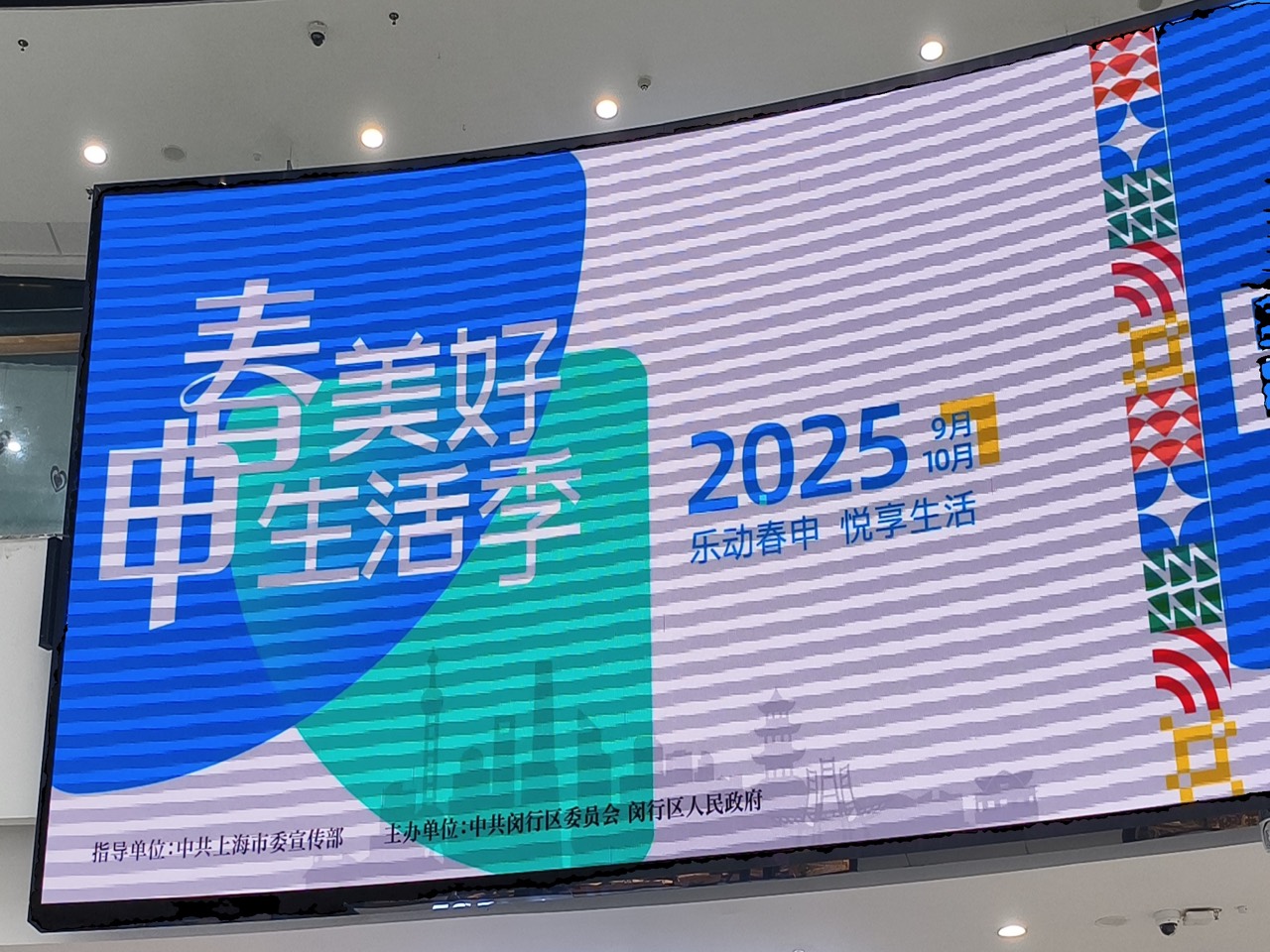} \\
    \small GT & \small LQ & \small GT & \small LQ \\
  \end{tabular}
  \caption{Our real-world dataset partial visualization (part 2).}
  \label{fig:Real-world dataset visualization2}
\end{figure*}

\begin{figure*}[t]
  \centering
  \setlength{\tabcolsep}{2pt}
  \renewcommand{\arraystretch}{1.0}
  \begin{tabular}{cccc}
    \includegraphics[width=0.23\linewidth,height=0.1725\linewidth,trim=60 0 60 0,clip]{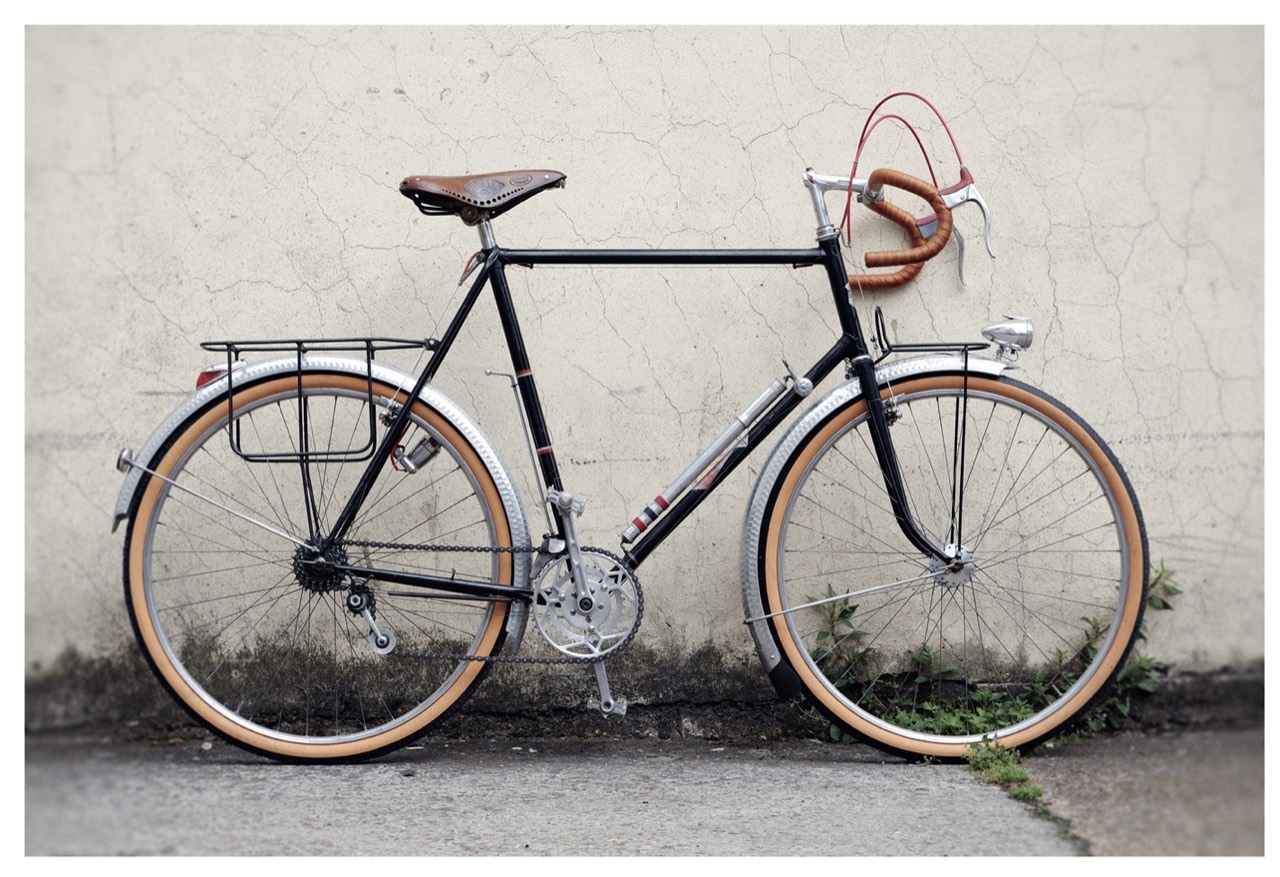} &
    \includegraphics[width=0.23\linewidth,height=0.1725\linewidth,trim=60 0 60 0,clip]{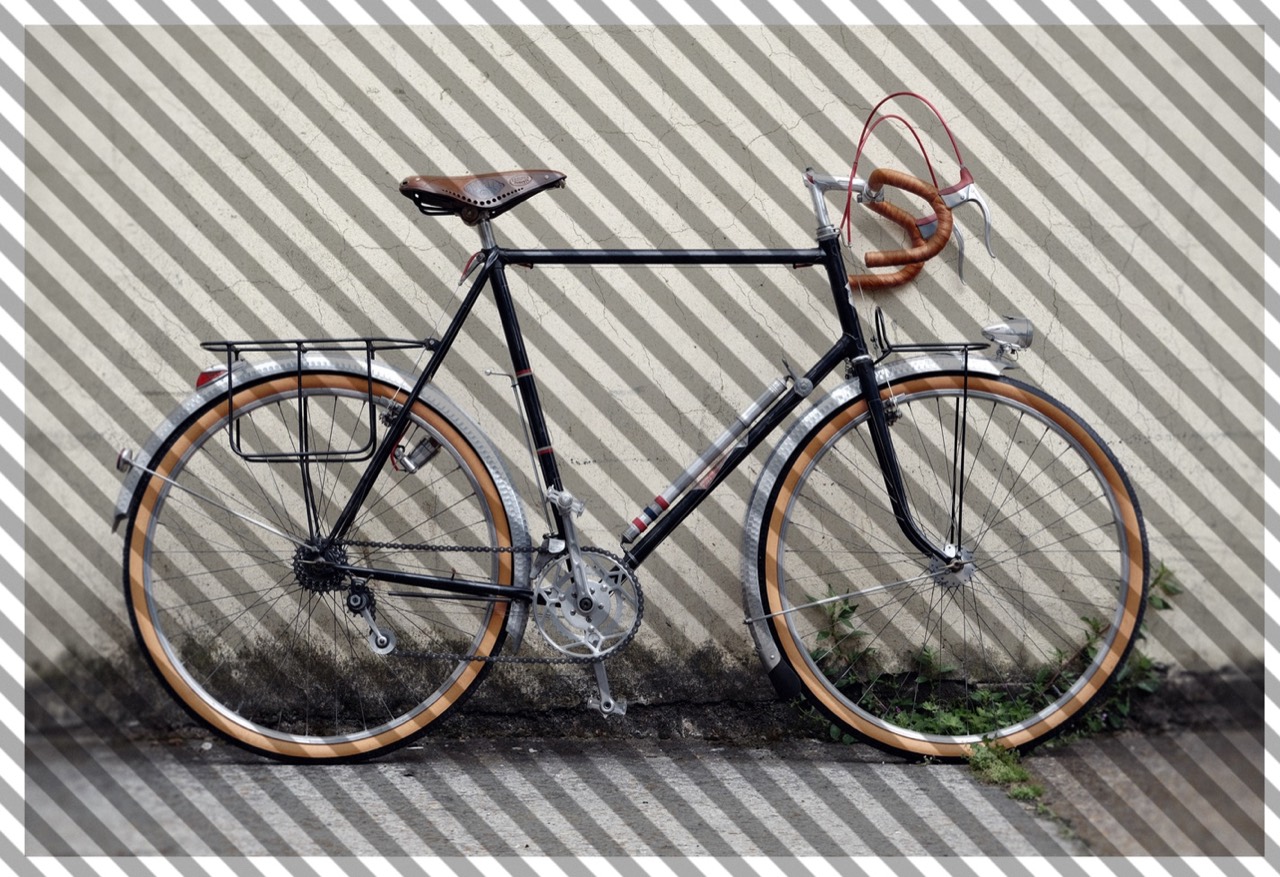} &
    \includegraphics[width=0.23\linewidth,height=0.1725\linewidth,trim=60 0 60 0,clip]{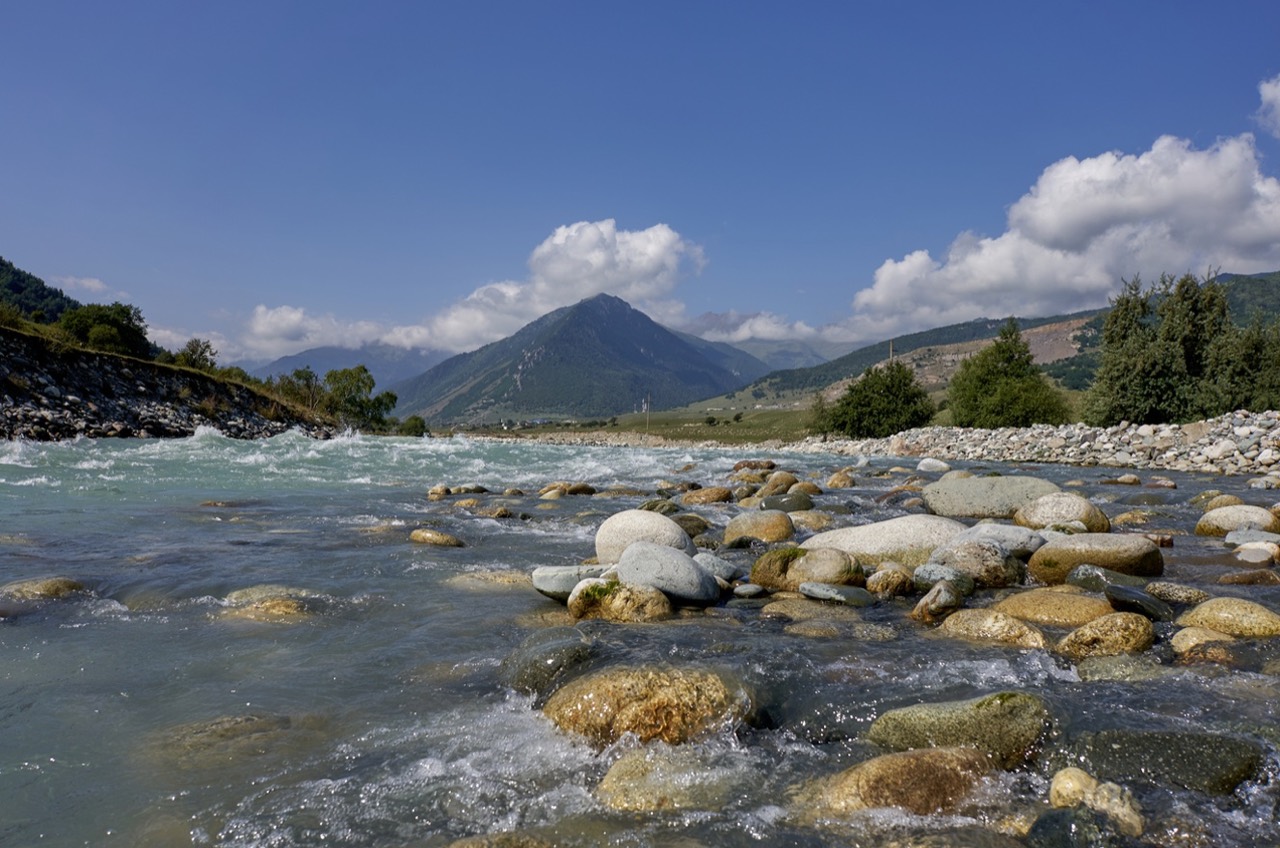} &
    \includegraphics[width=0.23\linewidth,height=0.1725\linewidth,trim=60 0 60 0,clip]{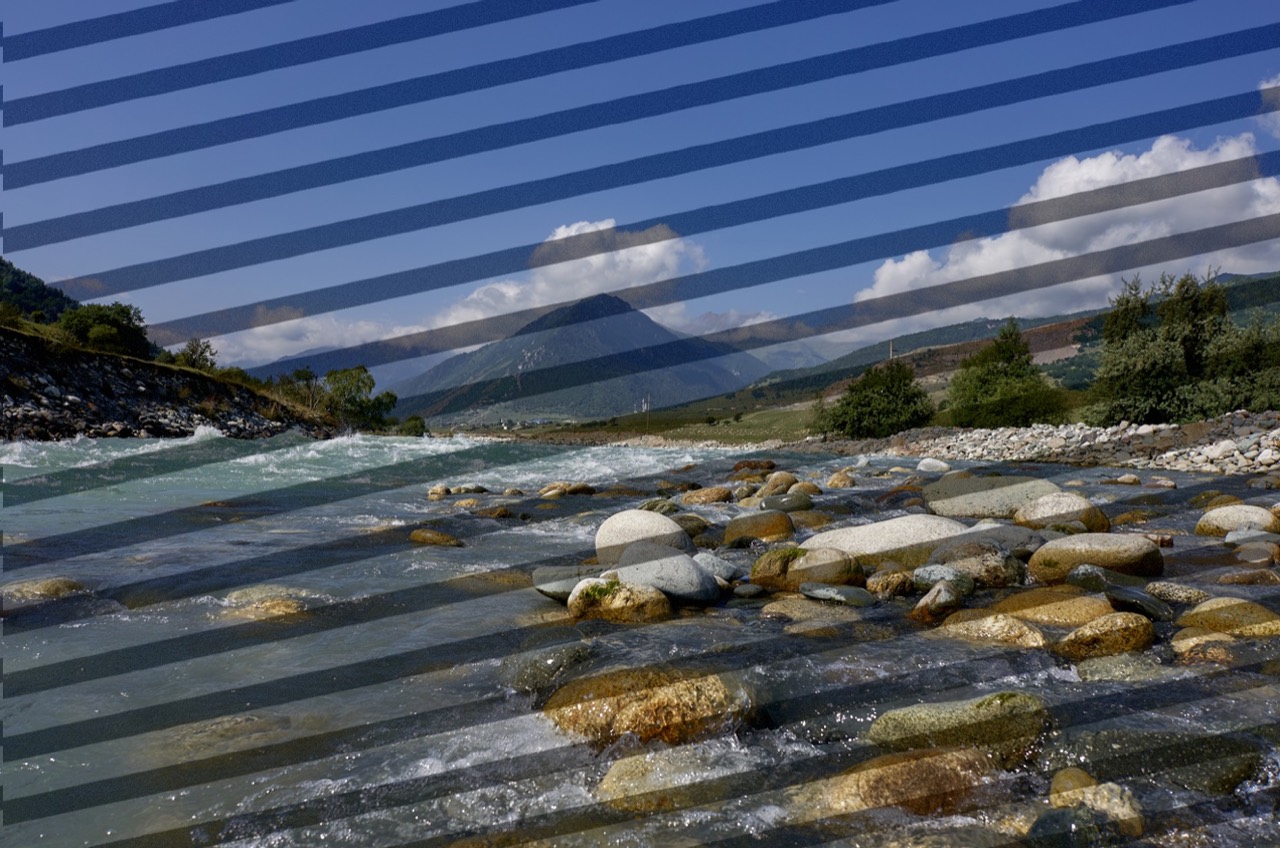} \\
    \small GT & \small LQ & \small GT & \small LQ \\
    \includegraphics[width=0.23\linewidth,height=0.1725\linewidth,trim=60 0 60 0,clip]{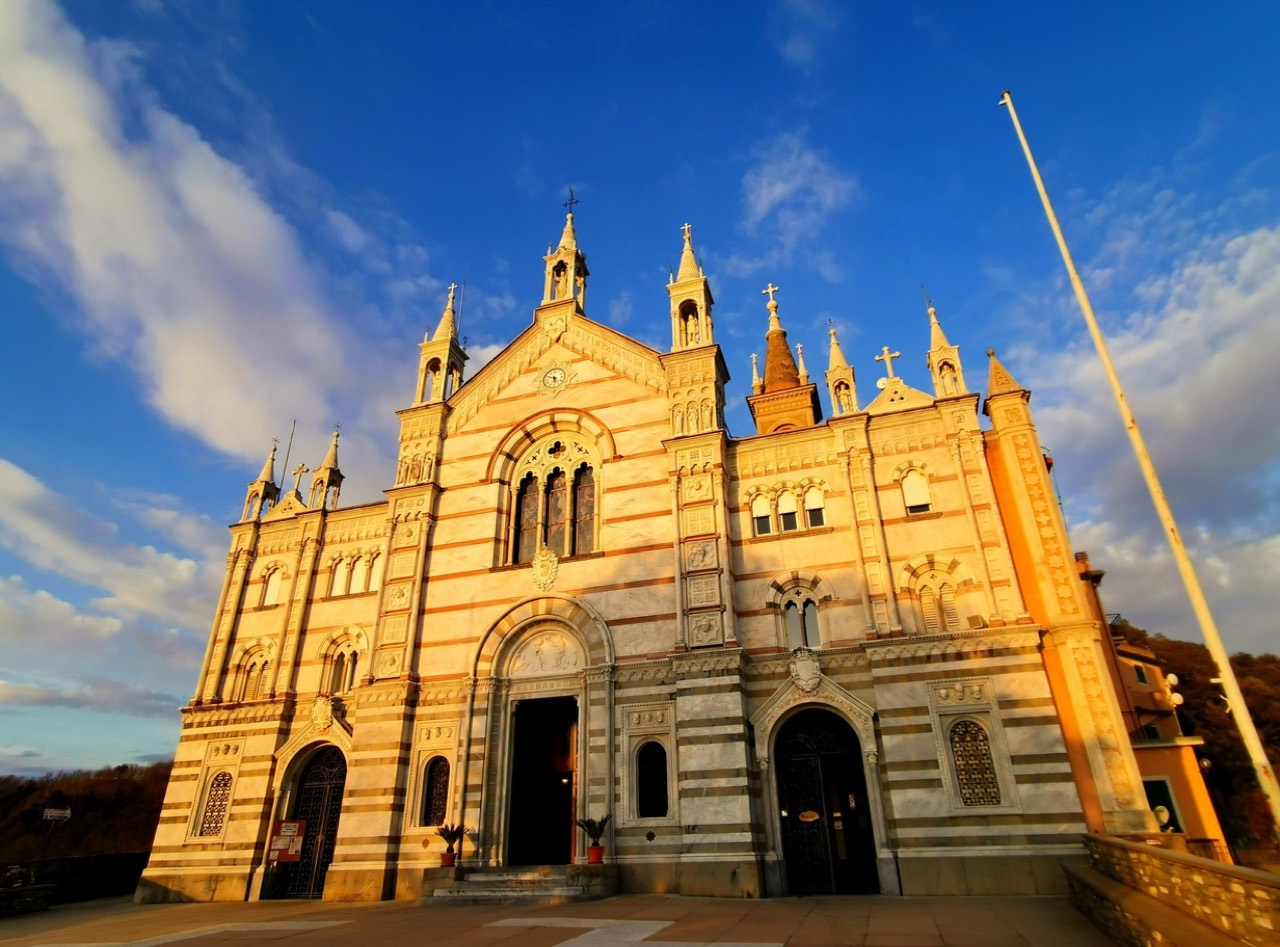} &
    \includegraphics[width=0.23\linewidth,height=0.1725\linewidth,trim=60 0 60 0,clip]{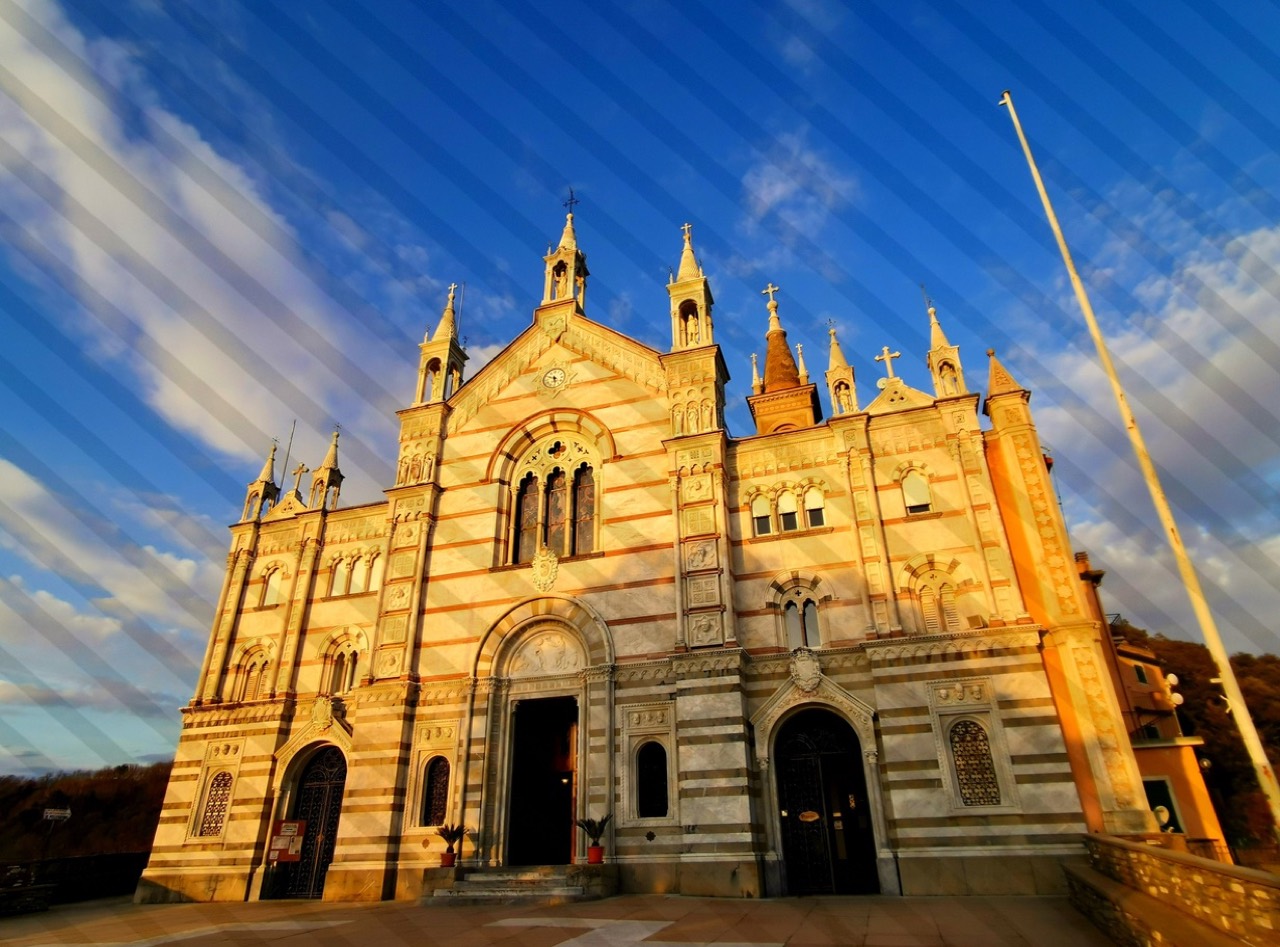} &
    \includegraphics[width=0.23\linewidth,height=0.1725\linewidth,trim=60 0 60 0,clip]{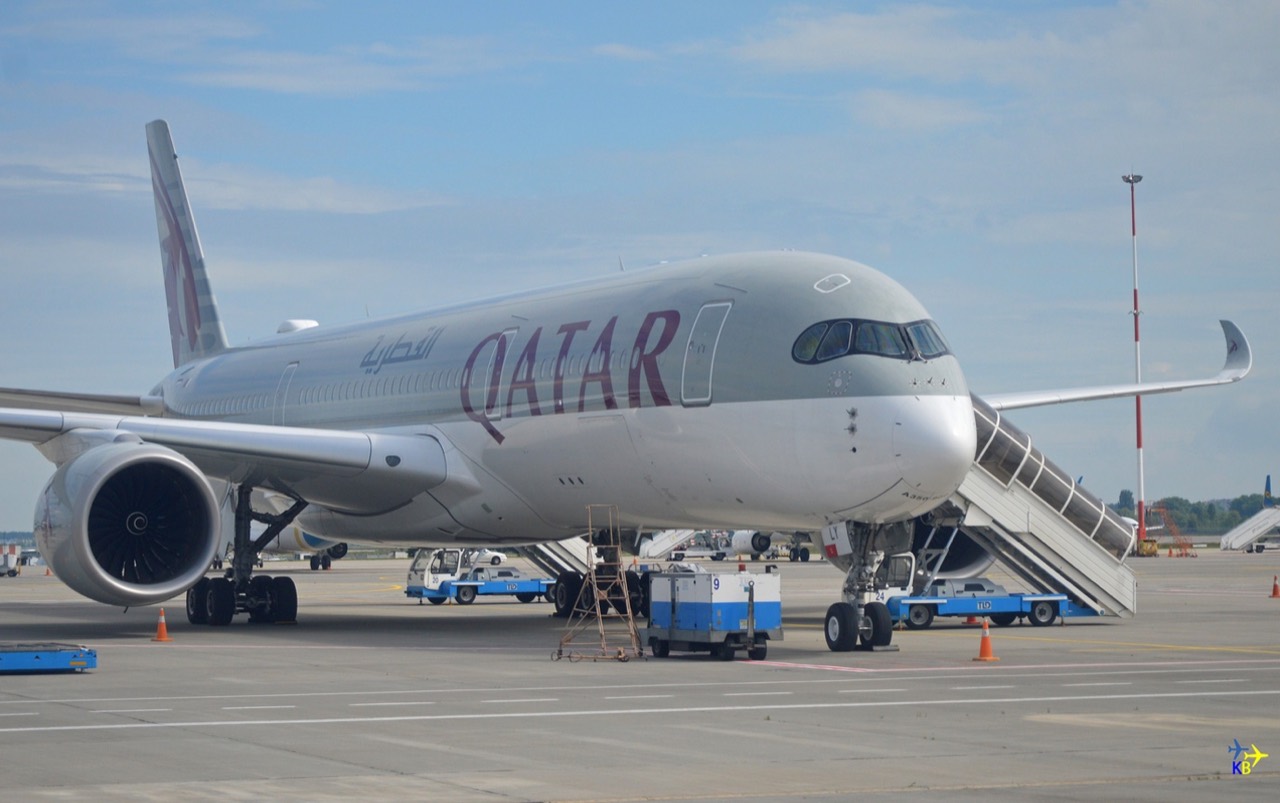} &
    \includegraphics[width=0.23\linewidth,height=0.1725\linewidth,trim=60 0 60 0,clip]{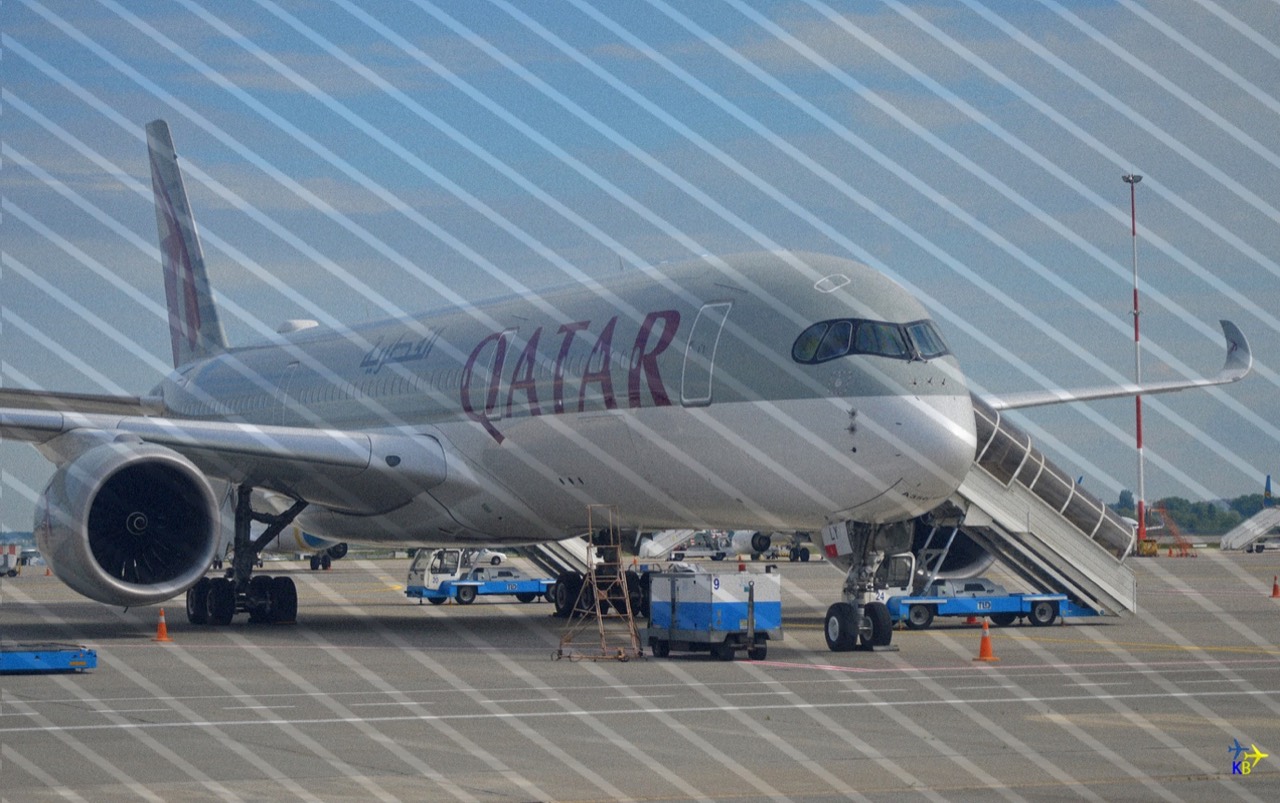} \\
    \small GT & \small LQ & \small GT & \small LQ \\
    \includegraphics[width=0.23\linewidth,height=0.1725\linewidth,trim=66 0 66 0,clip]{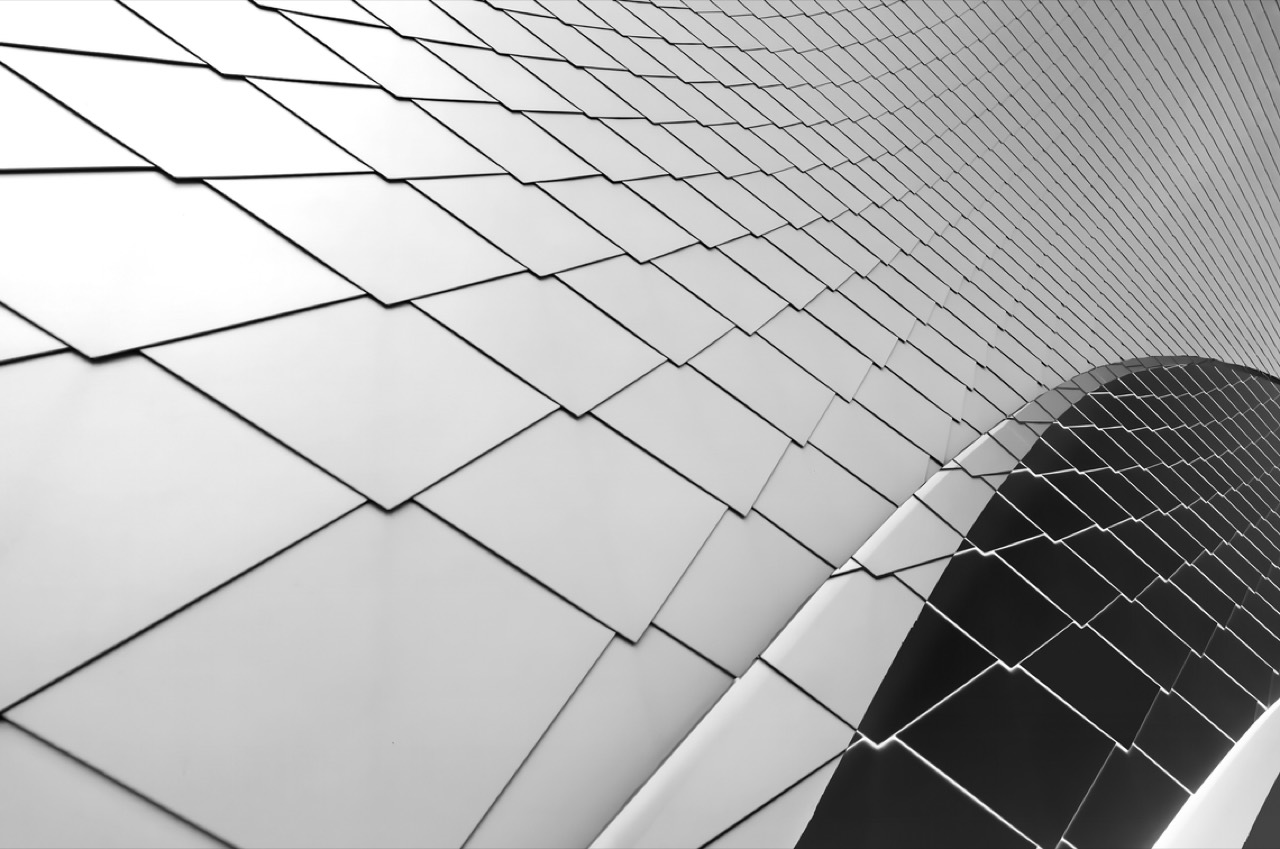} &
    \includegraphics[width=0.23\linewidth,height=0.1725\linewidth,trim=66 0 66 0,clip]{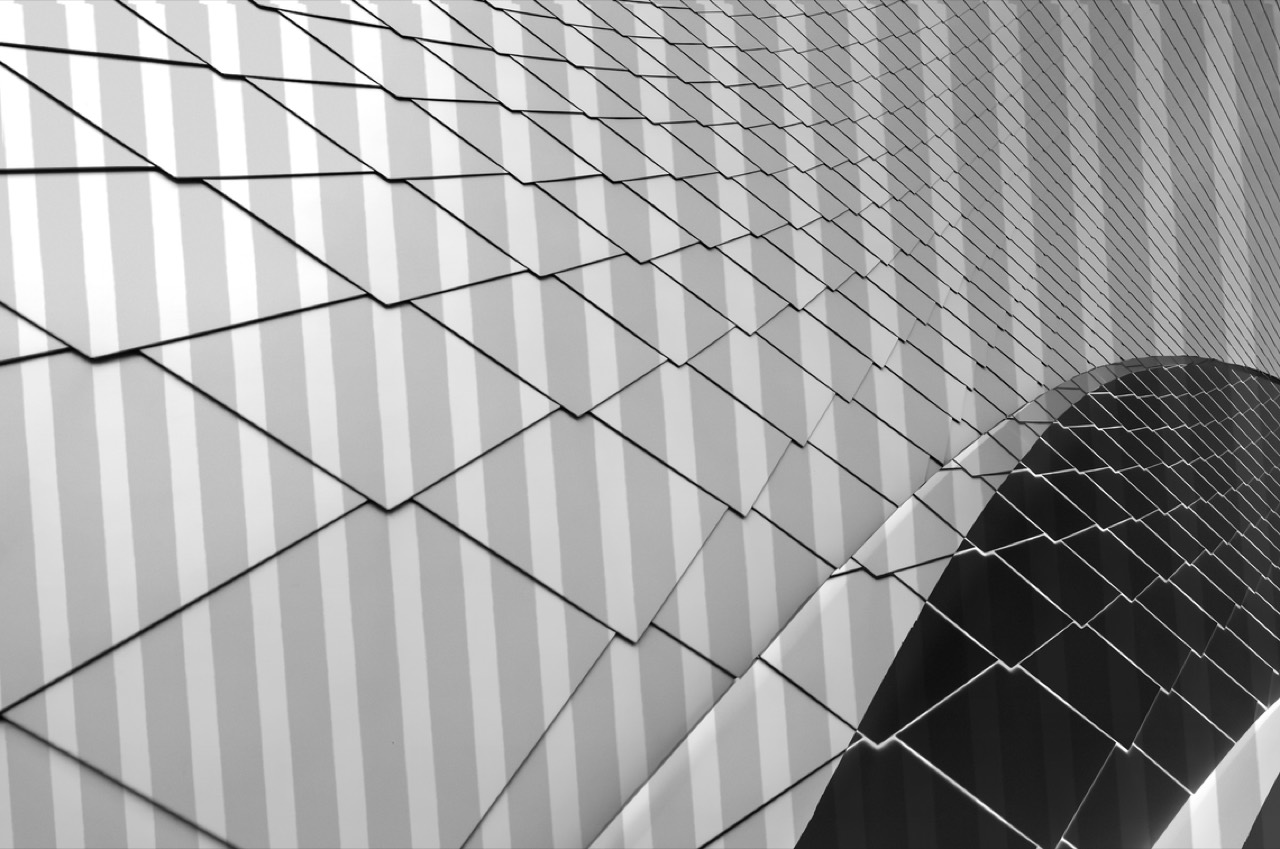} &
    \includegraphics[width=0.23\linewidth,height=0.1725\linewidth]{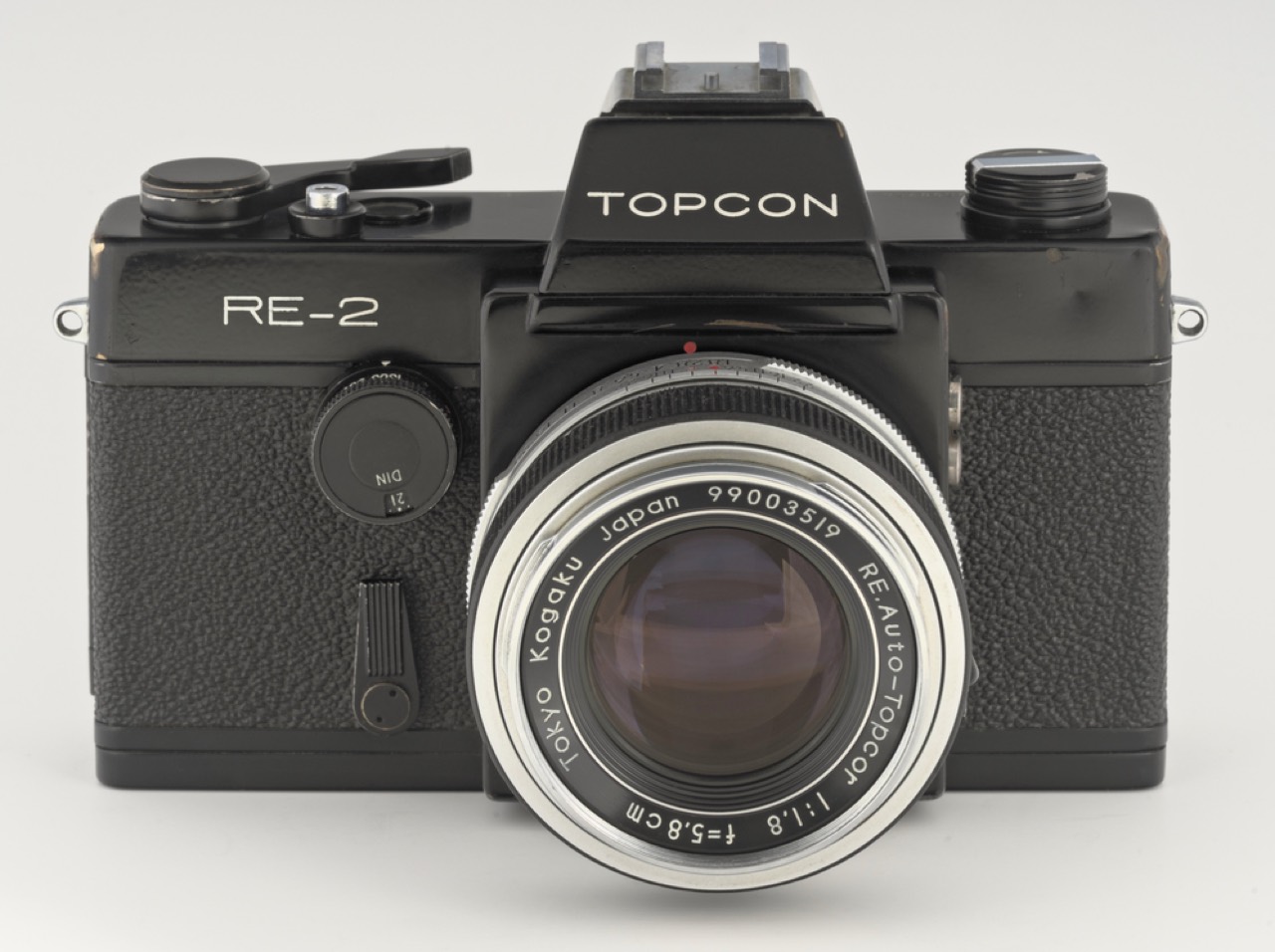} &
    \includegraphics[width=0.23\linewidth,height=0.1725\linewidth]{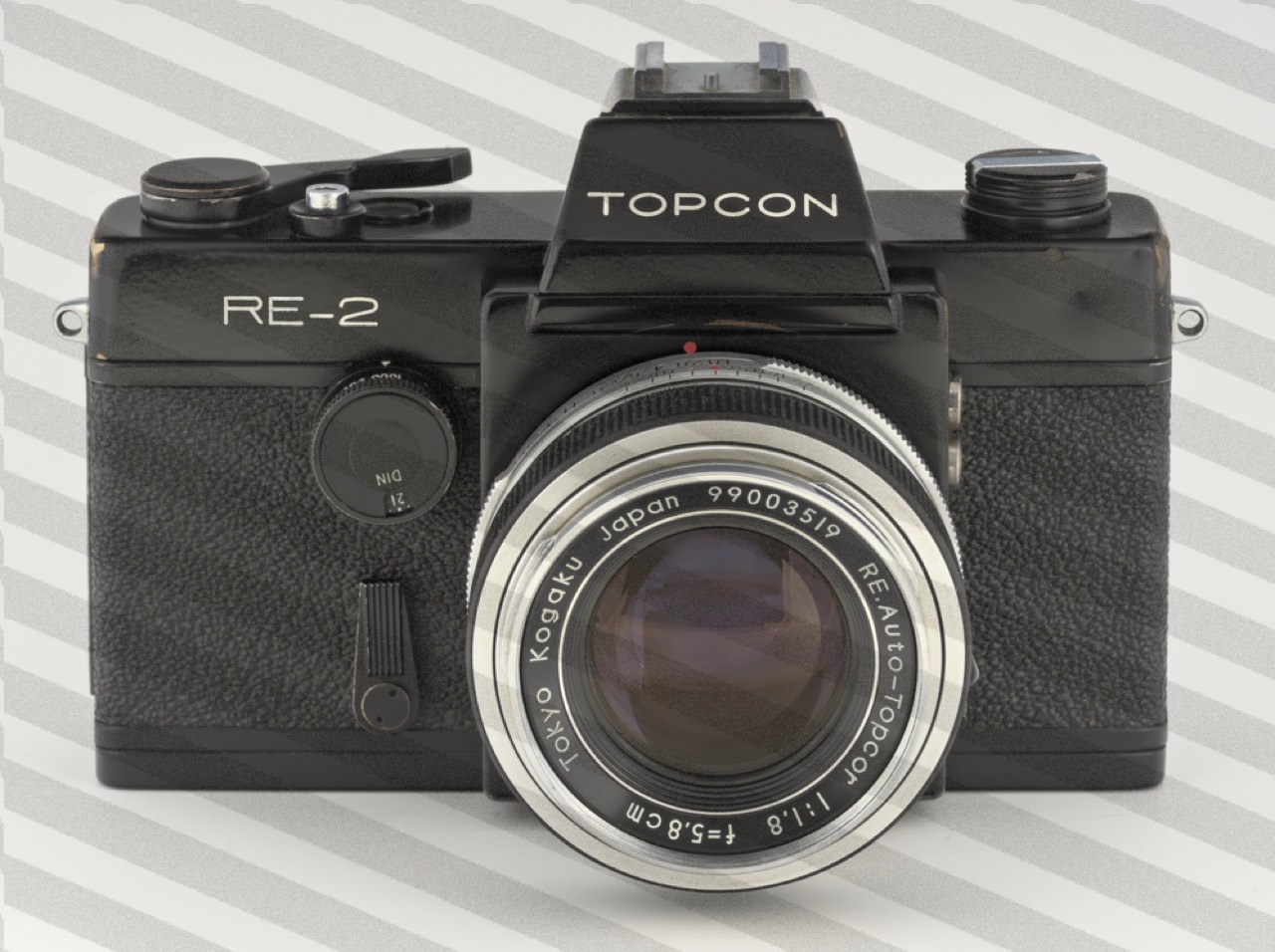} \\
    \small GT & \small LQ & \small GT & \small LQ \\
  \end{tabular}
  \caption{Our simulated dataset based on LSDIR~\citep{li2023lsdir} partial visualization (part 1).}
  \label{fig:simulated dataset visualization1}
\end{figure*}

\begin{figure*}[t]
  \centering
  \setlength{\tabcolsep}{2pt}
  \renewcommand{\arraystretch}{1.0}
  \begin{tabular}{cccc}
    \includegraphics[width=0.23\linewidth,height=0.1725\linewidth]{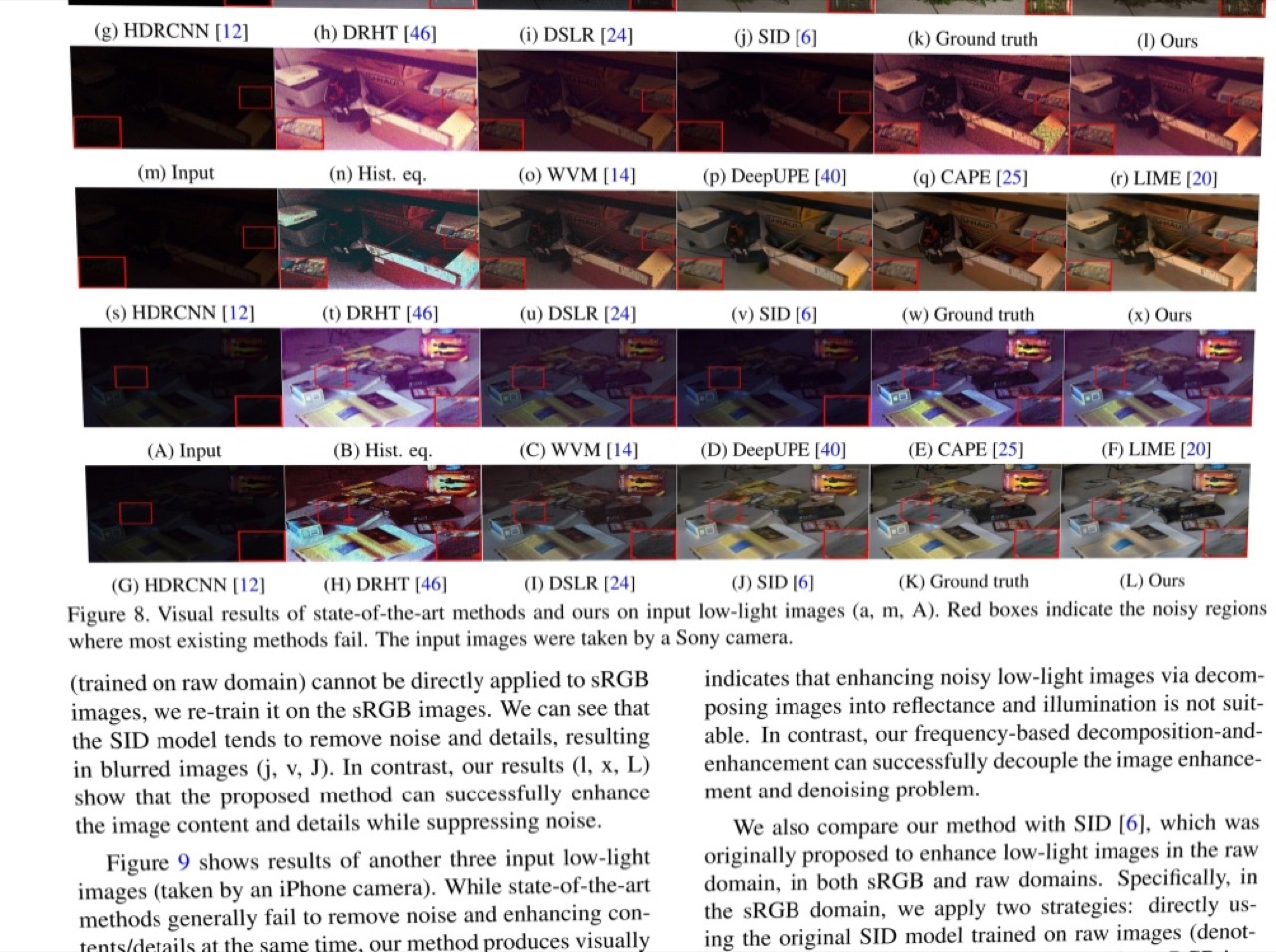} &
    \includegraphics[width=0.23\linewidth,height=0.1725\linewidth]{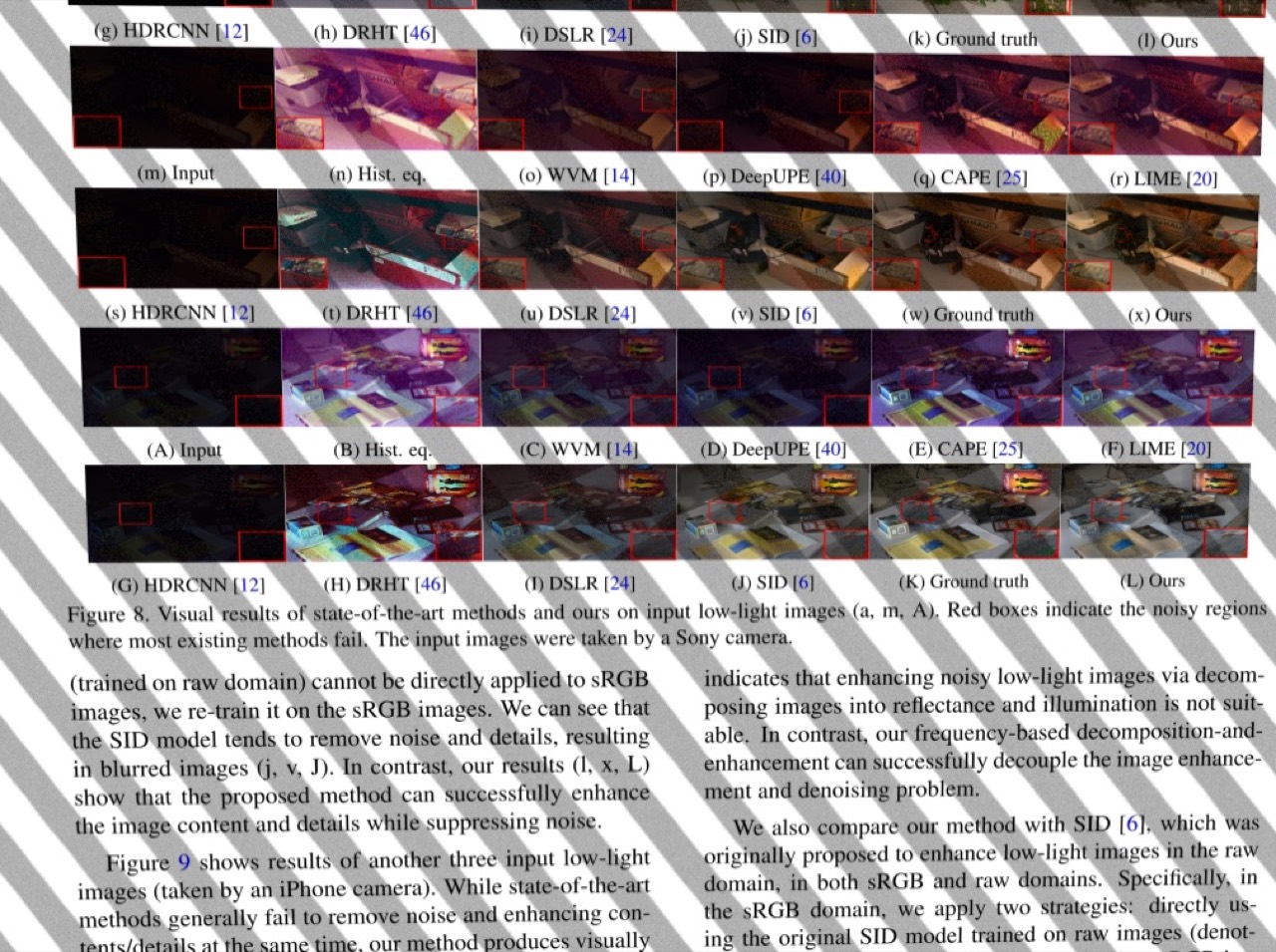} &
    \includegraphics[width=0.23\linewidth,height=0.1725\linewidth]{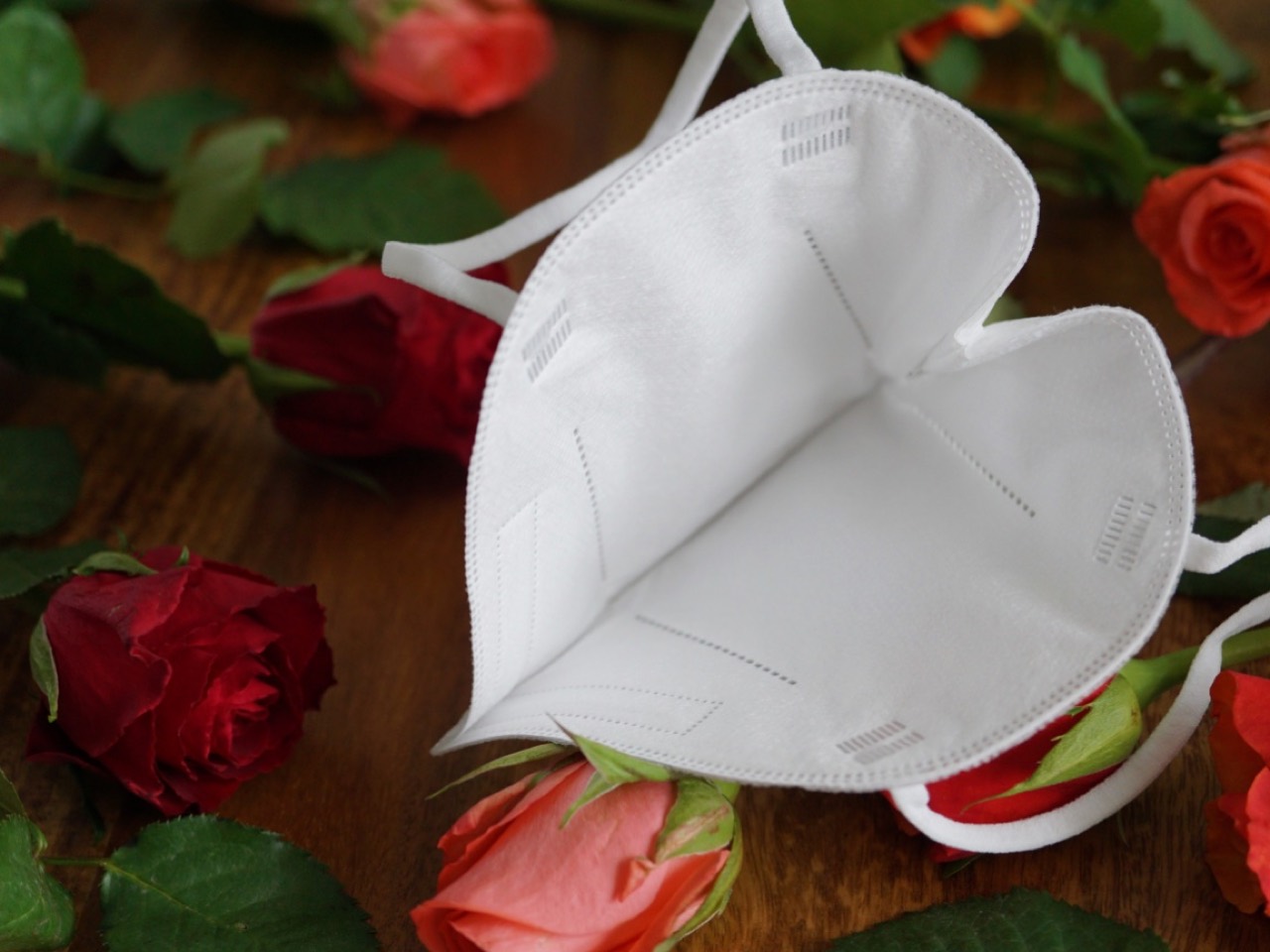} &
    \includegraphics[width=0.23\linewidth,height=0.1725\linewidth]{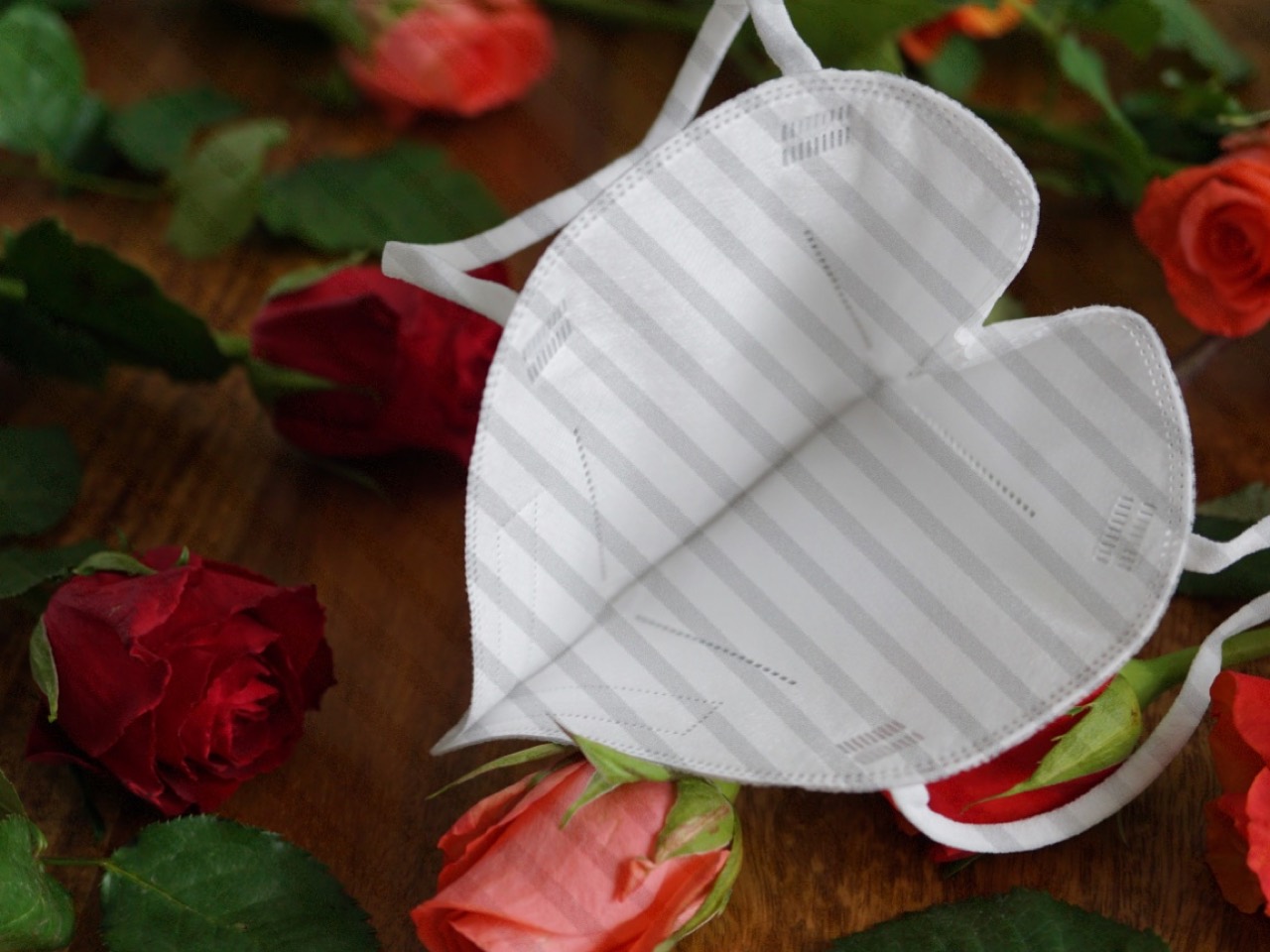} \\
    \small GT & \small LQ & \small GT & \small LQ \\
    \includegraphics[width=0.23\linewidth,height=0.1725\linewidth]{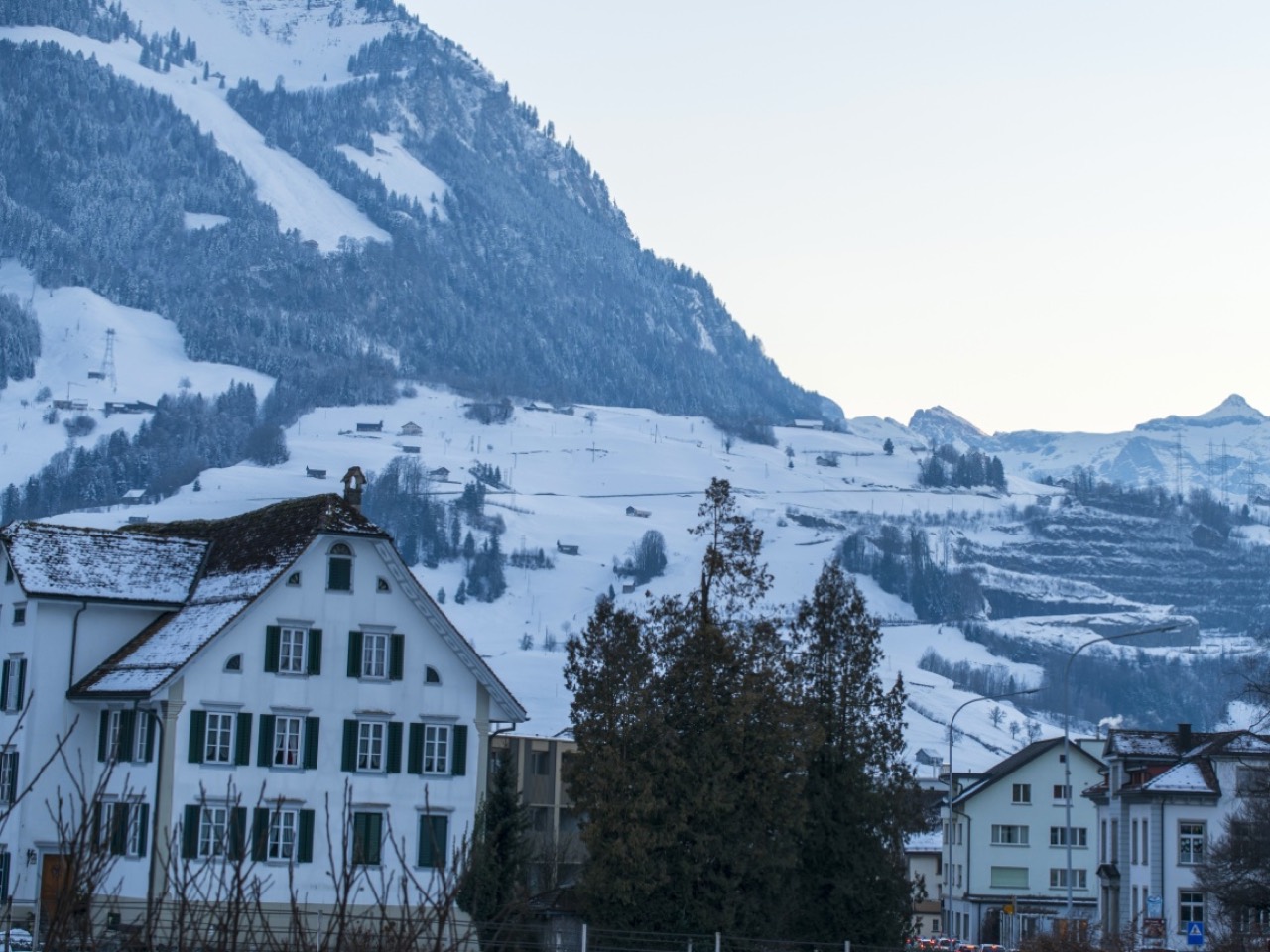} &
    \includegraphics[width=0.23\linewidth,height=0.1725\linewidth]{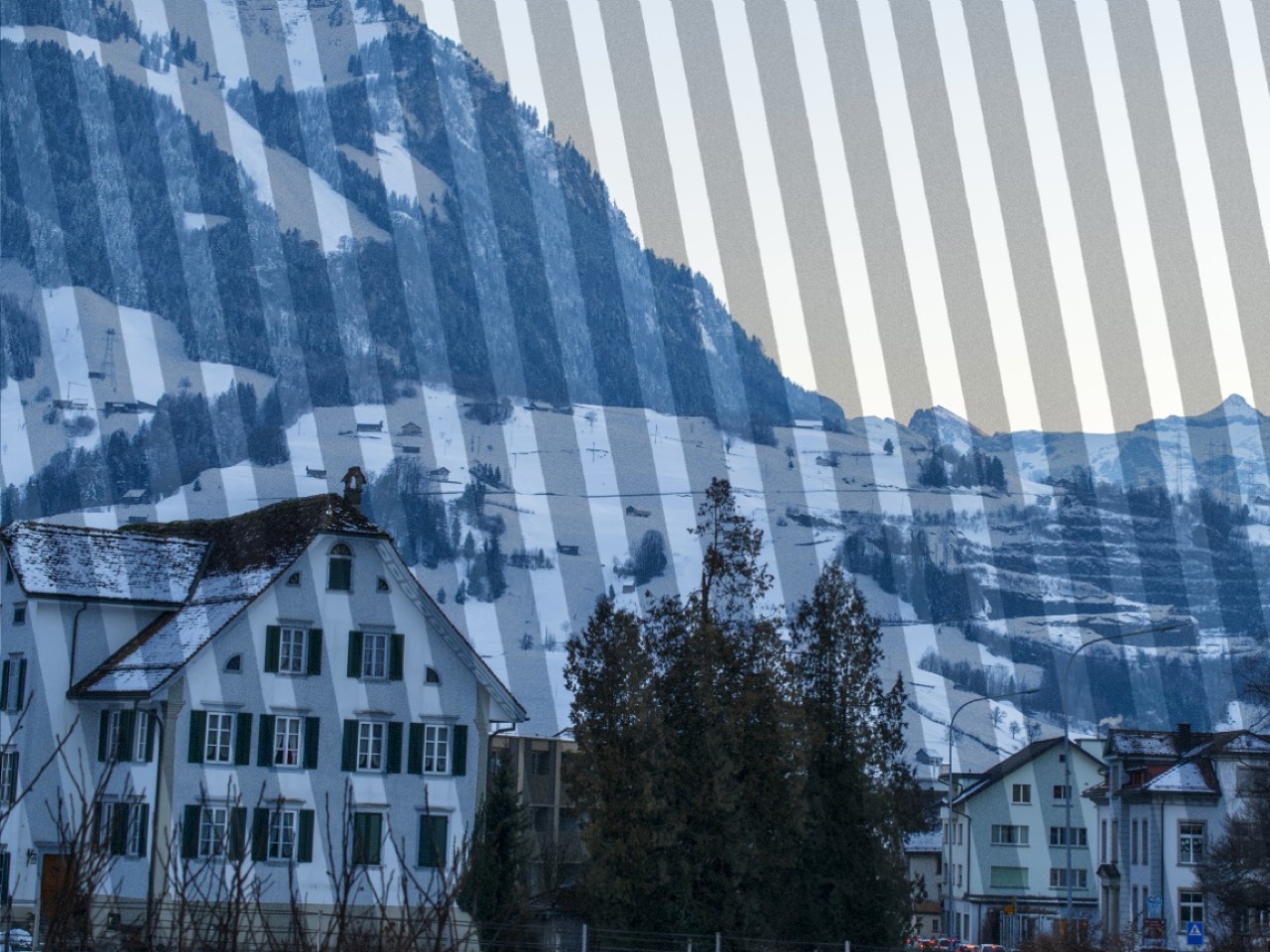} &
    \includegraphics[width=0.23\linewidth,height=0.1725\linewidth]{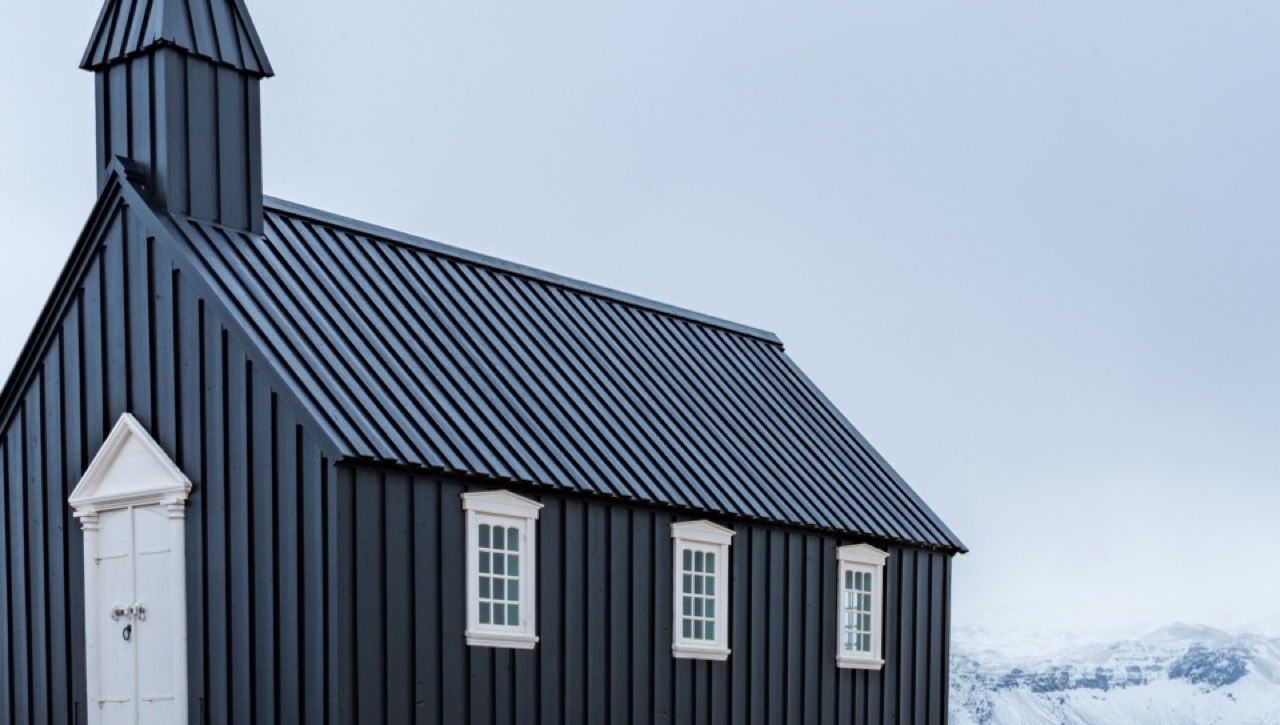} &
    \includegraphics[width=0.23\linewidth,height=0.1725\linewidth]{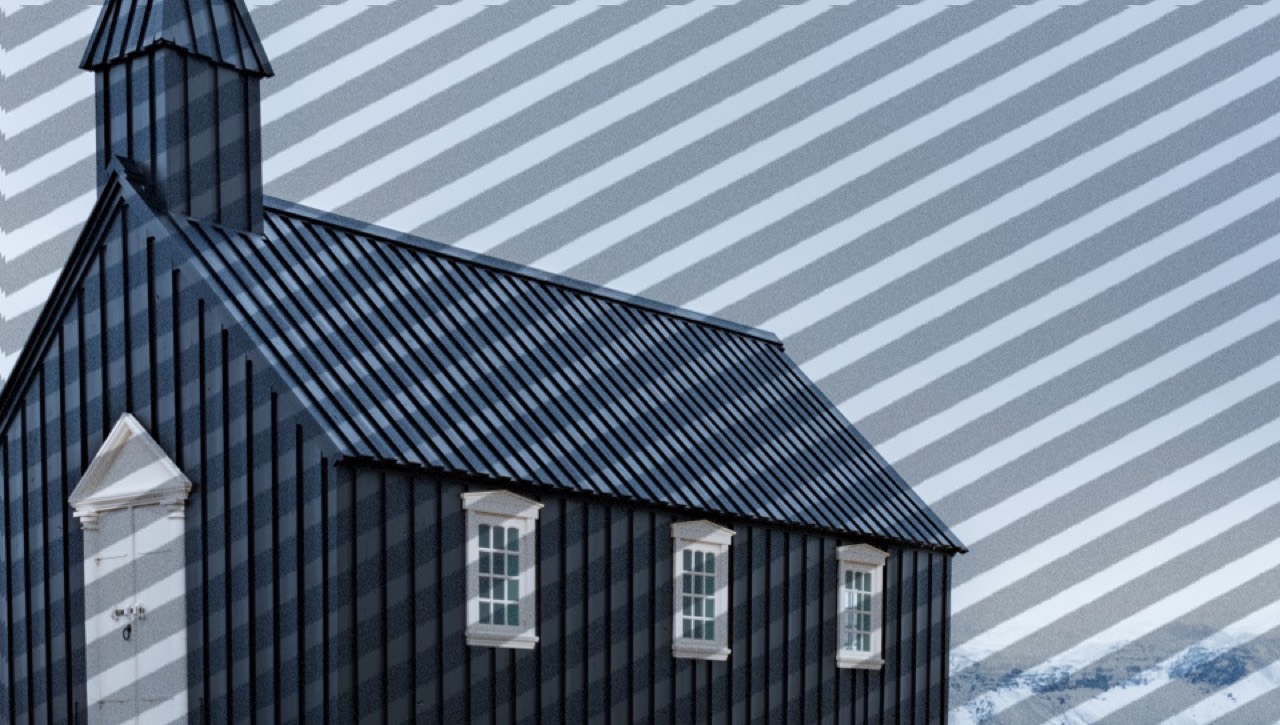} \\
    \small GT & \small LQ & \small GT & \small LQ \\
    \includegraphics[width=0.23\linewidth,height=0.1725\linewidth]{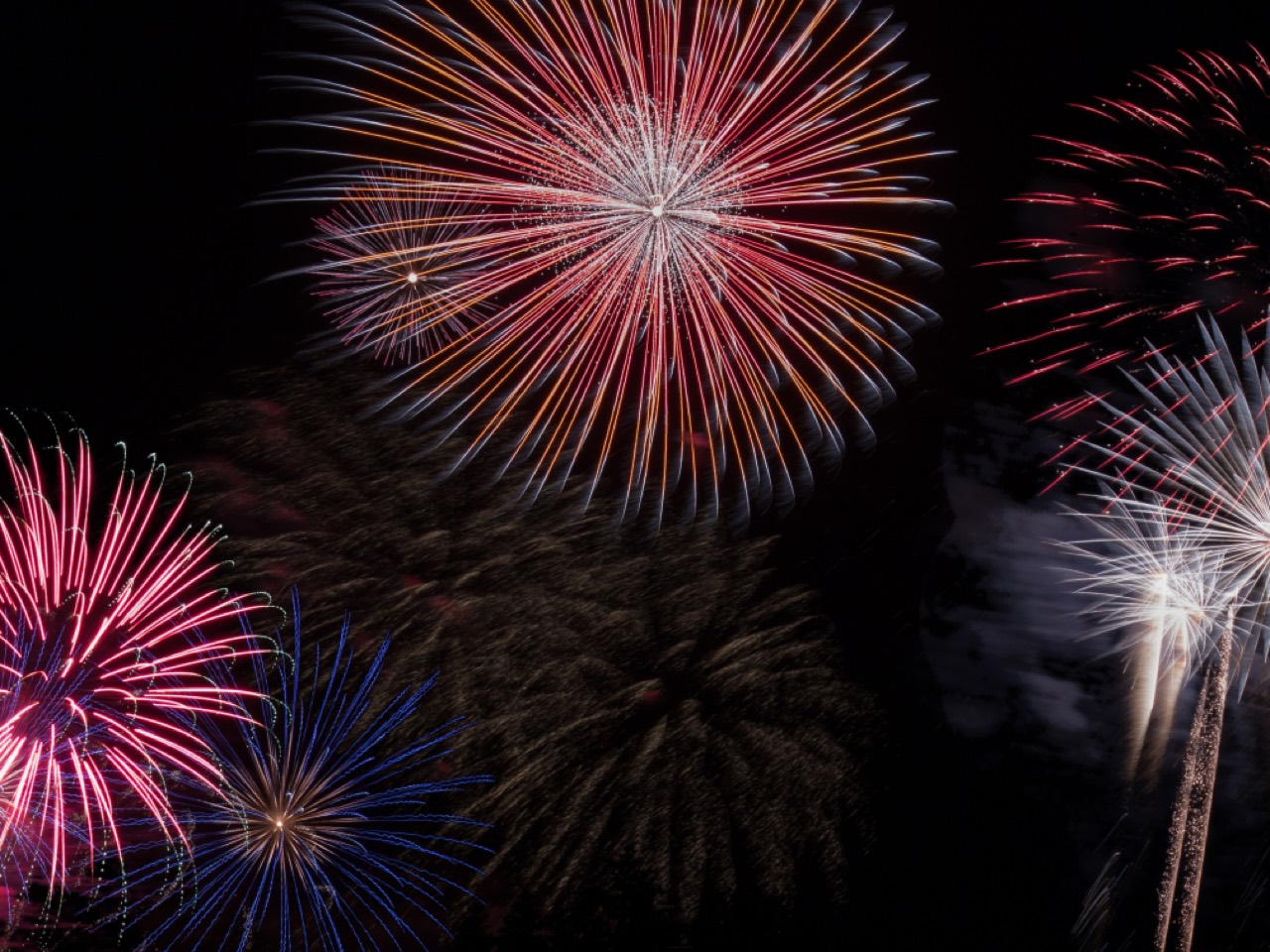} &
    \includegraphics[width=0.23\linewidth,height=0.1725\linewidth]{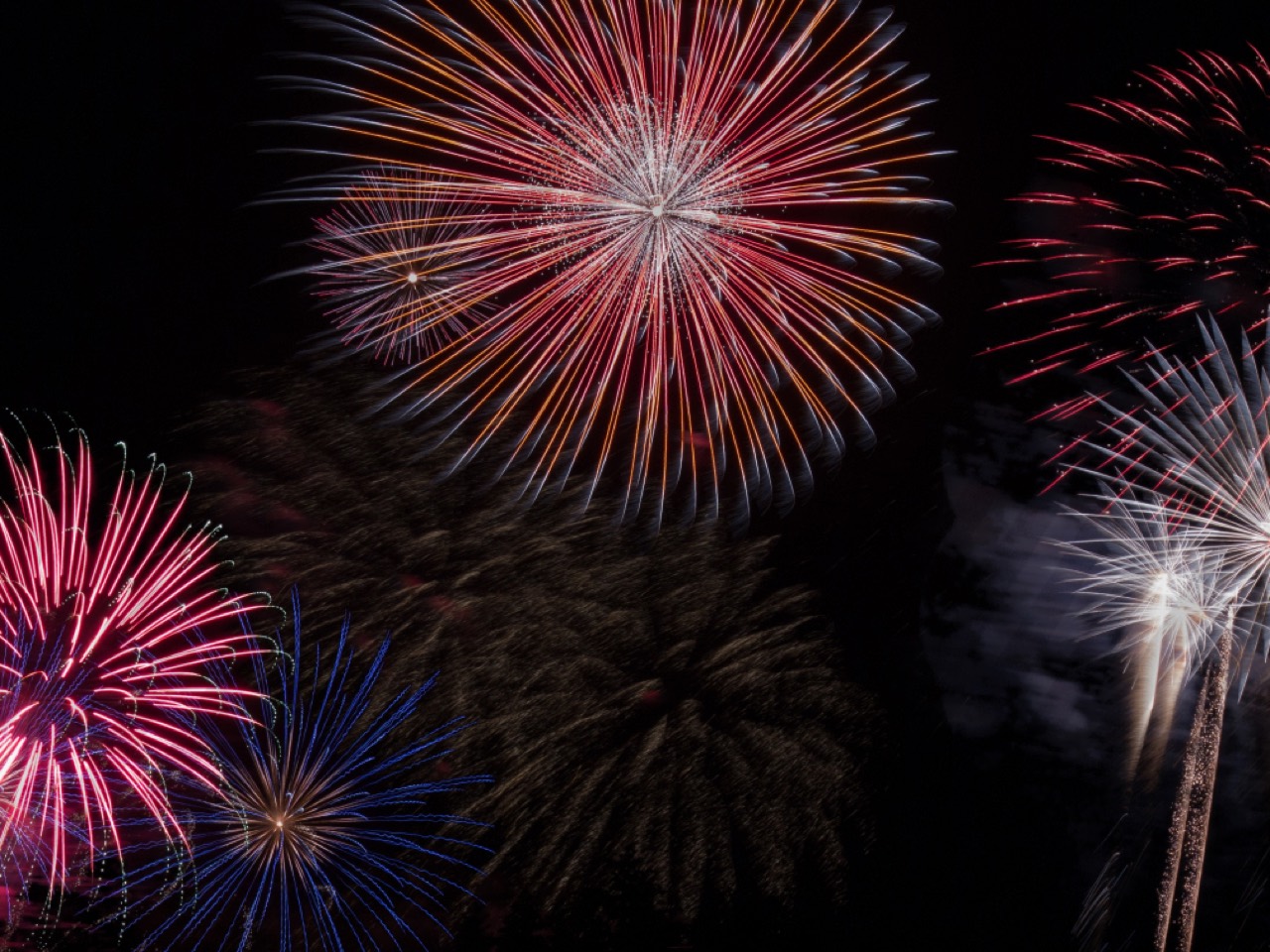} &
    \includegraphics[width=0.23\linewidth,height=0.1725\linewidth,trim=66 0 66 0,clip]{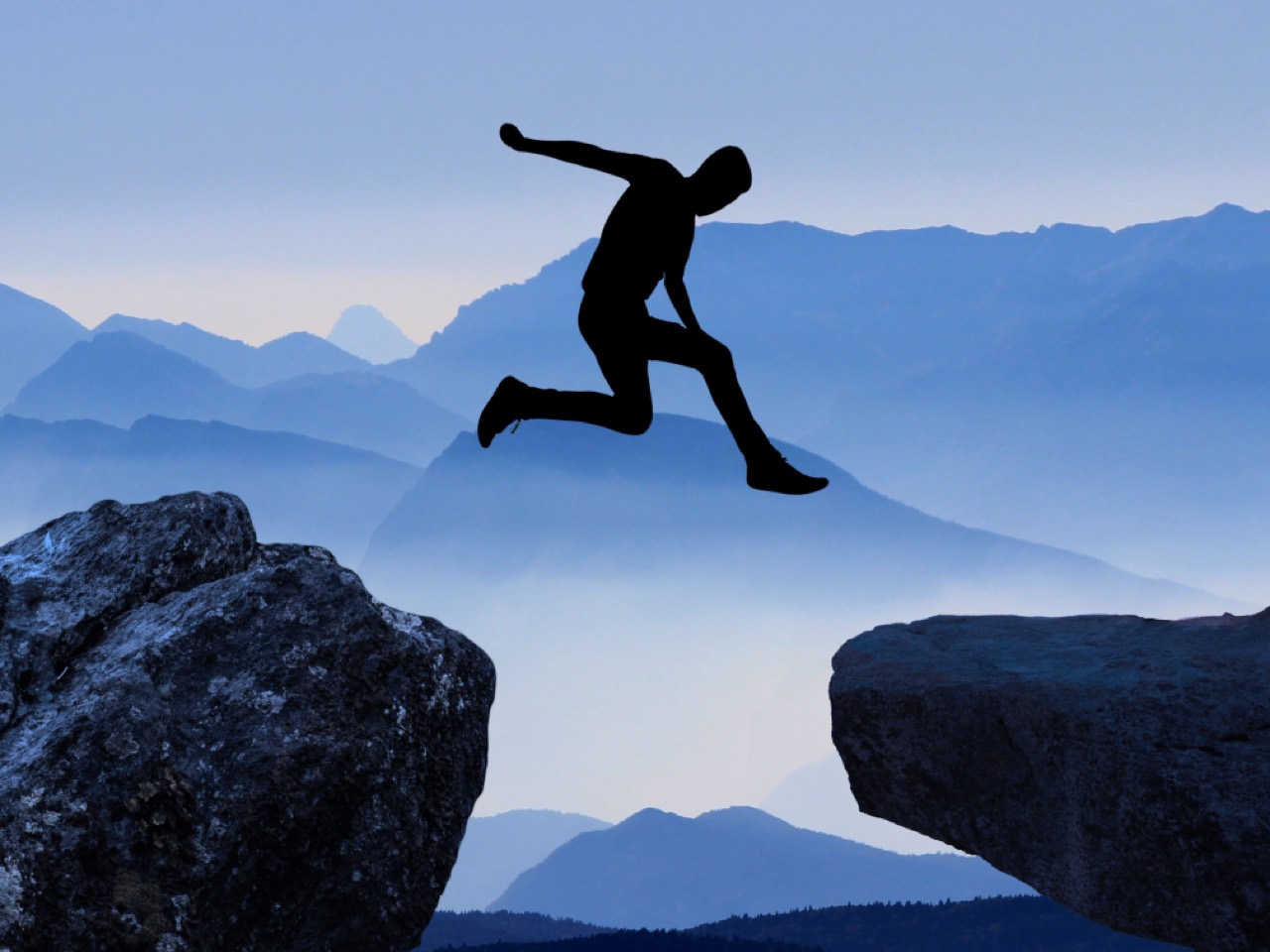} &
    \includegraphics[width=0.23\linewidth,height=0.1725\linewidth,trim=66 0 66 0,clip]{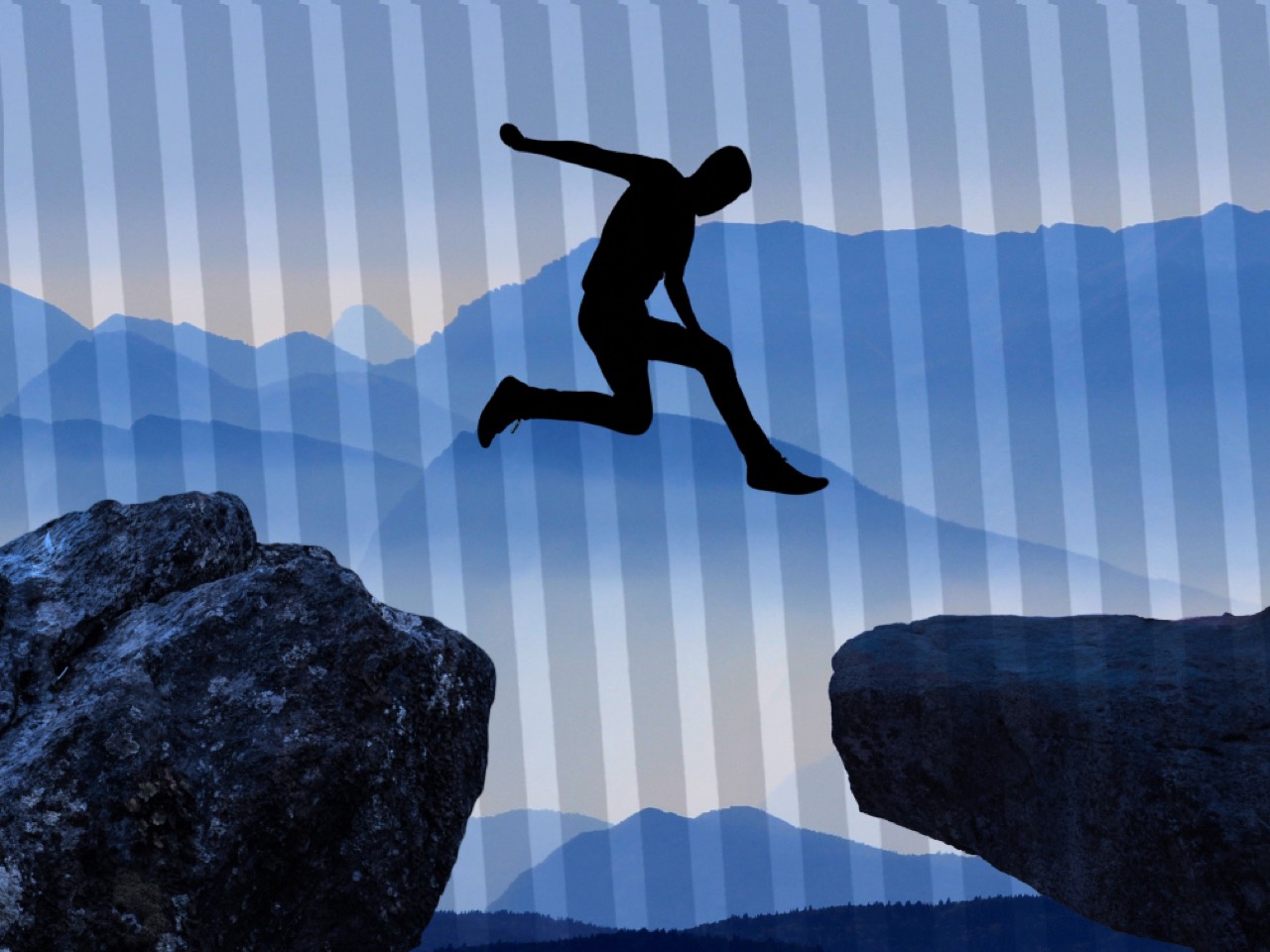} \\
    \small GT & \small LQ & \small GT & \small LQ \\
  \end{tabular}
  \caption{Our simulated dataset based on UHDM~\citep{yu2022towards} partial visualization (part 2).}
  \label{fig:simulated dataset visualization3}
\end{figure*}

Please refer to Fig.~\ref{fig:Real-world dataset visualization1}-\ref{fig:simulated dataset visualization3}

\clearpage

\subsection{Real-World Acquisition Settings}
\label{app:acquisition}
The five scenes were captured from commercial LED/OLED-type emissive advertising displays using a vivo X200s under indoor ambient lighting. Table~\ref{tab:acquisition-settings} gives the exposure and ISO settings reported for each scene. Short- and long-exposure pairs share a fixed viewpoint, with unchanged ambient lighting. These conditions represent the predefined dark-FB target rather than defining the applicable range after collection. Registration, valid-region cropping, visual screening, and paired patch extraction are described in Sec.~\ref{sec:methods}.

\begin{table}[tbp]
\centering
\small
\caption{Real-world acquisition settings for the 105 valid image pairs.}
\label{tab:acquisition-settings}
\begin{tabular}{crllll}
\toprule
Scene & Pairs & LQ exposure & LQ ISO & GT exposure & GT ISO \\
\midrule
1 & 15 & 1/2500 or 1/500 s & 360--1800 & 1/80 s & 250 \\
2 & 25 & 1/2500 or 1/800 s & 1000--2500 & 1/60 s & 72 \\
3 & 33 & 1/2000 s & 332--1440 & 1/80 s & 72 \\
4 & 27 & 1/2000 s & 417--2400 & 1/80 s & 72 \\
5 & 5 & 1/1600 s & 390--1152 & 1/80 s & 72 \\
\bottomrule
\end{tabular}
\end{table}

\subsection{Training and Inference Costs}
\label{app:compute}
Tables~\ref{tab:training-cost} and~\ref{tab:inference-cost} report the training settings used to obtain the compared checkpoints and the measured inference costs. Inference uses the same RTX A6000, $512\times512$ inputs, and batch size 1. The direct PiSA-SR/RIFLE comparison uses FP16 inference. Memory units in the training table follow the recorded measurements; the RIFLE training value is reported in GB, while the other recorded training values are in GiB.

\begin{table}[tbp]
\centering
\scriptsize
\caption{Training settings and costs for the compared checkpoints. N/A denotes a method without training.}
\label{tab:training-cost}
\begin{tabular}{lrrcrr}
\toprule
Method & Batch & Iterations & Learning rate & Peak memory & Time (h) \\
\midrule
Notch Filter & N/A & N/A & N/A & N/A & N/A \\
MAT & 2 & 45,000 & $2\times10^{-4}$ & 23.1 GiB & 16.2 \\
InvSR & 4 & 4,000 & $5\times10^{-5}$ & 30.8 GiB & 20.3 \\
Step1X & 8 & 25,000 & $10^{-4}$ & 20.5 GiB & 69.0 \\
ESDNet & 16 & 186,875 & $2\times10^{-4}$ & 26.4 GiB & 50.8 \\
FlickerFormer & 1 & 20,000 & $5\times10^{-5}$ & 17.7 GiB & 6.4 \\
PiSA-SR & 4 & 50,000 & $5\times10^{-5}$ & 43.2 GiB & 25.7 \\
RIFLE & 4 & 50,000 & $5\times10^{-5}$ & 43.3 GB & 25.8 \\
\bottomrule
\end{tabular}
\end{table}

\begin{table}[tbp]
\centering
\scriptsize
\caption{Inference costs on $512\times512$ inputs. M and G denote $10^6$ and $10^9$. Notch filtering runs on the CPU and is not included in GPU-memory or neural-operation profiling.}
\label{tab:inference-cost}
\begin{tabular}{lrrrrr}
\toprule
Method & Params (M) & Memory (GiB) & MACs (G) & FLOPs (G) & Latency (ms) \\
\midrule
Notch Filter & 0 & N/A & N/A & N/A & 58.4 \\
MAT & 0.690 & 0.4839 & 169 & 337 & 238.5 \\
InvSR & 983.402 & 3.9520 & 7,521 & 15,042 & 2,607.6 \\
Step1X & 12,511.620 & 39.9809 & 999,122 & 1,998,244 & 24,548.4 \\
ESDNet & 5.934 & 0.2238 & 70 & 140 & 64.1 \\
FlickerFormer & 4.155 & 3.7336 & 131 & 263 & 169.5 \\
PiSA-SR & 949.565 & 4.6580 & 2,177.477 & 4,354.953 & 278.1 \\
RIFLE & 950.804 & 4.6593 & 2,177.479 & 4,354.957 & 316.3 \\
\bottomrule
\end{tabular}
\end{table}

Relative to PiSA-SR, RIFLE adds 1.239 M parameters, 0.0013 GiB peak memory, 0.002 G profiled MACs, and 0.004 G profiled FLOPs. Latency increases by 38.2 ms (13.7\%), mainly because BSHP runs on the CPU; the neural-operation profiles do not capture this CPU preprocessing cost. These measurements characterize patch-level inference under the stated hardware and precision, rather than end-to-end full-resolution mobile inference. RIFLE remains substantially larger and more expensive than lightweight baselines; model compression and efficient prior computation are directions for reducing deployment cost.

\subsection{OCR-Based Content Fidelity}
\label{app:ocr-fidelity}
Following the OCR-A evaluation protocol used in TADiSR~\cite{hu2025tadisr}, we detect and recognize text in GT and use the recognized strings as reference transcriptions. We reuse exactly the same text regions in LQ, PiSA-SR, and RIFLE and recognize them using the same OCR model. OCR-A converts the Levenshtein edit distance, including insertions, deletions, and substitutions, to a normalized similarity score and averages over the text regions.

\begin{table}[tbp]
\centering
\caption{OCR-based text fidelity. Higher OCR-A is better.}
\label{tab:ocr-fidelity}
\begin{tabular}{lrrr}
\toprule
Metric & LQ & PiSA-SR & RIFLE \\
\midrule
OCR-A $\uparrow$ & 0.9594 & 0.8718 & 0.8841 \\
\bottomrule
\end{tabular}
\end{table}

RIFLE scores above PiSA-SR but below LQ. Both diffusion restorers therefore incur some text-fidelity loss relative to the input, despite improvements in visual quality. Character-level fidelity is not guaranteed, and outputs intended for text-sensitive use should be verified.

\subsection{Video Evaluation and Temporal Consistency}
\label{app:video-evaluation}
We evaluate PiSA-SR and RIFLE on the complete collected video debanding dataset: 108 videos, 9,849 paired frames, and 9,741 adjacent-frame transitions. All frames use the same $512\times512$ preprocessing, and both methods restore each frame independently without neighboring frames. We report the six frame-level image-quality metrics and the optical-flow-based warping error $E_{\mathrm{warp}}$ for temporal consistency. For each metric, we first average within each video and then report the unweighted mean over the 108 videos.

\begin{table}[tbp]
\centering
\scriptsize
\caption{Full-dataset video evaluation. $E_{\mathrm{warp}}$ values are in units of $10^{-3}$.}
\label{tab:video-evaluation}
\begin{tabular}{lrrrrrrr}
\toprule
Method & SSIM $\uparrow$ & MS-SSIM $\uparrow$ & LPIPS $\downarrow$ & DISTS $\downarrow$ & FSIM $\uparrow$ & GMSD $\downarrow$ & $E_{\mathrm{warp}}\downarrow$ \\
\midrule
LQ & 0.6266 & 0.7018 & 0.3279 & 0.1994 & 0.7668 & 0.2028 & 17.6871 \\
PiSA-SR & 0.4921 & 0.6319 & 0.4077 & 0.2373 & 0.7086 & 0.2418 & 33.7324 \\
RIFLE & 0.6227 & 0.7074 & 0.3397 & 0.2031 & 0.7709 & 0.1965 & 20.5202 \\
\bottomrule
\end{tabular}
\end{table}

RIFLE improves all six frame-level metrics relative to PiSA-SR and reduces $E_{\mathrm{warp}}$ by 39.2\%. Its warping error is closer to LQ, although a gap remains (20.5202 versus 17.6871). These results indicate encouraging temporal stability without an explicit temporal module, rather than guaranteed temporal consistency. Recent video diffusion models use cross-frame aggregation and video-level guidance for space-time super-resolution~\cite{chen2026diffst}, or spatial and temporal priors for face restoration~\cite{chen2026dvface}. A video extension of RIFLE could explore such temporal conditioning while retaining the current frame-restoration objective.

\section*{NeurIPS Paper Checklist}

\begin{enumerate}

\item {\bf Claims}
    \item[] Question: Do the main claims made in the abstract and introduction accurately reflect the paper's contributions and scope?
    \item[] Answer: \answerYes{}.
    \item[] Justification: In the abstract and introduction, we state the main contributions of this paper, including the simulation pipeline, the paired real-world benchmark, and the RIFLE model. We support these claims with the method in Sec.~\ref{sec:methods} and with the quantitative results, visual comparisons, and ablation study in Sec.~\ref{sec:experiments}.

\item {\bf Limitations}
    \item[] Question: Does the paper discuss the limitations of the work performed by the authors?
    \item[] Answer: \answerYes{}.
    \item[] Justification: We discuss limitations and future directions in Appendix~\ref{app:limitations}. We also note the diversity of display mechanisms and the difficulty of handling more complex banding patterns in practice in Appendix~\ref{app:display-patterns}.

\item {\bf Theory assumptions and proofs}
    \item[] Question: For each theoretical result, does the paper provide the full set of assumptions and a complete (and correct) proof?
    \item[] Answer: \answerNA{}.
    \item[] Justification: Our work is an empirical image restoration study and does not contain theorem-proof style theoretical results.

\item {\bf Experimental result reproducibility}
    \item[] Question: Does the paper fully disclose all the information needed to reproduce the main experimental results of the paper to the extent that it affects the main claims and/or conclusions of the paper (regardless of whether the code and data are provided or not)?
    \item[] Answer: \answerYes{}.
    \item[] Justification: We specify the problem setup in Sec.~\ref{sec:flicker-banding}, the degradation simulation process, dataset construction, model components, and loss design in Sec.~\ref{sec:methods}, the evaluation setting and ablations in Sec.~\ref{sec:experiments}, and additional details in Sec.~\ref{sec:appendix}. These details are sufficient to understand and reproduce our main experiments.

\item {\bf Open access to data and code}
    \item[] Question: Does the paper provide open access to the data and code, with sufficient instructions to faithfully reproduce the main experimental results, as described in supplemental material?
    \item[] Answer: \answerNo{}.
    \item[] Justification: We commit to publicly releasing the dataset and model code at \url{https://github.com/libozhu03/RIFLE}, as stated in the abstract; a release plan alone does not establish current open access.

\item {\bf Experimental setting/details}
    \item[] Question: Does the paper specify all the training and test details (e.g., data splits, hyperparameters, how they were chosen, type of optimizer) necessary to understand the results?
    \item[] Answer: \answerYes{}.
    \item[] Justification: We describe training data construction, model design, and loss design in Sec.~\ref{sec:methods}; test data, evaluation metrics, implementation details, and ablation settings in Sec.~\ref{sec:experiments}; and supplementary details in Sec.~\ref{sec:appendix}. We also report the key hyperparameters needed to interpret the results.

\item {\bf Experiment statistical significance}
    \item[] Question: Does the paper report error bars suitably and correctly defined or other appropriate information about the statistical significance of the experiments?
    \item[] Answer: \answerYes{}.
    \item[] Justification: Appendix~\ref{app:user-study} reports a decision-level 95\% Wilson confidence interval and a one-sided exact binomial test for user preferences, with the analysis unit and its limitations stated explicitly. Restoration metrics are single-run results; we do not claim across-run confidence intervals for them.

\item {\bf Experiments compute resources}
    \item[] Question: For each experiment, does the paper provide sufficient information on the computer resources (type of compute workers, memory, time of execution) needed to reproduce the experiments?
    \item[] Answer: \answerNo{}.
    \item[] Justification: Appendix~\ref{app:compute} reports training settings, peak memory, training time, and inference profiling on an RTX A6000 with the input size and batch size specified. However, the complete training hardware configuration and total development compute are not documented, so we retain the conservative \answerNo{} answer.
    
\item {\bf Code of ethics}
    \item[] Question: Does the research conducted in the paper conform, in every respect, with the NeurIPS Code of Ethics \url{https://neurips.cc/public/EthicsGuidelines}?
    \item[] Answer: \answerYes{}.
    \item[] Justification: To the best of our knowledge, our research complies with the NeurIPS Code of Ethics. We study an image restoration problem, credit prior assets, and do not involve sensitive personal data collection or high-risk deployment claims.

\item {\bf Broader impacts}
    \item[] Question: Does the paper discuss both potential positive societal impacts and negative societal impacts of the work performed?
    \item[] Answer: \answerNo{}.
    \item[] Justification: Our current manuscript does not include a dedicated discussion of broader societal impacts. Although this work can improve screen-photography readability and usability, we do not yet explicitly discuss potential positive and negative downstream impacts.
    
\item {\bf Safeguards}
    \item[] Question: Does the paper describe safeguards that have been put in place for responsible release of data or models that have a high risk for misuse (e.g., pre-trained language models, image generators, or scraped datasets)?
    \item[] Answer: \answerNA{}.
    \item[] Justification: We do not release a high-risk generative model, scraped dataset, or other asset that would typically require special misuse safeguards. Our contribution is an image restoration method and a paired benchmark for flicker-banding removal.

\item {\bf Licenses for existing assets}
    \item[] Question: Are the creators or original owners of assets (e.g., code, data, models), used in the paper, properly credited and are the license and terms of use explicitly mentioned and properly respected?
    \item[] Answer: \answerNo{}.
    \item[] Justification: We cite the original datasets and baseline methods that we build on, but we do not yet explicitly document the licenses or terms of use of all existing assets in the manuscript or appendix.

\item {\bf New assets}
    \item[] Question: Are new assets introduced in the paper well documented and is the documentation provided alongside the assets?
    \item[] Answer: \answerNo{}.
    \item[] Justification: We introduce a new paired real-world dataset and a code release plan, but documentation for these assets is not yet provided alongside an anonymized release in the current submission materials.

\item {\bf Crowdsourcing and research with human subjects}
    \item[] Question: For crowdsourcing experiments and research with human subjects, does the paper include the full text of instructions given to participants and screenshots, if applicable, as well as details about compensation (if any)? 
    \item[] Answer: \answerNo{}.
    \item[] Justification: We include a user study and provide screenshots of the evaluation interface in Appendix~\ref{app:user-study}, but we do not include the full participant instructions or compensation details in the current version.

\item {\bf Institutional review board (IRB) approvals or equivalent for research with human subjects}
    \item[] Question: Does the paper describe potential risks incurred by study participants, whether such risks were disclosed to the subjects, and whether Institutional Review Board (IRB) approvals (or an equivalent approval/review based on the requirements of your country or institution) were obtained?
    \item[] Answer: \answerNo{}.
    \item[] Justification: We include a user study, but our current manuscript does not describe participant-risk disclosure or any IRB or equivalent review status.

\item {\bf Declaration of LLM usage}
    \item[] Question: Does the paper describe the usage of LLMs if it is an important, original, or non-standard component of the core methods in this research? Note that if the LLM is used only for writing, editing, or formatting purposes and does \emph{not} impact the core methodology, scientific rigor, or originality of the research, declaration is not required.
    \item[] Answer: \answerNA{}.
    \item[] Justification: LLMs are not an important or non-standard component of our proposed method, experiments, or scientific claims.

\end{enumerate}

\end{document}